\documentclass[10pt,journal,compsoc]{IEEEtran}

\usepackage{cite}

\ifCLASSINFOpdf
\else
\fi
\usepackage{color}
\usepackage{tikz}
\usepackage{comment}
\usepackage{amsmath,amssymb,amssymb}
\usepackage{dsfont}
\usepackage{booktabs} 
\usepackage{mathrsfs}
\usepackage{multirow}
\usepackage{comment}
\usepackage{changepage}
\usepackage{wrapfig}
\usepackage{colortbl} 
\usepackage{arydshln} 
\usepackage[colorlinks,linkcolor=red]{hyperref}
\usepackage{cleveref}
\usepackage{enumitem}
\usepackage{pifont}

\usepackage[noend]{algpseudocode}
\usepackage{algorithmicx,algorithm}

\definecolor{darkgreen}{RGB}{0,0,0}

\usepackage[noend]{algpseudocode}
\usepackage{algorithmicx,algorithm}

\newtheorem{theorem}{Theorem}

\usepackage[switch]{lineno}

\begin{document}
%
\title{HiDe-PET: Continual Learning via Hierarchical Decomposition of Parameter-Efficient Tuning}
%
%
%

\author{Liyuan Wang, Jingyi Xie, Xingxing Zhang, Hang Su, \textit{Member}, \textit{IEEE}, Jun Zhu, \textit{Fellow}, \textit{IEEE} \thanks{Liyuan Wang, Jingyi Xie, Xingxing Zhang, Hang Su, and Jun Zhu are with Dept. of Comp. Sci. \& Tech., Institute for AI, BNRist Center, THBI Lab, Tsinghua-Bosch Joint Center for ML, Tsinghua University, Beijing, China (email: wly19@tsinghua.org.cn; jingyi\_xie96@163.com; xxzhang1993@gmail.com; \{suhangss, dcszj\}@tsinghua.edu.cn). Corresponding authors: Jun Zhu and Hang Su.} 
}

%
%

\markboth{Journal of \LaTeX\ Class Files,~Vol.~14, No.~8, August~2015}%
{Wang \MakeLowercase{\textit{et al.}}}
\IEEEtitleabstractindextext{%
\begin{abstract}

The deployment of pre-trained models (PTMs) has greatly advanced the field of continual learning (CL), enabling positive knowledge transfer and resilience to catastrophic forgetting. To sustain these advantages for sequentially arriving tasks, a promising direction involves keeping the pre-trained backbone frozen while employing parameter-efficient tuning (PET) techniques to instruct representation learning. 
Despite the popularity of Prompt-based PET for CL, its empirical design often leads to sub-optimal performance in our evaluation of different PTMs and target tasks. To this end, we propose a unified framework for CL with PTMs and PET that provides both theoretical and empirical advancements. We first perform an in-depth theoretical analysis of the CL objective in a pre-training context, decomposing it into hierarchical components namely within-task prediction, task-identity inference and task-adaptive prediction. We then present Hierarchical Decomposition PET (HiDe-PET), an innovative approach that explicitly optimizes the decomposed objective through incorporating task-specific and task-shared knowledge via mainstream PET techniques along with efficient recovery of pre-trained representations. Leveraging this framework, we delve into the distinct impacts of implementation strategy, PET technique and PET architecture, as well as adaptive knowledge accumulation amidst pronounced distribution changes. Finally, across various CL scenarios, our approach demonstrates remarkably superior performance over a broad spectrum of recent strong baselines. Our code is available at \url{https://github.com/thu-ml/HiDe-PET}.
\end{abstract}

}

\maketitle

\IEEEdisplaynontitleabstractindextext

%
\IEEEpeerreviewmaketitle

\IEEEraisesectionheading{\section{Introduction}\label{sec:introduction}}

\IEEEPARstart{T}he proficiency of artificial intelligence (AI) in accommodating real-world dynamics is largely contingent on its capability of continual learning (CL), which has benefited significantly in recent years from the deployment of pre-trained models (PTMs).
In essence, PTMs provide CL with not only positive knowledge transfer but also resilience to catastrophic forgetting \cite{ramasesh2021effect,mehta2021empirical,wang2023comprehensive,zhang2023slca}, which are  critical to improve the performance of CL methods.
Given that adapting PTMs to sequentially arriving tasks may compromise these advantages via progressive overwriting of pre-trained knowledge, numerous efforts have been devoted into implementing parameter-efficient tuning (PET) techniques for CL, i.e., keeping the pre-trained backbone frozen and introducing a few lightweight parameters to instruct representation learning. However, current advances predominantly center around empirical designs of Prompt-based PET \cite{wang2022learning_l2p,wang2022dualprompt,wang2022sprompts,smith2023coda}, which struggle to adequately achieve the CL objective and therefore often exhibit sub-optimal performance across different PTMs and target tasks (see Sec.~\ref{sec:exp_result}). In response to this critical challenge, there is an urgent need for a more profound understanding of CL with PTMs and PET, coupled with endeavors to enhance its effectiveness and generality.

In this work, we propose a unified framework for CL with PTMs and PET, seeking to advance this direction with both theoretical and empirical insights.
We initiate our explorations with an in-depth theoretical analysis of the CL objective in a pre-training context. Considering the significant impact of pre-trained knowledge on CL, this objective can be decomposed into three hierarchical components in response to sequentially arriving tasks, namely within-task prediction (WTP), task-identity inference (TII) and task-adaptive prediction (TAP). They prove to be sufficient and necessary to ensure low errors in common CL settings.
Based on the theoretical analysis, we devise an innovative approach named Hierarchical Decomposition PET (HiDe-PET) to explicitly optimize WTP, TII and TAP.

The principal concept behind HiDe-PET leverages the unique strengths of PTMs for CL, with a particular focus on the effective instruction and efficient recovery of pre-trained representations. As a generic approach applicable to mainstream PET techniques (e.g., Prompt \cite{lester2021power,li2021prefix}, Adapter \cite{rebuffi2017learning}, LoRA \cite{hu2021lora}, etc.), we construct an ensemble of task-specific parameters that incorporates the knowledge of each task to optimize WTP, and a set of task-shared parameters that accumulates knowledge in a task-agnostic manner to improve TII.
We further recover the distribution of uninstructed and instructed representations through preserving their statistical information, so as to optimize TII and TAP, respectively. In this way, our HiDe-PET is able to adeptly predict the identity of task-specific parameters from uninstructed representations and collect appropriate instructed representations for final predictions.

The proposed framework facilitates a thorough assessment of important factors emerged in CL with PTMs and PET, including the implementation strategy, PET technique and PET architecture. For example, we dissect representative strategies for stabilizing task-shared parameters and for preserving pre-trained representations, empirically analyzing their effectiveness.
Moreover, we evaluate the behaviors of different PET techniques in achieving the CL objective, where the Prompt-based PET tends to be less effective in WTP, consequently displaying lower sensitivity to TII and clearly lagging behind the LoRA/Adapter-based PET. Motivated by the inherent connections between TII and out-of-distribution (OOD) detection, we further devise a PET hierarchy tailored for adaptive knowledge accumulation, and unravel the relationship between task-specific and task-shared PET architectures in representation learning.

\begin{figure}[ht]
    \centering
    \vspace{-0.1cm}
    \includegraphics[width=0.48\textwidth]{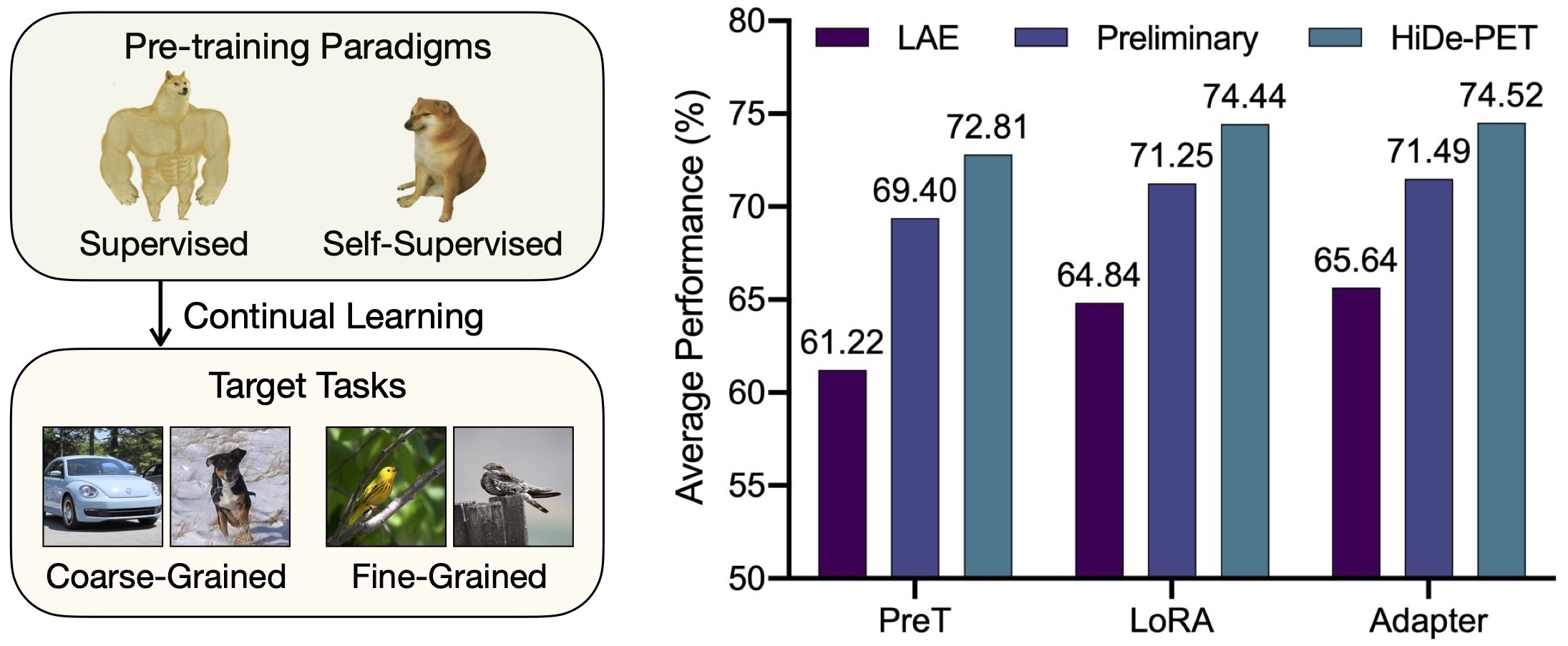}
    \vspace{-0.1cm}
\caption{Summary of CL performance with different PET techniques. We compare our HiDe-PET, our preliminary version \cite{wang2023hierarchical} and LAE \cite{gao2023unified}, and report the final average accuracy (FAA) over three pre-trained checkpoints and four CL benchmarks.} 
\vspace{-0.1cm}
\label{Comparison_PET}
\end{figure}

Upon extensive experiments, our HiDe-PET clearly outperforms a wide range of recent strong baselines, and demonstrates remarkable generality across a variety of PET techniques, pre-trained checkpoints and CL benchmarks (summarized in Fig.~\ref{Comparison_PET}). We further provide empirical analysis to elucidate the respective contributions of the three hierarchical components. 
Note that some results have been presented in our preliminary version \cite{wang2023hierarchical}, which mainly considered a specific implementation of task-specific parameters via Prompt-based PET. In contrast, the current version presents a unified framework for CL with PTMs and PET. It incorporates substantial extensions to the implementation strategy, PET technique and PET architecture, culminating in a comprehensive analysis and improved performance.

Overall, our main contributions are as follows:
\begin{itemize}[nolistsep] 
\item We present an in-depth theoretical analysis of the CL objective in a pre-training context, decomposing it into three hierarchical components for model design; 
\item We propose an innovative approach that exploits mainstream PET techniques and pre-trained representations to explicitly optimize the decomposed objective; 
\item We conduct extensive experiments to demonstrate the effectiveness and generality of our approach, coupled with a thorough assessment of important factors emerged in CL with PTMs and PET.
\end{itemize}

\section{Related Work}\label{sec:related}

\textbf{Continual Learning (CL)} is characterized by learning sequentially arriving tasks and performing well on them. The primary challenge of CL stems from the dynamic nature of data distribution, which leads to catastrophic forgetting of old tasks while acquiring new ones \cite{wang2023comprehensive,parisi2019continual}.
As summarized in a recent survey \cite{wang2023comprehensive}, representative methods include selective stabilization of important parameters for old tasks \cite{kirkpatrick2017overcoming,wang2021afec,aljundi2018memory_mas}, approximation and recovery of old data distributions \cite{rebuffi2017icarl,shin2017continual_dgr,wang2021memory}, exploitation of robust and well-distributed representations \cite{pham2021dualnet,cha2021co2l,ostapenko2022foundational}, manipulation of optimization programs via gradient projection \cite{lopez2017gradient_gem,wang2021training,saha2020gradient}, construction of (relatively) dedicated parameters for each task \cite{serra2018overcoming,wang2022coscl,wang2023incorporating}, etc. 

In the realm of CL, current efforts have predominantly centered around computer vision (CV), specifically for visual classification tasks.
These efforts have progressively expanded their scope to include more complex visual tasks, as well as natural language processing (NLP), reinforcement learning (RL) and other related applications. 
Across various representative CL settings, especially those lacking the oracle of task identity during the testing phase, robust CL models often necessitate the storage and rehearsal of old training samples \cite{wu2019large_bic,knoblauch2020optimal,wang2021memory}, which creates additional resource overheads and potential privacy concerns. These issues have been largely avoided through effective use of pre-trained knowledge in recent work, as discussed later.

\textbf{Pre-training and Fine-tuning}: Transfer learning with pre-trained models (PTMs) can significantly improve the performance of target tasks and has therefore become a prevalent paradigm in many areas of artificial intelligence (AI). 
Since the pre-trained knowledge is usually generalized and difficult to cover all specific domains, PTMs necessitate further fine-tuning for better adaptation. The most straightforward way is to update all model parameters, but involves keeping a separate copy of fine-tuned model parameters for each task. This leads to considerable resource overheads especially for advanced PTMs of increasing size.

In order to improve the efficiency of fine-tuning, some lightweight alternatives have been proposed that update only a few extra parameters with most of the model parameters frozen, collectively referred to as the parameter-efficient tuning (PET) techniques. These PET techniques were originally proposed for NLP and are now widely used for CV as well. Representative practices include \emph{Prompt Tuning} (ProT) \cite{lester2021power} and \emph{Prefix Tuning} (PreT) \cite{li2021prefix} that prepend short prompt tokens into the original inputs or intermediate inputs, \emph{Adapter} \cite{rebuffi2017learning} that inserts small neural modules between backbone layers, \emph{Low-Rank Adaptation} (LoRA) \cite{hu2021lora} that approximates the updates of model parameters with low rank matrices, etc. A recent work \cite{he2021towards} unified these PET techniques in a similar form, corresponding to modifying specific hidden states of the PTMs.

\textbf{CL with PTMs and PET:}
While much of the past work in CL has focused on training from scratch, a growing body of efforts have delved into the benefits of PTMs, which provide not only positive knowledge transfer but also resilience to catastrophic forgetting \cite{ramasesh2021effect, mehta2021empirical}. 
Meanwhile, PTMs also need to improve the ability of CL to accommodate sequentially arriving tasks and to refine pre-trained knowledge from them. 
However, fine-tuning PTMs becomes much more difficult when considering CL, since repetitive adaptation of the same PTM may lead to progressive overwriting of pre-trained knowledge, whereas separate saving of the fine-tuned PTMs creates an additional linear increase in resource overhead on the time scale \cite{zhang2023slca}.

Therefore, applying PET techniques for CL has become an emerging direction in recent years, with Prompt-based PET being predominant. 
Many state-of-the-art methods have focused on designing appropriate prompt architectures for CL, such as task-shared prompts \cite{wang2022dualprompt,mcdonnell2023ranpac,gao2023unified}, task-specific prompts \cite{wang2022dualprompt,wang2022sprompts,wang2023hierarchical,smith2023coda} and their combinations \cite{wang2022learning_l2p,wang2022dualprompt}.
Since the frozen backbone with pre-trained knowledge can provide robust and well-distributed representations, these methods have almost achieved the performance pinnacle of rehearsal-free CL under adequate supervised pre-training and general tasks.
However, their sub-optimality in achieving the objective of CL has been clearly exposed under the more realistic self-supervised pre-training and fine-grained tasks \cite{zhang2023slca,wang2023hierarchical}, in part due to the limited fitting ability of Prompt-based PET \cite{jia2022visual,yoo2023improving}.
LAE \cite{gao2023unified} is a recent work that assembles a pair of task-shared parameters to be implemented with mainstream PET techniques, but exhibits only moderate improvements in CL performance. 

\section{Preliminaries}\label{sec:preliminaries}
In this section, we first introduce the problem formulation of continual learning (CL) in a pre-training context. Then we describe representative parameter-efficient tuning (PET) techniques and PET-based CL methods.  

\subsection{Problem Formulation}\label{sec:problem}
The CL problem can be generally defined as learning sequentially arriving tasks from their respective training sets $\mathcal{D}_1, ..., \mathcal{D}_t$ in order to perform well on their corresponding test sets. The training set and test set of each task are assumed to follow the same distribution.
For supervised CL, each training set $\mathcal{D}_{t} = \{(\boldsymbol{x}_{t,n}, y_{t,n})\}_{n=1}^{N_t}$ comprises multiple sample-label pairs, where $\boldsymbol{x}_{t, n} \in \mathcal{X}_t$ and $y_{t, n} \in \mathcal{Y}_t$ represent the sample and label elements, respectively, and $N_t$ denotes the size of $\mathcal{D}_{t}$. 
Let us consider a neural network model composed of a backbone $f_\theta$ with parameters $\theta$, and an output layer $h_\psi$ with parameters $\psi$. 
This model aims to learn a projection from $\mathcal{X} = \bigcup_{i=1}^{t} \mathcal{X}_i$ to $\mathcal{Y} = \bigcup_{i=1}^{t} \mathcal{Y}_i$, so that it can correctly predict the label $\hat{y} = h_\psi(f_\theta(\boldsymbol{x})) \in \mathcal{Y}$ of an unseen test sample $\boldsymbol{x}$ from observed tasks.
Since $\mathcal{D}_1, ..., \mathcal{D}_t$ are usually limited in size and distinct in distribution, learning $f_\theta$ from scratch can easily converge to an undesirable local minimum. In contrast, initialization of $f_\theta$ with a substantial quantity of training samples external to $\mathcal{D}_1, ..., \mathcal{D}_t$, i.e., applying adequate pre-training, helps $\theta$ converge to a wide low-error region and thus can greatly improve the CL performance \cite{ramasesh2021effect,mehta2021empirical}.

Depending on the split of label space and the availability of task identity, there are three representative setups for CL \cite{van2019three}, including task-incremental learning (TIL), domain-incremental learning (DIL), and class-incremental learning (CIL). Specifically, the label space $\mathcal{Y}_i$ with $i \in [t]$ is the same for DIL whereas disjoint for TIL and CIL. The task identity $i \in [t]$ is provided at test time for TIL whereas not available for DIL and CIL. As a result, CIL is often considered more challenging and realistic. Of note, although CIL is named from classification tasks, its definition can be generalized to other task types. To avoid additional resource overhead and potential privacy issues, the CL process is further restricted to be \emph{rehearsal-free} \cite{smith2023coda}, i.e., the sample elements of each $\mathcal{D}_{i}$ are available only when learning task $i$, which particularly increases the challenge of CIL \cite{wang2023comprehensive}.

\subsection{PET Techniques}\label{sec:pet_tech}
The backbone $f_\theta$ of advanced pre-trained models (PTMs) often employs a transformer architecture \cite{vaswani2017attention} based on multi-head attention mechanisms. 
For example, a pre-trained vision transformer (ViT) \cite{dosovitskiy2020image_vit} consists of multiple consecutive multi-head self-attention (MSA) layers that transform an input sample into a sequence-like output representation $\boldsymbol{r} \in \mathbb{R}^{d_{\boldsymbol{r}} \times d}$ of sequence length $d_{\boldsymbol{r}}$ and embedding dimension $d$. Let us consider the $l$-th layer with input $\boldsymbol{h}_{(l)}$ and output $\boldsymbol{h}_{(l)}'$, where $\boldsymbol{h}_{(L)}'$ is equivalent to $\boldsymbol{r}$ for total $L$ layers.
Since the output $\boldsymbol{h}_{(l)}'$ then becomes the input $\boldsymbol{h}_{(l+1)} \in \mathbb{R}^{d_{\boldsymbol{h}_{(l+1)}} \times d}$ of the next layer, we omit the layer identity $l$ for the sake of clarity. 
Then, the input $\boldsymbol{h} \in \mathbb{R}^{d_{\boldsymbol{h}} \times d}$ is further specified as the query $\boldsymbol{h}_Q$, key $\boldsymbol{h}_K$ and value $\boldsymbol{h}_V$, and the output $\boldsymbol{h}' \in \mathbb{R}^{d_{\boldsymbol{h}} \times d}$ of the current layer is
\begin{small}
\begin{equation}
\begin{split}
    \boldsymbol{h}' = {\rm{MSA}}(\boldsymbol{h}_Q, \boldsymbol{h}_K, \boldsymbol{h}_V) = {\rm{Concat}} (h_1, ..., h_m) \boldsymbol{W}_O,\\
\end{split}
\end{equation}
\vspace{-0.4cm}
\begin{equation}
\begin{split}
    h_i = {\rm{Attn}} (\boldsymbol{h}_Q \boldsymbol{W}_{Q,i}, \boldsymbol{h}_K \boldsymbol{W}_{K,i}, \boldsymbol{h}_V \boldsymbol{W}_{V,i}), i\in[m],
\end{split}
\end{equation}
\end{small}
where $\boldsymbol{W}_O$, $\boldsymbol{W}_{Q,i}$, $\boldsymbol{W}_{K,i}, \boldsymbol{W}_{V,i}$ are projection matrices, $m$ is the number of heads, and $\boldsymbol{h}_Q = \boldsymbol{h}_K = \boldsymbol{h}_V = \boldsymbol{h}$ in ViT. The concatenation (Concat) and attention (Attn) functions are specified in their original papers \cite{vaswani2017attention,dosovitskiy2020image_vit}. 

To facilitate effective transfer of pre-trained knowledge while preventing its catastrophic forgetting, the backbone parameters $\theta$ are usually frozen and additional lightweight parameters are introduced to instruct representation learning, referred to as the PET techniques \cite{he2021towards}. Here we describe some representative ones (see Fig.~\ref{PET_Techniques}):

\begin{figure}[t]
    \centering
    \includegraphics[width=0.42\textwidth]{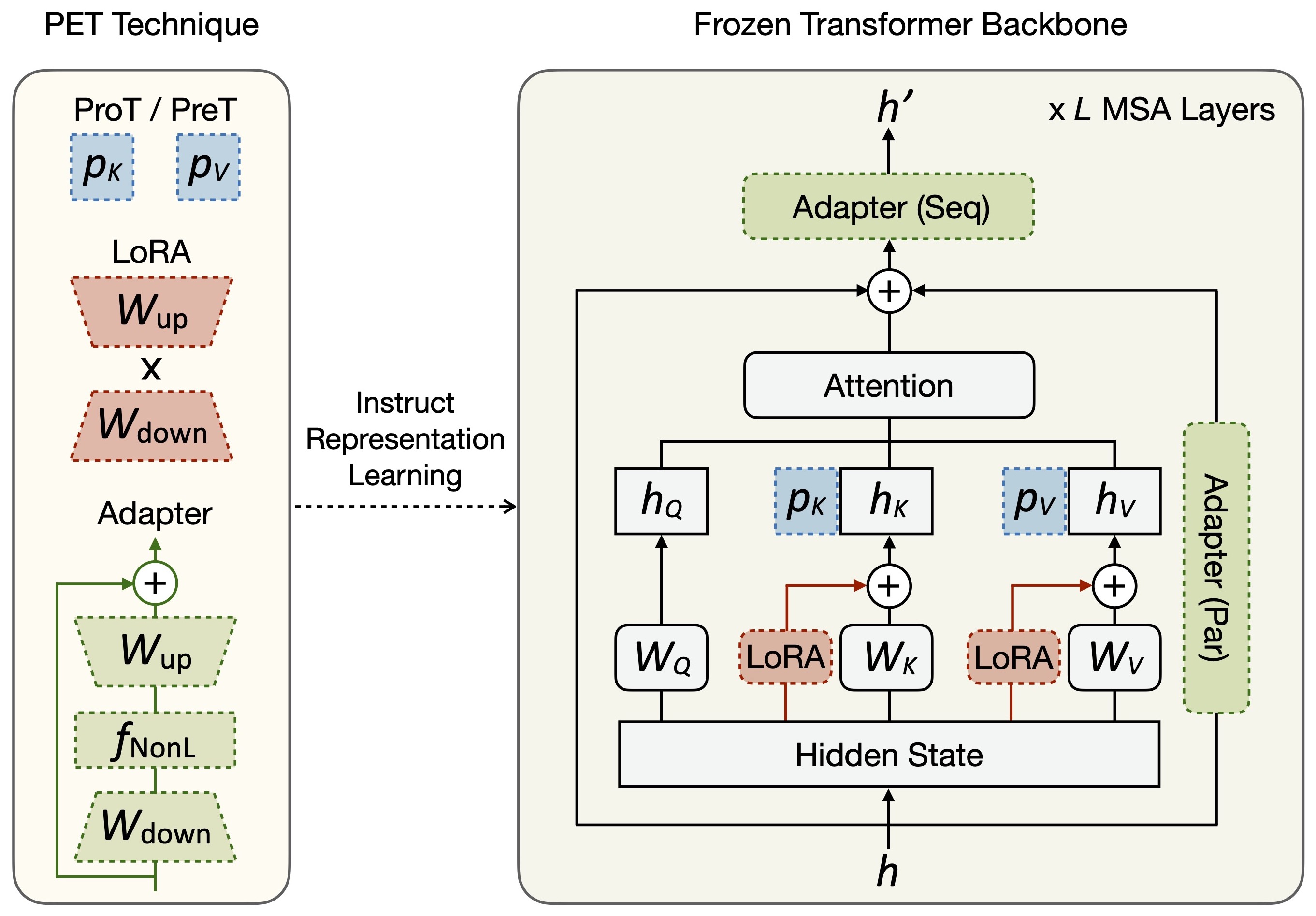}
    \vspace{-0.2cm}
\caption{Implementation of PET techniques for representation learning. These PET techniques all mount to modulating the (intermediate) representations of the backbone and ensure lightweight implementations. 
}
\vspace{-0.3cm}
\label{PET_Techniques}
\end{figure}

\textbf{Prompt Tuning} (ProT) \cite{lester2021power} and \textbf{Prefix Tuning} (PreT) \cite{li2021prefix} both prepend a few learnable parameters $\boldsymbol{p} \in \mathbb{R}^{d_{\boldsymbol{p}} \times d}$ of sequence length $d_{\boldsymbol{p}}$ and embedding dimension $d$ to $\boldsymbol{h}$, collectively known as the \emph{Prompt-based PET}. For ProT in ViT, an identical $\boldsymbol{p}$ is prepended to $\boldsymbol{h}_Q$, $\boldsymbol{h}_K$ and $\boldsymbol{h}_V$:
\begin{equation}
    \boldsymbol{h}' = {\rm{MSA}}([\boldsymbol{p};\boldsymbol{h}_Q], [\boldsymbol{p};\boldsymbol{h}_K], [\boldsymbol{p};\boldsymbol{h}_V]),
\end{equation}
where $[\cdot \, ; \cdot]$ represents the concatenation operation along the dimension of sequence length. Since the output in $\mathbb{R}^{(d_{\boldsymbol{h}}+d_{\boldsymbol{p}}) \times d}$ has increased dimensions, ProT is often used for only the last layer in CL \cite{wang2022learning_l2p,wang2022sprompts}. In contrast, PreT splits $\boldsymbol{p}$ into $\boldsymbol{p}_K \in \mathbb{R}^{d_{\boldsymbol{p}}/2 \times d}$ for $\boldsymbol{h}_K$ and $\boldsymbol{p}_V \in \mathbb{R}^{d_{\boldsymbol{p}}/2 \times d}$ for $\boldsymbol{h}_V$:
\begin{equation}
    \boldsymbol{h}' = {\rm{MSA}}(\boldsymbol{h}_Q, [\boldsymbol{p}_K;\boldsymbol{h}_K], [\boldsymbol{p}_V;\boldsymbol{h}_V]),
\end{equation}
where the output has the same dimension as the input $\boldsymbol{h} \in \mathbb{R}^{d_{\boldsymbol{h}} \times d}$, allowing PreT to be implemented in multiple layers.
In particular, the output of PreT can be reframed as 
\begin{equation}
\boldsymbol{h}' \leftarrow  (1-\lambda(\boldsymbol{h})) \boldsymbol{h}' + \lambda(\boldsymbol{h}) \, f_{{\rm{NonL}}}(\boldsymbol{h}\boldsymbol{W}_{Q}\boldsymbol{p}_K^{\top})\boldsymbol{p}_V,
\label{Eq_h_PreT}
\end{equation}
where $f_{{\rm{NonL}}}$ is the nonlinear (NonL) softmax function, and $\lambda(\boldsymbol{h})$ is a scalar that depends on the input \cite{he2021towards}.

\textbf{Adapter} \cite{rebuffi2017learning} inserts lightweight neural modules between backbone layers, each usually composed of a down-projection matrix $\boldsymbol{W}_{{\rm{down}}} \in \mathbb{R}^{d \times r}$ that reduces the dimension of $\boldsymbol{h}$ with bottleneck $r$, a nonlinear (NonL) activation function $f_{{\rm{NonL}}}$ and an up-projection matrix $\boldsymbol{W}_{{\rm{up}}} \in \mathbb{R}^{r \times d}$. These modules are implemented with residual connections, acting on the output $\boldsymbol{h}'$ in a \emph{sequential} (Seq) manner, i.e., 
\begin{equation}
\boldsymbol{h}' \leftarrow \boldsymbol{h}' + f_{{\rm{NonL}}}(\boldsymbol{h}' \boldsymbol{W}_{{\rm{down}}}) \boldsymbol{W}_{{\rm{up}}},
\label{Eq_h_Adapter_S}
\end{equation}
as well as on the input $\boldsymbol{h}$ in a \emph{parallel} (Par) manner, i.e.,
\begin{equation}
\boldsymbol{h}' \leftarrow \boldsymbol{h}' + f_{{\rm{NonL}}}(\boldsymbol{h} \boldsymbol{W}_{{\rm{down}}}) \boldsymbol{W}_{{\rm{up}}}.
\label{Eq_h_Adapter_P}
\end{equation}

\textbf{LoRA} \cite{hu2021lora} approximates the updates of pre-trained parameter matrix $\boldsymbol{W} \in \mathbb{R}^{d \times k}$ with a low-rank decomposition $\boldsymbol{W} + \bigtriangleup \boldsymbol{W} = \boldsymbol{W} + \boldsymbol{W}_{{\rm{down}}} \boldsymbol{W}_{{\rm{up}}}$, where $\boldsymbol{W}_{{\rm{down}}} \in \mathbb{R}^{d \times r}$, $\boldsymbol{W}_{{\rm{up}}} \in \mathbb{R}^{r \times k}$ and $r$ is the low-rank bottleneck. For ViT, LoRA is typically used to update the $\boldsymbol{W}_Q$ and $\boldsymbol{W}_V$ of a backbone layer. 
As a special case, when $\boldsymbol{W}$ performs linear projection of the input $\boldsymbol{h}$, the output is modified as 
\begin{equation}
\boldsymbol{h}' \leftarrow \boldsymbol{h}' + s \cdot \boldsymbol{h} \boldsymbol{W}_{{\rm{down}}} \boldsymbol{W}_{{\rm{up}}},
\label{Eq_h_LoRA}
\end{equation}
where $s \geq 1$ is a scalar hyperparameter \cite{he2021towards}.  

As we can see, these PET techniques all amount to modulating the (intermediate) representations of $f_\theta$, though differing in their specific implementations.

\subsection{PET-Based CL Methods}\label{sec:pet_method}

With the widespread use of PTMs in CL, there have been a variety of methods that incorporate PET techniques on a continual basis.
Most of these methods have focused on designing appropriate PET architectures to formulate lightweight parameters tailored for CL, which can be conceptually summarized into task-specific parameters, task-shared parameters, and their explicit or implicit combinations.
In particular, task-specific parameters require their identity to be predicted during the testing phase, while task-shared parameters need to mitigate catastrophic forgetting when learning each task. We briefly describe these methods here, with a comprehensive summary in Appendix Table~\ref{table:survey} for further comparison.

\textbf{L2P} \cite{wang2022learning_l2p} constructs a prompt pool $ \boldsymbol{p}_1, ..., \boldsymbol{p}_M $ of size $M$ and instructs pre-trained representations in a ProT manner. Each prompt is associated with a learnable key, optimized by the cosine distance of the top-$N$ keys to a query function $q(\boldsymbol{x}) = f_{\theta}(\boldsymbol{x})[0]$. The most relevant prompt(s) can therefore be selected via uninstructed representations.

\textbf{DualPrompt} \cite{wang2022dualprompt} employs the task-specific prompt $\boldsymbol{p}^{l}_{t}$ and task-shared prompt $\boldsymbol{p}^{l}$ to instruct respective layers in a PreT manner. The layer-wise $\boldsymbol{p}^{l}_{t}$ is associated with a task-specific key, optimized by its cosine distance to $q(\boldsymbol{x})$, and the best-matched key is selected via uninstructed representations.

\textbf{S-Prompt} \cite{wang2022sprompts} employs only the task-specific prompt $\boldsymbol{p}_{t}$ and instructs pre-trained representations in a ProT manner. The task identity is inferred from preserved task centroids with $k$-Nearest Neighbors ($k$NNs). S-Prompt also employs a multi-head output layer associated with the task identity.

\textbf{CODA-Prompt} \cite{smith2023coda} performs a weighted summation of the prompt pool, i.e., $\boldsymbol{p} = \sum_{i=1}^{M} \alpha_i \boldsymbol{p}_i$, and instructs multiple layers in a PreT manner. Each weighting factor $\alpha_i$ for $i \in [M]$ is associated to a learnable key, optimized by its cosine distance to $q(\boldsymbol{x})$. The inference of $\alpha_i$ can therefore construct an appropriate prompt for each input sample. 

\textbf{LAE} \cite{gao2023unified} employs two kinds of task-shared parameters to incorporate knowledge from more recent tasks and more remote tasks, respectively, applicable to PreT, Adapter and LoRA for multiple layers. Their update speeds are regulated via combinatorial strategies such as temporary freezing and exponential moving average (EMA).

Besides, there are several relevant methods with different focuses. 
For example, SLCA \cite{zhang2023slca} updates the entire backbone with a reduced learning rate, and further preserves pre-trained representations via dedicated covariance matrices.
RanPAC \cite{mcdonnell2023ranpac} projects pre-trained representations to a high-dimension space and preserves them via a shared covariance matrix. 
These methods are often parameter-inefficient, i.e., the parameter cost is comparable to $\theta$ due to a complexity much larger than $O(d^2)$, and therefore not prioritized in this work.\footnote{Our preliminary version \cite{wang2023hierarchical} also employed dedicated covariance matrices as the main implementation to acquire better performance.} 

Taken together, three notable limitations have surfaced in current progress of harnessing PET techniques for CL. First, the above methods often rely on empirical designs, making it difficult to ascertain their effectiveness in achieving the objective of CL in a pre-training context. In particular, their performance exhibits significant variations across different PTMs and target tasks, as demonstrated in Sec.~\ref{sec:exp_result}. 
Second, these methods predominantly center around Prompt-based PET, which has been shown to be less effective under self-supervised pre-training \cite{yoo2023improving} and fine-grained tasks \cite{ma2023visual}, leaving underexplored the particular behaviors and potential benefits of other alternatives. 
Third, these methods share some analogous strategies, such as stabilizing task-shared parameters and recovering pre-trained representations, without a comprehensive analysis of different implementations and assimilation of their respective strengths.
Therefore, there is an urgent need to establish a unified framework that incorporates both theoretical and empirical insights for CL with PTMs and PET, which constitutes our main motivation.

\section{Theoretical Analysis}\label{sec:theory}
In this section, we present an in-depth theoretical analysis of the CL objective in a pre-training context, so as to inspire the better design of PET-based CL methods. We first decompose this objective into three hierarchical components, which demonstrate the impact of pre-trained knowledge on CL and are both sufficient and necessary to optimize the CL performance (Sec.~\ref{sec:hier_decom}). This analysis motivates us to develop an innovative CL method to explicitly optimize the objective (Sec.~\ref{sec:three_component}). We then illustrate the connection of the decomposed objective to OOD detection, which is shown to play a similar role as inferring the task identity (Sec.~\ref{sec:tii_ood}). This analysis motivates us to adaptively accumulate knowledge with the proposed CL method to facilitate the learning of subsequent tasks (Sec.~\ref{sec:ood_method}).

\subsection{Three Hierarchical Components}\label{sec:hier_decom}

Let us consider a CL problem for sequentially arriving tasks $i \in [t]$ with training set $\mathcal{D}_{i}$ of domain $\mathcal{X}_i=\bigcup_{j}\mathcal{X}_{i,j}$ and label $\mathcal{Y}_i=\bigcup_{j}\mathcal{Y}_{i,j}$, where $j \in [|\mathcal{Y}_i|]$ denotes the $j$-th class of task $i$. The goal is to learn a projection from $\mathcal{X} = \bigcup_{i=1}^{t} \mathcal{X}_i$ to $\mathcal{Y} = \bigcup_{i=1}^{t} \mathcal{Y}_i$ in order to predict the label of an unseen test sample $\boldsymbol{x}$ from all observed tasks, referred to as task-adaptive prediction (TAP). As summarized in Sec.~\ref{sec:problem}, there are many representative setups of CL, such as CIL, DIL and TIL. Here we take CIL as a typical scenario for theoretical analysis, and leave the results of DIL and TIL to Appendix~\ref{Appendix_A_theory}.

\textcolor{darkgreen}{\textbf{CL from Scratch:}}
When training from scratch, the TAP performance $P(\boldsymbol{x} \in \mathcal{X}^{y}|\mathcal{D})$ is to predict across all classes without distinguishing tasks, where $\mathcal{D} = \{\mathcal{D}_1, ..., \mathcal{D}_{t}\}$, $y \in [|\bigcup_{i=1}^{t} \mathcal{Y}_i|]$ denotes the ground truth label of $\boldsymbol{x}$, and $\mathcal{X}^{y}$ denotes the domain of class $y$. The restricted definition of CIL further posits two assumptions~\cite{kim2022theoretical}: the domains of tasks are disjoint (i.e., $\mathcal{X}_i \cap \mathcal{X}_{i'} =\emptyset$, $\forall i \neq i'$), and the domains of classes of the same task are disjoint (i.e., $\mathcal{X}_{i,j} \cap \mathcal{X}_{i,j'} =\emptyset$, $\forall j \neq j'$). Through predicting which task to perform and then performing that task (i.e., there is an execution order), the CIL probability can be expressed as a hierarchical process of task-identity inference (TII) and within-task prediction (WTP):
\begin{small}
\begin{align}\label{BayesTheorem_Scratch}
\underbrace{P(\boldsymbol{x} \in \mathcal{X}_{i,j}|\mathcal{D})}_{\text{CIL}}  = 
   \underbrace{P(\boldsymbol{x} \in \mathcal{X}_{i}|\mathcal{D})}_{\text{TII}} 
   \underbrace{P(\boldsymbol{x} \in \mathcal{X}_{i,j}|\boldsymbol{x} \in \mathcal{X}_{i},\mathcal{D})}_{\text{WTP}}.
\end{align}
\end{small}
Eq.~(\ref{BayesTheorem_Scratch}) is exactly the main conclusion of a previous theoretical study of CL from scratch~\cite{kim2022theoretical}. Given the assumptions of disjoint domains and the omitted impact of randomly initialized $\theta$, the TAP performance $P(\boldsymbol{x} \in \mathcal{X}^{y}|\mathcal{D})$ is essentially equivalent to the decomposed CIL performance ${P}(\boldsymbol{x} \in \mathcal{X}_{\bar{i},\bar{j}}|\mathcal{D})$, where $\bar{i}\in [t]$ and $\bar{j} \in [ |\mathcal{Y}_{\bar{i}}| ]$ denote the ground truth of an $\boldsymbol{x}$ w.r.t. the task identity and within-task index. 
Here we provide a more intuitive explanation with the definition of class labels and the implementation of output heads. The decomposed CIL performance can be naturally computed in a multi-head manner with two steps: inferring the task identity $\bar{i} \in [t]$, i.e., the TII performance; and predicting the within-task index $\bar{j} \in [|\mathcal{Y}_{\bar{i}}|]$, i.e., the WTP performance. In comparison, the TAP performance $P(\boldsymbol{x} \in \mathcal{X}^{y}|\mathcal{D})$ is computed by performing single-head prediction for global class $y \in [|\bigcup_{i=1}^{t} \mathcal{Y}_i|]$. For CL from scratch, the decomposed CIL performance of multi-head inference \& prediction is equivalent to the TAP performance of single-head prediction. In this case, the decomposed CIL performance can also be computed in a single-head manner, as many traditional continual learning methods do.

\textcolor{darkgreen}{\textbf{CL with Pre-training:}}
When considering the impact of pre-trained knowledge carried by parameters $\theta$, the TAP is redefined as $P(\boldsymbol{x} \in \mathcal{X}^{y}|\mathcal{D},\theta)$, while the CIL probability of TII and WTP in Eq.~(\ref{BayesTheorem_Scratch}) is re-written as
\begin{small}
\begin{align}\label{BayesTheorem}
    \underbrace{P(\boldsymbol{x} \in \mathcal{X}_{i,j}|\mathcal{D},\theta)}_{\text{CIL}} \triangleq 
   \underbrace{P(\boldsymbol{x} \in \mathcal{X}_{i}|\mathcal{D},\theta)}_{\text{TII}}
   \underbrace{P(\boldsymbol{x} \in \mathcal{X}_{i,j}|\boldsymbol{x} \in \mathcal{X}_{i},\mathcal{D},\theta)}_{\text{WTP}}.
\vspace{-0.2cm}
\end{align}
\end{small}
It can be seen that both the TAP performance $P(\boldsymbol{x} \in \mathcal{X}^{y}|\mathcal{D},\theta)$ and the CIL performance ${P}(\boldsymbol{x} \in \mathcal{X}_{\bar{i},\bar{j}}|\mathcal{D},\theta)$ are now affected by $\theta$, but in different ways. The pre-trained parameters $\theta$ in TAP affect simultaneously all observed classes (i.e., a large output space $[|\bigcup_{i=1}^{t} \mathcal{Y}_i|]$), while in CIL affect separately TII and WTP (i.e., two small output spaces $[t]$ and $[|\mathcal{Y}_{\bar{i}}|]$). 
Accordingly, the CIL performance is upper bounded by either the TII performance $P(\boldsymbol{x} \in \mathcal{X}_{\bar{i}}|\mathcal{D},\theta)$ or the WTP performance $P(\boldsymbol{x} \in \mathcal{X}_{\bar{i},\bar{j}}|\boldsymbol{x} \in \mathcal{X}_{\bar{i}},\mathcal{D},\theta)$, whereas the TAP performance is not, as the CL tasks and pre-trained knowledge are not conditionally independent from a statistical perspective. For example, an incorrectly predicted task identity results in full errors in predicting within-task index, remaining a performance gap from rectifying the predictions of inter-task classes to make them properly balanced. 
This also underscores the notable difference of the multi-head inference \& prediction from the single-head prediction in the context of pre-training. 
Such difference tends to be more pronounced if the task structure is not clear (e.g., randomly split classes of the same dataset), and has been empirically validated in our preliminary experiments~\cite{wang2023hierarchical}, where the multi-head inference \& prediction significantly underperforms the single-head prediction in CL with PTMs.

Therefore, we propose to further optimize TAP along with the improved TII and WTP, formulating the ultimate goal of CL as a multi-objective optimization problem, i.e., 
\begin{small}
\begin{align}\label{FinalObjective}
\max [\, \underbrace{P(\boldsymbol{x} \in \mathcal{X}_{\bar{i},\bar{j}}|\mathcal{D},\theta)}_{\text{CIL}}, 
\underbrace{P(\boldsymbol{x} \in \mathcal{X}^{y}|\mathcal{D},\theta)}_{\text{TAP}} \,],
\end{align}
\end{small}
where $P(\boldsymbol{x} \in \mathcal{X}_{\bar{i},\bar{j}}|\mathcal{D},\theta)$ follows a similar decomposition as in Eq.~(\ref{BayesTheorem}), with TII and WTP having an execution order.

To resolve the above WTP, TII and TAP, we derive the sufficient and necessary conditions with the widely-used cross-entropy loss. 
Specifically, we define 
\begin{small}
\begin{align}\label{H_WTP}
   {H}_{\rm{WTP}}(\boldsymbol{x}) &= \mathcal{H}(\boldsymbol{1}_{\bar{j}},\{P(\boldsymbol{x} \in \mathcal{X}_{\bar{i},j}|\boldsymbol{x} \in \mathcal{X}_{\bar{i}},\mathcal{D},\theta)\}_j),\\
    {H}_{\rm{TII}}(\boldsymbol{x}) &= \mathcal{H}(\boldsymbol{1}_{\bar{i}},\{P(\boldsymbol{x} \in \mathcal{X}_{i}|\mathcal{D},\theta)\}_{i}),\\
   {H}_{\rm{TAP}}(\boldsymbol{x}) &= \mathcal{H}(\boldsymbol{1}_{y}, \{P(\boldsymbol{x} \in \mathcal{X}^{c}|\mathcal{D},\theta)\}_{c} ),
\end{align}
\end{small}
where ${H}_{\rm{WTP}}$, ${H}_{\rm{TII}}$ and ${H}_{\rm{TAP}}$ are the cross-entropy values of WTP, TII and TAP, respectively.
$c \in [|\bigcup_{i=1}^{t} \mathcal{Y}_i|]$ denotes the index of all observed classes. 
$\mathcal{H}(p,q) \triangleq -\mathbb{E}_{p}[\log q]=-\sum_{k}p_k \log q_k$. $\boldsymbol{1}_{\cdot}$ is a one-hot encoding function.

We now present the first theorem under the CIL setting (see Appendix~\ref{Appendix_A_theory} for the complete proof and the corresponding extensions to DIL and TIL settings):
\begin{theorem}
    \label{LossError1}
    For continual learning (CL) in a pre-training context, 
    if $\mathbb{E}_{\boldsymbol{x}} [{H}_{\rm{WTP}}(\boldsymbol{x})] \leq \delta$, $\mathbb{E}_{\boldsymbol{x}} [{H}_{\rm{TII}}(\boldsymbol{x})] \leq \epsilon$, and $\mathbb{E}_{\boldsymbol{x}} [{H}_{\rm{TAP}}(\boldsymbol{x})] \leq \eta$, we have the loss error $\mathcal{L} \in [0, \max\{{\delta +\epsilon},\eta\}]$, regardless whether the WTP predictor, TII predictor and TAP predictor are trained together or separately.
\end{theorem}
With the use of cross-entropy, the CL performance tends to be better as the bounds are tightened.
In Theorem~\ref{LossError1} we have shown that good performance of WTP, TII and TAP is sufficient to ensure good performance of CL. 
For completeness, we now study the necessary conditions of a well-performed CL model in Theorem~\ref{LossError2}.
\begin{theorem}
    \label{LossError2}
     For CL in a pre-training context, if the loss error $\mathcal{L}\leq \xi $, then there always exist (1) a WTP predictor, s.t. ${H}_{\rm{WTP}} \leq \xi$; (2) a TII predictor, s.t. ${H}_{\rm{TII}} \leq \xi$; and (3) a TAP predictor, s.t. ${H}_{\rm{TAP}} \leq \xi$.
\end{theorem}
Theorem~\ref{LossError2} suggests that if a CL model is well trained (i.e., with low loss error), then the WTP error, TII error and TAP error for sequentially arriving tasks are always implied to be small. Similar to the connection between TAP and CIL under different assumptions, Theorem~\ref{LossError1} and Theorem~\ref{LossError2} would degenerate into the main conclusion of the previous theoretical study~\cite{kim2022theoretical} if the pre-trained knowledge carried by $\theta$ is not considered. This suggests that the presented theorems are particularly directed to the impact of pre-training on CL.

\subsection{Connection of TII to OOD Detection}\label{sec:tii_ood}

In essence, the TII probability specifies the CL problem with task-wise input samples. Although the definition of ``task'' in literature can generalize to an incoming training batch of distinct distribution~\cite{wang2023comprehensive}, it may not be pertinent in describing realistic data streams with apparent similarity and dissimilarity.
Indeed, the CL problem is often associated with the out-of-distribution (OOD) detection~\cite{yang2021generalized}, i.e., the ability of a model to detect an unseen input sample, which has been shown to behave similarly as task prediction when training from scratch~\cite{kim2022theoretical}. 
Inspired by this, we further explore the necessary conditions to optimize TII/OOD for CL in a pre-training context, in order to facilitate adaptive knowledge accumulation from more pronounced distribution changes.

We again use cross-entropy to measure the performance of TII and OOD detection, so as to establish their connection in each task. We first define the OOD detector of the $i$-th task as $P_i(\boldsymbol{x} \in \mathcal{X}_{i}|\mathcal{D},\theta)$. 
Different from the TII probability, the OOD detection probability here is a Bernoulli distribution: 
\begin{small}
\begin{equation}
H_{{\rm{OOD}},i}(\boldsymbol{x}) =
\begin{cases}
\mathcal{H}(1, P_i(\boldsymbol{x} \in \mathcal{X}_{i}|\mathcal{D},\theta)), & \boldsymbol{x} \in \mathcal{X}_{i}\\ 
\mathcal{H}(0, P_i(\boldsymbol{x} \in \mathcal{X}_{i}|\mathcal{D},\theta)), & \boldsymbol{x} \notin \mathcal{X}_{i} \\
\end{cases},\\
\end{equation}
\end{small}
where $H_{{\rm{OOD}},i}$ is the cross-entropy value of an OOD detector of task $i$, and $P_i(\boldsymbol{x} \in \mathcal{X}_{i}|\mathcal{D},\theta)$ can be predicted with an appropriate distance function. The TII probability can then be defined with the OOD detectors, i.e., $P(\boldsymbol{x} \in \mathcal{X}_{i}|\mathcal{D},\theta)=\frac{P_i(\boldsymbol{x} \in \mathcal{X}_{i}|\mathcal{D},\theta)}{\sum_{j}P_j(\boldsymbol{x} \in \mathcal{X}_{j}|\mathcal{D},\theta)}$.

Now we have the following theorem to explore the connection between TII and OOD detection in a pre-training context (see Appendix~\ref{LossOODProof} for the complete proof):
\begin{theorem}
    \label{LossOOD1}
    For CL in a pre-training context, 
     if $H_{{\rm{OOD}},i}(\boldsymbol{x}) \leq \epsilon_i$ for $i \in [t]$, 
     then we have $H_{\rm{TII}}(\boldsymbol{x})\leq (\sum_i \boldsymbol{1}_{\boldsymbol{x}\in \mathcal{X}_{i}}e^{\epsilon_i})(\sum_i \boldsymbol{1}_{\boldsymbol{x}\notin \mathcal{X}_{i}}(1-e^{-\epsilon_i}))$. 
     Likewise, if $H_{\rm{TII}}(\boldsymbol{x})\leq \epsilon$, 
     then ${H}_{{\rm{OOD}},i}(\boldsymbol{x}) \leq \epsilon$ for $i \in [t]$. 
\end{theorem}
Theorem~\ref{LossOOD1} shows that the TII performance improves if the OOD detection performance improves, and vice versa. In particular, $H_{\rm{TII}}(\boldsymbol{x})$ converges to 0 as $\epsilon_i$ converges to 0. Note that the connection between TII and OOD detection in Theorem~\ref{LossOOD1} for CL with pre-training is similar in form to that for  CL from scratch \cite{kim2022theoretical}.
We further derive the sufficient and necessary conditions of improving CL with WTP, OOD detection and TAP, as detailed in Appendix~\ref{LossOODProof}.

\section{Hierarchical Decomposition PET}\label{sec:method} 
\begin{figure}[t]
    \centering
    \includegraphics[width=0.50\textwidth]{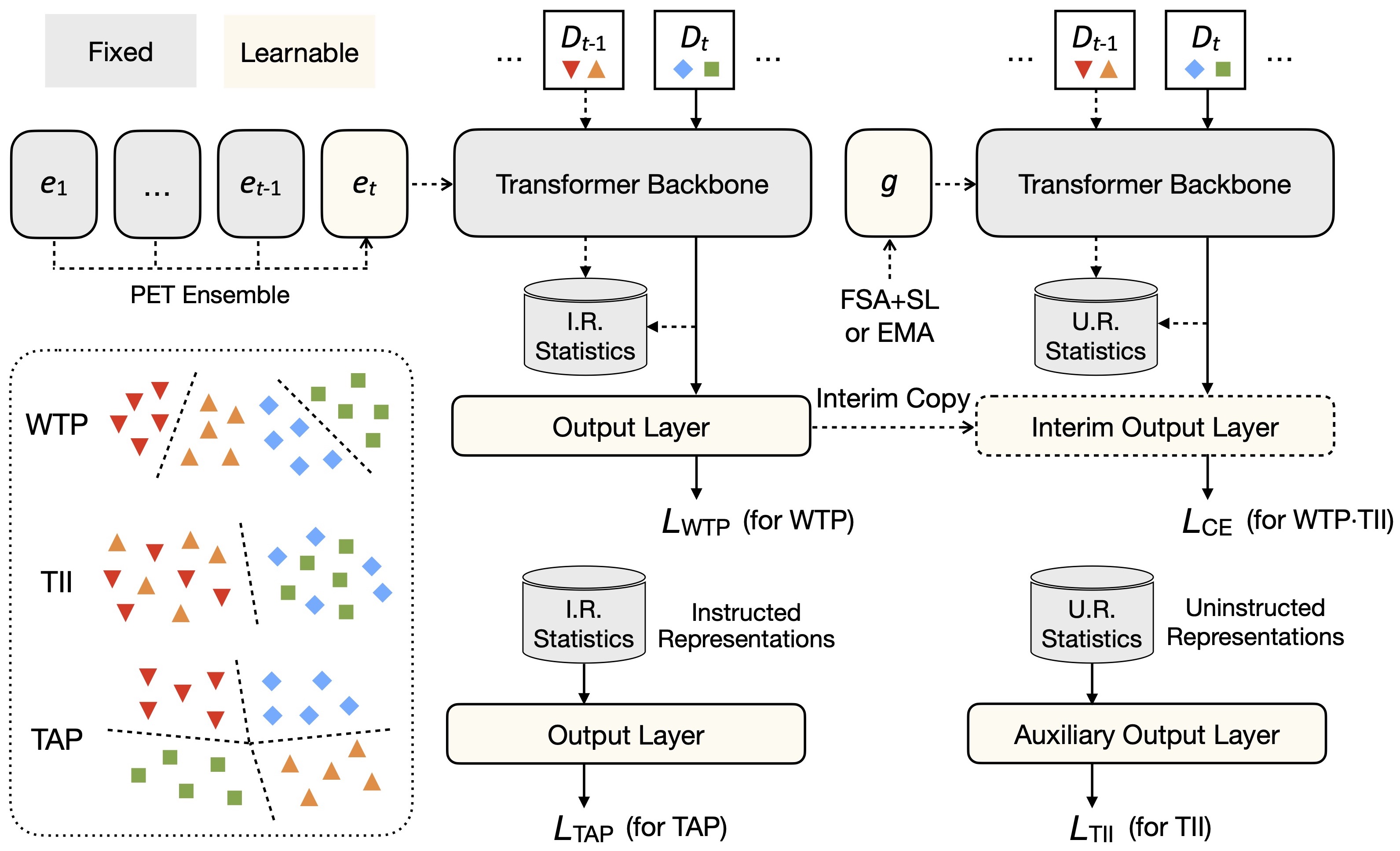}
    \vspace{-0.3cm}
\caption{Illustration of HiDe-PET. See Fig.~\ref{PET_Techniques} for detailed implementations of PET techniques and the frozen transformer backbone.
\textcolor{darkgreen}{Here we have an example of 2 tasks with 2 classes each. 
WTP aims to classify each 2 classes well, optimized by the PET ensemble of task-specific parameters. TII aims to select appropriately one of the 2 tasks, optimized by the task-shared parameters and the recovery of uninstructed representations. 
TAP aims to classify the total of 4 classes well, optimized by the recovery of instructed representations upon WTP and TII.
}
}
\vspace{-0.3cm}
\label{HiDe_Method}
\end{figure}

Based on the above analysis, we now present an innovative approach named Hierarchical Decomposition PET (HiDe-PET) to explicitly optimize the three hierarchical components tailored for CL in a pre-training context (see Fig.~\ref{HiDe_Method}). 
Our HiDe-PET is applicable to mainstream PET techniques for learning sequentially arriving tasks, from which the pre-trained knowledge can also be evolved.

\begin{algorithm}[tb]
    \caption{Training Algorithm of HiDe-PET}
    \label{alg:algorithm}
    \textbf{Input}: Pre-trained transformer backbone $f_{\theta}$, training sets $\mathcal{D}_1, ..., \mathcal{D}_t$, number of tasks $t$, number of epochs $E$ and $E'$. 
    \\ 
    \textbf{Output}: Parameters $\boldsymbol{e}_{1}, ..., \boldsymbol{e}_{t}$, $\boldsymbol{g}$, $\omega$ and $\psi$
    \begin{algorithmic}[1] 
          \State Initialize $\boldsymbol{g}$, $\omega$ and $\psi$
          \For{$i = 1, ..., t$}
              \State Initialize $\hat{\psi}$ with $\psi$
              \State Construct $\boldsymbol{e}_{i}$ with $\boldsymbol{e}_{1},...,\boldsymbol{e}_{i-1}$
              \For{$epoch = 1, ..., E$} 
              \State Update $\boldsymbol{g}$ and $\hat{\psi}$ with $\mathcal{L}_{{\rm{CE}}}(\hat{\psi}, \boldsymbol{g})$ 
              \State Update $\boldsymbol{e}_{i}$ and $\psi$ with $\mathcal{L}_{{\rm{WTP}}}(\psi, \boldsymbol{e}_{i})$ in Eq.~(\ref{eq.wtp_loss}) 
              \EndFor

              \For{$c \in \mathcal{Y}_i$} 
              \State Calculate $\hat{\mathcal{G}}_{i,c}$ from $f_{\theta}$, $\boldsymbol{g}$ and $\mathcal{D}_i$
              \State Calculate $\mathcal{G}_{i,c}$ from $f_{\theta}$, $\boldsymbol{e}_{i}$ and $\mathcal{D}_i$
              \EndFor

              \For{$epoch = 1, ..., E'$} 
              \State Optimize $\omega$ with $\mathcal{L}_{{\rm{TII}}}(\omega)$ in Eq.~(\ref{eq.tii_loss}) 
               \State Optimize $\psi$ with $\mathcal{L}_{{\rm{TAP}}}(\psi)$ in Eq.~(\ref{eq.tap_loss})
              \EndFor
                   
        \EndFor
       \State \textbf{return} $(\boldsymbol{e}_{1}, ..., \boldsymbol{e}_{t}$, $\boldsymbol{g}$, $\omega$, $\psi)$
    \end{algorithmic}
\end{algorithm}

\subsection{Optimization of Decomposed Objective}\label{sec:three_component}
The principal concept behind HiDe-PET stems from two unique strengths of PTMs in CL: 
(1) the pre-trained representations can be effectively adapted to the distribution of target tasks through implementing PET techniques, and (2) the distribution of target tasks can be efficiently recovered through preserving statistical information of their pre-trained representations. The optimization of WTP, TII and TAP is described as below.

First, we improve \textbf{WTP} through incorporating task-specific knowledge from $\mathcal{D}_{i}$ to capture the distribution of any task $i \in [t]$. Specifically, we employ multiple sets of task-specific parameters $\boldsymbol{e}_{1},...,\boldsymbol{e}_{t}$ to instruct pre-trained representations, implemented via mainstream PET techniques. 
Their concrete forms can be defined as the prompt parameters $\{\boldsymbol{p}\}$ for ProT and PreT, the projection matrices $\{\boldsymbol{W}_{{\rm{down}}}, \boldsymbol{W}_{{\rm{up}}}\}$ for Adapter and LoRA, etc.
When learning task $t$, we keep the previous parameters $\boldsymbol{e}_{1},...,\boldsymbol{e}_{t-1}$ frozen to avoid catastrophic forgetting. 
In order to inherit knowledge obtained from CL, we employ a PET ensemble strategy that initializes $\boldsymbol{e}_{t}$ with $\boldsymbol{e}_{t-1}$ and optimizes $\boldsymbol{e}_{t}$ with a weighted combination of all previous parameters $\boldsymbol{e}_{t} \leftarrow \alpha \sum_{i \in [t-1]} \boldsymbol{e}_{i} + (1-\alpha) \boldsymbol{e}_{t}$, where $\alpha\in[0,1]$ is a hyperparameter that controls the strength of obtained knowledge that facilitates $\boldsymbol{e}_{t}$ in learning task $t$. 
For ${H}_{\rm{WTP}}$, we then optimize $\boldsymbol{e}_{t}$ and $\psi$ via the cross-entropy (CE) loss:
\begin{small}
\begin{equation}
\begin{split}
&\mathcal{L}_{{\rm{WTP}}}(\psi, \boldsymbol{e}_{t}) 
= \mathcal{L}_{{\rm{CE}}}(\psi, \boldsymbol{e}_{t}) \\
&= \frac{1}{|\mathcal{D}_{t}|} \sum_{(\boldsymbol{x},y) \in \mathcal{D}_{t}} - \log \frac{\exp(h_{\psi}(f_{\theta, \boldsymbol{e}_{t}}(\boldsymbol{x})[y]))}{\sum_{y' \in \mathcal{Y}_t} \exp(h_{\psi}(f_{\theta, \boldsymbol{e}_{t}}(\boldsymbol{x}))[y'])}.
\end{split}
\label{eq.wtp_loss}
\end{equation}
\end{small}

Second, we improve \textbf{TII} and \textbf{TAP} through recovering the distributions of pre-trained representations to achieve the optimum over all tasks. 
Specifically, after learning each task $i \in [t]$, we collect the uninstructed and instructed representations, i.e., the backbone projection of $\mathcal{D}_{i}$ with $f_{\theta}(\cdot)$ and $f_{\theta, \boldsymbol{e}_{i}}(\cdot)$, respectively. We then approximate the distributions of these representations with their statistical information for subsequent recovery.
Taking classification tasks as an example, for each class $c \in \mathcal{Y}_i$ and $i\in [t]$, we denote the approximated distributions of uninstructed and instructed representations as $\hat{\mathcal{G}}_{i,c}$ and $\mathcal{G}_{i,c}$, respectively. 

For ${H}_{\rm{TII}}$, we build an auxiliary output layer $\hat{h}_{\omega}(\cdot): \mathbb{R}^d \rightarrow \mathbb{R}^t$ with parameters $\omega$, learning the projection from uninstructed representations to the task identity: 
\begin{small}
\begin{equation}
\mathcal{L}_{{\rm{TII}}}(\omega) = \sum_{i \in [t]} \sum_{c \in \mathcal{Y}_i} \sum_{\hat{\boldsymbol{r}} \in \hat{\mathcal{S}}_{i,c}} - \log \frac{\exp(\hat{h}_{\omega}(\hat{\boldsymbol{r}})[i])}{\sum_{j \in [t]} \exp(\hat{h}_{\omega}(\hat{\boldsymbol{r}})[j])},
\label{eq.tii_loss}
\end{equation}
\end{small}
where $\hat{\mathcal{S}}_{i,c}$ is a collection of uninstructed representations sampled in a class-balanced manner from $\hat{\mathcal{G}}_{i,c}$ for $c \in \mathcal{Y}_i$ and $i \in [t]$.
Therefore, we can determine the task identity via uninstructed representations and then obtain the corresponding instructed representations. 

For ${H}_{\rm{TAP}}$, we use a similar strategy to optimize the final output layer $h_\psi(\cdot): \mathbb{R}^d \rightarrow \mathbb{R}^{|\mathcal{Y}|}$ with parameters $\psi$, learning the projection from instructed representations to all observed classes: 
\begin{small}
\begin{equation}
\mathcal{L}_{{\rm{TAP}}}(\psi) = \sum_{i \in [t]} 
\sum_{c \in \mathcal{Y}_i} \sum_{\boldsymbol{r} \in \mathcal{S}_{i,c}} - \log \frac{\exp(h_{\psi}(\boldsymbol{r})[c])}{\sum_{j \in [t]}\sum_{c' \in \mathcal{Y}_j} \exp(h_{\psi}(\boldsymbol{r})[c'])}, 
\label{eq.tap_loss}
\end{equation}
\end{small}
where $\mathcal{S}_{i,c}$ is a collection of instructed representations sampled in a class-balanced manner from $\mathcal{G}_{i,c}$ for $c \in \mathcal{Y}_i$ and $i \in [t]$.

\textbf{Improvement of Uninstructed Representations:} Since the TII process depends heavily on the uninstructed representations of $\mathcal{D}_{i}$ collected from $f_{\theta}(\cdot)$, its effectiveness tends to be severely affected by the pre-trained checkpoints and target tasks. 
This issue becomes particularly pronounced when implementing more advanced PET techniques, which better incorporate task-specific knowledge for WTP and are thus more sensitive to TII.
To address this issue, we further deploy a set of task-shared parameters $\boldsymbol{g}$ to improve TII in a task-agnostic manner. $\boldsymbol{g}$ is implemented via mainstream PET techniques analogous to $\boldsymbol{e}_{i}$, while optimized with the cross-entropy $\mathcal{L}_{{\rm{CE}}}(\hat{\psi}, \boldsymbol{g})$ to accumulate task-shared knowledge from sequentially arriving $\mathcal{D}_{i}$ for $i \in [t]$,
where $\hat{\psi}$ is an interim copy of $\psi$ to avoid overwriting. 
We then use $f_{\theta, \boldsymbol{g}}(\cdot)$ instead of $f_{\theta}(\cdot)$ to collect uninstructed representations\footnote{For naming consistency, we still use ``uninstructed representations'' to denote the projection of $f_{\theta, \boldsymbol{g}}(\cdot)$.}, approximate each $\hat{\mathcal{G}}_{i,c}$ and optimize Eq.~\eqref{eq.tii_loss}.

Notably, $\boldsymbol{g}$ needs to overcome its own catastrophic forgetting that leads to not only the loss of information in representation learning but also the representation shift in subsequent recovery. There are many CL methods attempting to address this challenge~\cite{gao2023unified,mcdonnell2023ranpac,zhang2023slca}, but their strategies remain sub-optimal in balancing sequentially arriving tasks (see Table~\ref{table:shared_pet}). 
Here we propose a simple yet effective strategy by taking advantages of first-session adaptation (FSA) \cite{mcdonnell2023ranpac,panos2023first} and slow learner (SL) \cite{zhang2023slca,gao2023unified}. Specifically, we learn $\boldsymbol{g}$ in the first task with a larger learning rate that is adequately strong for representation learning, and then in subsequent tasks with a smaller learning rate for further fine-tuning. In this way, task-shared knowledge is effectively incorporated into $\boldsymbol{g}$ and accumulates over time.

\begin{figure}[ht]
    \centering
    \includegraphics[width=0.48\textwidth]{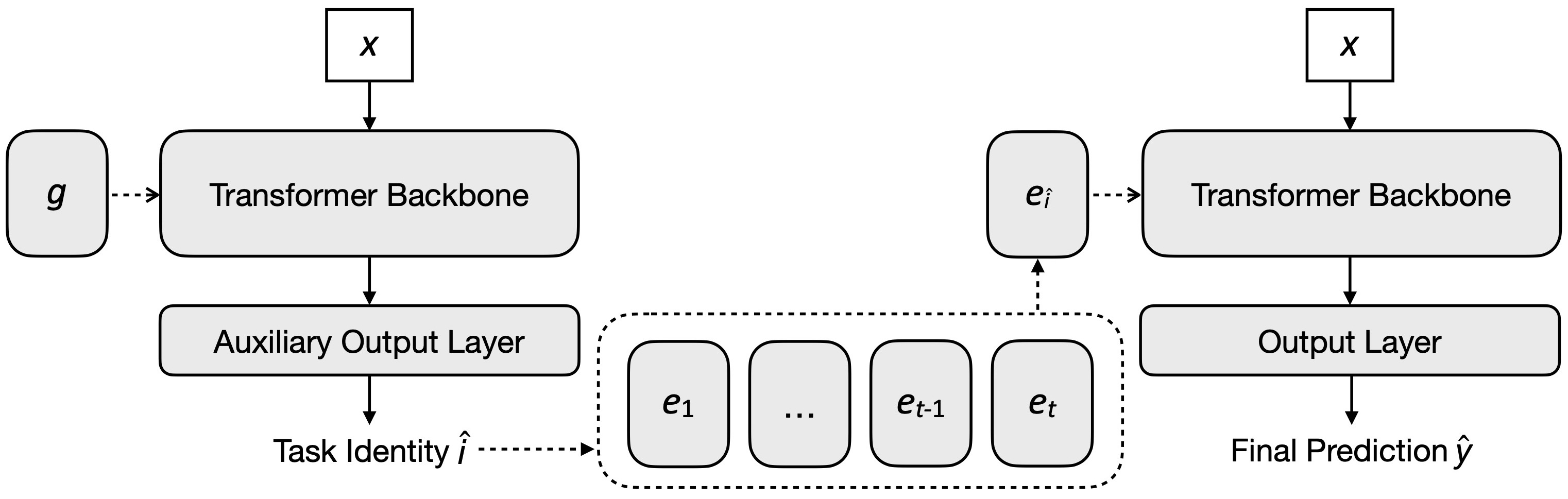}
    \vspace{-0.2cm}
\caption{Inference of HiDe-PET during the testing phase.
\textcolor{darkgreen}{HiDe-PET first employs the task-shared parameters and the auxiliary output layer to infer task identity, and then employs the corresponding task-specific parameters and the output layer to obtain final prediction.
}
}
\vspace{-0.1cm}
\label{HiDe_Inference}
\end{figure}

\textbf{Recovery of Task Distributions:}
Since the pre-trained representations are well-distributed in general, there are many feasible strategies to approximate and preserve their distributions $\hat{\mathcal{G}}_{i,c}$ and $\mathcal{G}_{i,c}$. The most straightforward option is to save randomly selected prototypes \cite{tran2023koppa}, yet not adequately employing the relationships between them. For classification tasks, the class-wise distribution is typically single-peaked and thus can be naturally approximated as a Gaussian with its mean and covariance \cite{zhang2023slca,wang2023hierarchical}. In order to reduce storage complexity, dedicated covariance matrices need to be further simplified for practical use, suffering from information loss to varying degrees \cite{zhang2023slca,mcdonnell2023ranpac,wang2023hierarchical}. 
Considering both storage efficiency and task-type generality, our default implementation is to obtain multiple representation centroids with $k$-Nearest Neighbors ($k$NNs) and add Gaussian noise to them. We also provide an empirical comparison of different implementations in Table~\ref{table:representation}.

Overall, the entire HiDe-PET consists of two training stages (see Algorithm~\ref{alg:algorithm} and Fig.~\ref{HiDe_Method}), corresponding to the pre-trained transformer backbone and the (auxiliary) output layer.
At test time, HiDe-PET first predicts the task identity $\hat{i} = \hat{h}_{\omega}(f_{\theta,\boldsymbol{g}}(\boldsymbol{x}))$ and then the overall class label $\hat{y} = h_\psi(f_{\theta,\boldsymbol{e}_{\hat{i}}}(\boldsymbol{x}))$ (see Fig.~\ref{HiDe_Inference}). 
Compared to the backbone parameters $\theta$, the trainable parameters $\boldsymbol{e}_t$, $\boldsymbol{g}$, $\omega$ and $\psi$, as well as the representation recovery are all lightweight, thus ensuring resource efficiency.

\subsection{Adaptive Knowledge Accumulation}\label{sec:ood_method}

\begin{figure}[th]
    \centering
    \vspace{-0.2cm}
    \includegraphics[width=0.48\textwidth]{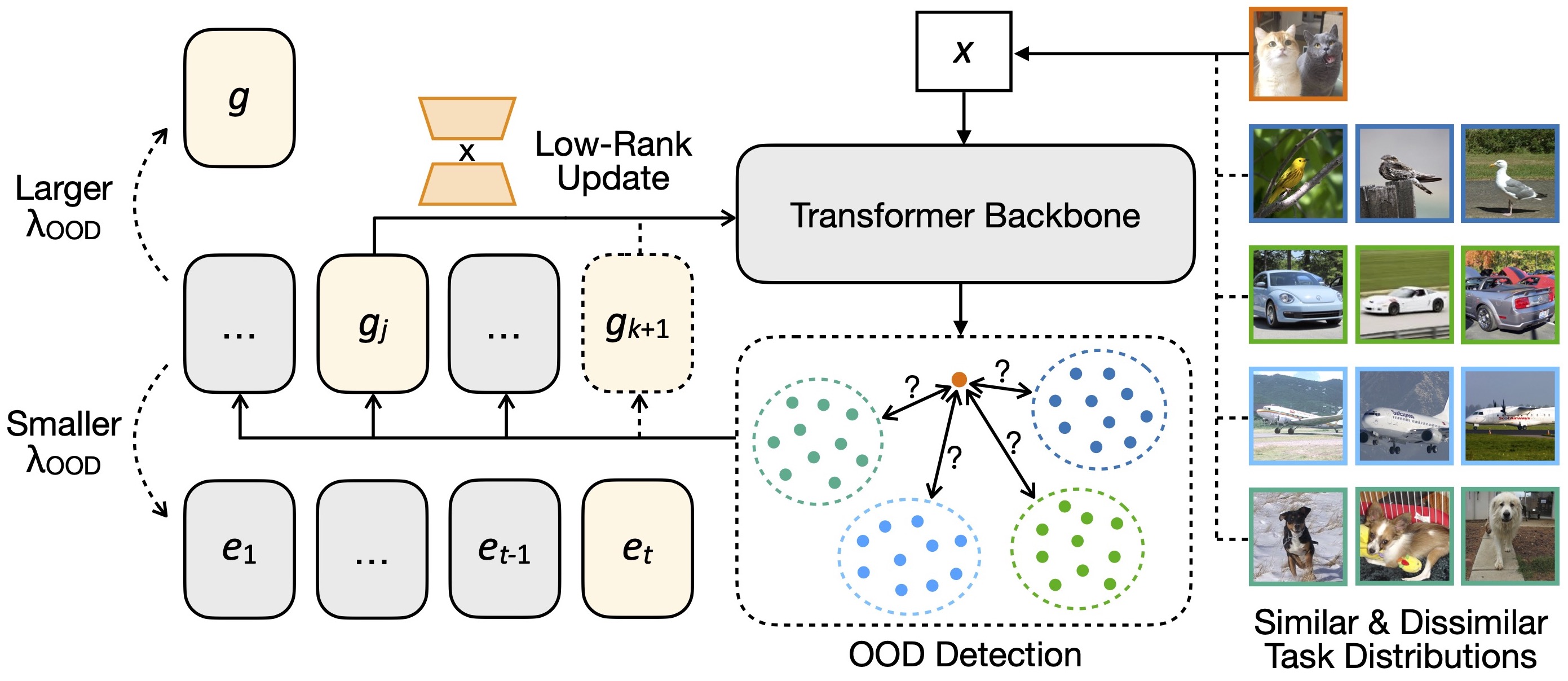}
    \vspace{-0.2cm}
\caption{Adaptive knowledge accumulation.
\textcolor{darkgreen}{HiDe-PET employs OOD detection to decide whether to expand a new set of parameters or to retrieve a previously learned set of parameters. Such parameters are further specified into LoRA-based PET to update the backbone.
}
}
\vspace{-0.1cm}
\label{HiDe_AKA}
\end{figure}

Within HiDe-PET, the parallel organization of $\boldsymbol{e}_1, ..., \boldsymbol{e}_t$ and $\boldsymbol{g}$ facilitates the incorporation of task-specific and task-shared knowledge for many representative CL scenarios.
In fact, the functions of $\boldsymbol{e}_1, ..., \boldsymbol{e}_t$ and $\boldsymbol{g}$ are usually not exclusive, depending on the similarity and dissimilarity between task distributions.
Motivated by the intrinsic connection between TII and OOD detection in our theoretical analysis, we unify the task-specific and task-shared PET architectures with a hierarchy of expandable parameter sets (see Fig.~\ref{HiDe_AKA}), which may degenerate into either case of Sec.~\ref{sec:three_component}. We further explore a particular implementation of this hierarchy in order to accumulate knowledge adaptively from more pronounced distribution changes. 

Let us assume the existence of multiple parameter sets that are implemented via mainstream PET techniques and are expanded or retrieved upon the input samples. 
For example, the sample elements of sequentially learning tasks $i \in [t-1]$ have derived $k$ parameter sets $ \boldsymbol{g}_1,...,\boldsymbol{g}_{k} $.
If the incoming $\boldsymbol{x} \in \mathcal{X}_{t}$ is identified as OOD from the previously observed distributions of $\mathcal{X}_{i}$, it learns an expanded set of parameters $\boldsymbol{g}_{k+1}$ through the task-specific loss $\mathcal{L}_{{\rm{CE}}}(\hat{\psi}, \boldsymbol{g}_{k+1})$, otherwise it retrieves and updates the most relevant one $\boldsymbol{g}_{j}$ with $j \in [k]$ through $\mathcal{L}_{{\rm{CE}}}(\hat{\psi}, \boldsymbol{g}_{j})$, where $\hat{\psi}$ is an interim copy of the output layer parameters $\psi$ to avoid overwriting.

\textbf{OOD Detection Strategy:}
Given that the previous $\mathcal{X}_{i}$ are not available in CL to describe their distributions, we take inspirations from recent metric-based OOD detectors \cite{sun2022out} and formulate an effective criterion with their uninstructed representations:
\begin{small}
\begin{equation}
P_i(\boldsymbol{x} \in \mathcal{X}_{i}|\mathcal{D},\theta) = 
\boldsymbol{1} ({\rm{Dis}}(\boldsymbol{x}, \hat{\mathcal{G}}_{i}) > \lambda_{{\rm{OOD}}}), \boldsymbol{x} \in \mathcal{X}_{t},
\label{eq.ood_detection}
\end{equation}
\end{small}
where $\hat{\mathcal{G}}_{i}$ for $i \in [t-1]$ denotes the approximated distribution of uninstructed representations. It can be further specified as $\hat{\mathcal{G}}_{i} = \bigcup_{c \in \mathcal{Y}_i} \hat{\mathcal{G}}_{i, c}$ for classification tasks. $\lambda_{{\rm{OOD}}}$ denotes the OOD detection threshold. $\boldsymbol{1}(\cdot)$ is the indicator function. ${\rm{Dis}}(\boldsymbol{x}, \hat{\mathcal{G}}_{i})$ measures the distance of task-wise distributions, which can be implemented via the average Euclidean distance between $f_{\theta,\boldsymbol{g}}(\boldsymbol{x})$ and $\hat{\boldsymbol{r}} \sim \hat{\mathcal{G}}_{i}$. 
Consequently, if $\boldsymbol{x}$ is identified as OOD for all tasks $i\in[t-1]$, then it will be associated with $\boldsymbol{g}_{k+1}$. Otherwise, it will retrieve the associated $\boldsymbol{g}_{j}$ for $j \in [k]$ corresponding to the majority of the most relevant task $\hat{i} = \arg \min_{i \in [t-1]} {\rm{Dis}}(\boldsymbol{x}, \hat{\mathcal{G}}_{i})$. To overcome catastrophic forgetting when updating $\boldsymbol{g}_{j}$, we employ the same strategy as for learning the task-shared parameters $\boldsymbol{g}$, e.g., a combination of FSA \cite{panos2023first} and SL \cite{zhang2023slca}. As a special case, we have $k = 1$ if all input samples are identified as in-distribution, for which only $\boldsymbol{g}_1$ exists and is equivalent to $\boldsymbol{g}$.

\textbf{Connection of PET Architectures:}
We now extend the above discussion with the criterion of OOD detection in Eq.~\eqref{eq.ood_detection}. Given a task sequence $1,...,t$, using a larger $\lambda_{{\rm{OOD}}}$ would make $k \rightarrow 1$ and $\{\boldsymbol{g}_{1},...,\boldsymbol{g}_{k}\} \rightarrow \{\boldsymbol{g}\}$, while using a smaller $\lambda_{{\rm{OOD}}}$ would make $k \rightarrow t$ and $\{\boldsymbol{g}_{1},...,\boldsymbol{g}_{k}\} \rightarrow \{\boldsymbol{e}_{1},...,\boldsymbol{e}_{t}\}$. 
Therefore, the representation learning of HiDe-PET in Sec.~\ref{sec:three_component} is equivalent to a parallel combination of these two extreme conditions for TII and WTP, respectively. This is a reasonable choice as most CL benchmarks employ randomly split classes of the same dataset as the task sequence, i.e., there is no actual task structure.

Instead, Eq.~\eqref{eq.ood_detection} applies to the apparent similarity and dissimilarity between task distributions, which is more realistic in applications and enables adaptive knowledge accumulation from CL for enhanced utilization.
Here we leverage the unique property of LoRA-based PET to construct a specialized implementation, serving as a plug-in module for Algorithm~\ref{alg:algorithm}. 
Unlike the commonly-used Prompt-based PET that updates only attached tokens, the LoRA-based PET specifies $\boldsymbol{g}_{1},...,\boldsymbol{g}_{k}$ as the approximated updates of $\theta$, where the most relevant $\boldsymbol{g}_{j}$ is selected and temporarily added to $\theta$ in CL. Therefore, the learning of subsequent tasks can be significantly improved from the accumulated knowledge and further contribute to it (see Fig.~\ref{HiDe_AKA}).
Moreover, this allows for the flexible evolution of pre-trained knowledge with target tasks in a lifelong manner, deviating from the conventional practice of fixing it at the initial checkpoint.

In brief, Sec.~\ref{sec:theory} and~\ref{sec:method} serve as a unified framework for CL with PTMs and PET, applicable to explore the distinct impacts of implementation strategy, PET technique and PET architecture, as well as the adaptive knowledge accumulation for enhanced utilization.

\section{Experiment}\label{sec:experiment}
In this section, we perform extensive experiments to demonstrate the effectiveness and generality of our HiDe-PET. 
We first describe the experimental setups, and then present the experimental results with a comprehensive analysis.

\subsection{Experimental Setup}
To ensure the breadth and adequacy of the experiments, we consider a variety of CL benchmarks, pre-trained checkpoints, recent strong baselines, PET techniques and evaluation metrics. For the sake of comparison fairness, we follow the official implementations to reproduce all baselines.

\newcommand{\tabincell}[2]{\begin{tabular}{@{}#1@{}}#2\end{tabular}}
\begin{table*}[t]
\centering
    \vspace{-0.2cm}
    \caption{Overall performance of continual learning. PTM: pre-trained model. FAA (\%): final average accuracy. CAA (\%): cumulative average accuracy. 
    } 
      \vspace{-0.2cm}
	\smallskip
      \renewcommand\arraystretch{1.15}
     \addtolength{\tabcolsep}{-2pt}
	\resizebox{0.86\textwidth}{!}{ 
	\begin{tabular}{c|l|cc|cc|cc|cc}
	 \hline
        \multirow{2}{*}{PTM} & \multirow{2}{*}{\,\,\,\,\,\,\,\,\,\,\,\, Method} & \multicolumn{2}{c|}{Split CIFAR-100} & \multicolumn{2}{c|}{Split ImageNet-R} & \multicolumn{2}{c|}{Split CUB-200} & \multicolumn{2}{c}{Split Cars-196} \\
        & & FAA ($\uparrow$) & CAA ($\uparrow$) & FAA ($\uparrow$) & CAA ($\uparrow$) & FAA ($\uparrow$) & CAA ($\uparrow$) & FAA ($\uparrow$) & CAA ($\uparrow$) \\
        \hline
        \multirow{10}*{\tabincell{c}{Sup-21/1K}} 
       &L2P \cite{wang2022learning_l2p} &84.25 &88.84 &71.34 &76.87 &70.90 &76.70 &41.06 &46.47 \\ 
       &DualPrompt \cite{wang2022dualprompt} &83.75 &89.11 &71.65 &77.51 &68.21 &75.15 &42.68 &51.60 \\ 
       &S-Prompt++ \cite{wang2022sprompts} &82.41 &87.68 &71.15 &77.16 &68.01 &75.04 &39.62 &47.85 \\ 
       &CODA-Prompt \cite{smith2023coda} &86.65 &90.78 &75.11 &81.45 &71.43 &78.61 &45.67 &53.28 \\ 
       &LAE-PreT \cite{gao2023unified} &87.36 &91.63 &74.95 &81.23 &78.46 &83.65 &42.80 &52.12 \\ 
       &LAE-LoRA \cite{gao2023unified} &88.38 &92.45 &76.27 &82.99 &80.02 &84.47 &50.90 &58.38 \\ 
       &LAE-Adapter \cite{gao2023unified} &88.37 &92.50 &75.69 &82.80 &80.52 &84.75 &55.20 &61.63 \\ 
       \cdashline{2-10}[2pt/2pt]
       &HiDe-PreT &91.11 &94.11 &78.93 &83.44 &87.95 &88.48 &68.73 &69.19 \\ 
       &HiDe-LoRA &91.21 &93.99 &\textbf{79.32} &\textbf{83.97} &\textbf{88.76} &\textbf{89.32} &69.65 &69.36 \\ 
       &HiDe-Adapter &\textbf{91.23} &\textbf{94.26} &78.65 &83.55 &88.49 &89.17 &\textbf{70.98} &\textbf{71.31} \\ 
        \hline
       \multirow{10}*{\tabincell{c}{iBOT-21K}} 
       &L2P \cite{wang2022learning_l2p} &79.32 &85.13 &61.31 &70.05 &45.93 &56.02 &45.25 &45.75 \\ 
       &DualPrompt \cite{wang2022dualprompt} &78.17 &85.15 &61.42 &70.06 &41.46 &54.57 &34.61 &42.28 \\ 
       &S-Prompt++ \cite{wang2022sprompts}  &79.85 &85.89 &60.84 &69.01 &39.88 &53.71 &36.46 &43.34 \\ 
       &CODA-Prompt \cite{smith2023coda} &81.58 &87.36 &67.15 &76.54 &47.79 &59.24 &39.50 &43.32 \\ 
       &LAE-PreT \cite{gao2023unified} &82.22 &88.05 &65.85 &75.34 &45.83 &60.31  &49.14 &52.59 \\ 
       &LAE-LoRA \cite{gao2023unified} &84.63 &90.24 &70.49 &79.06 & 56.16 &68.38  &58.66 &62.59 \\ 
       &LAE-Adapter \cite{gao2023unified} &84.68 &90.31 &69.93 &79.14 &58.04 &70.01 &61.76 &65.61 \\ 
       \cdashline{2-10}[2pt/2pt]
       &HiDe-PreT &88.13 &92.17 &70.57 &77.89 &70.72 &74.09 &63.98 &64.18 \\ 
       &HiDe-LoRA &\textbf{89.72} &\textbf{93.34} &\textbf{74.46} &\textbf{80.89} &\textbf{76.10} &\textbf{79.99} &67.73 &68.64 \\  
       &HiDe-Adapter &89.46 &93.12 &74.25 &80.48 &75.17 &79.42 &\textbf{69.62} &\textbf{70.11} \\ 
       \hline
       \multirow{10}*{\tabincell{c}{Sup-Weak}} 
       &L2P \cite{wang2022learning_l2p} &67.73 &78.84 &47.95 &56.51 &43.99 &58.85  &33.25 &38.97 \\ 
       &DualPrompt \cite{wang2022dualprompt} &69.09 &79.56 &51.21 &59.67 &46.05 &58.51 &35.08 &42.99  \\ 
       &S-Prompt++ \cite{wang2022sprompts} &71.05 &81.34 &47.87 &56.62 &42.91 &57.70 &36.20 &43.35 \\ 
       &CODA-Prompt \cite{smith2023coda} &65.45 &76.43 &53.21 &63.61 &44.91 &57.73 &35.59 &41.90 \\ 
       &LAE-PreT \cite{gao2023unified} &67.25 &77.34 &55.55 &64.78 &48.56 &61.73 &36.63 &41.56 \\ 
       &LAE-LoRA \cite{gao2023unified} &68.43 &78.57 &57.40 &66.84 &48.99 &61.50 &35.35 &39.93 \\ 
       &LAE-Adapter \cite{gao2023unified} &68.55 &78.59 &57.92 &67.79 &49.79 &62.25 &37.17 &41.72 \\ 
       \cdashline{2-10}[2pt/2pt]
       &HiDe-PreT &\textbf{77.65} &\textbf{85.14} &57.98 &65.79 &65.03 &71.63 &52.89 &55.09 \\ 
       &HiDe-LoRA &77.46 &84.89 &\textbf{59.40} &\textbf{67.05} &\textbf{66.84} &\textbf{71.91} &52.61 &54.78 \\ 
       &HiDe-Adapter &76.71 &84.55 &58.94 &67.53 &66.26 &71.24 &\textbf{54.38} &\textbf{56.23} \\
       \hline
	\end{tabular}
	} 
	\label{table:overall}
	\vspace{-0.2cm}
\end{table*}

\begin{figure*}[th]
    \centering
    \vspace{-0.1cm}
    \includegraphics[width=0.98\textwidth]{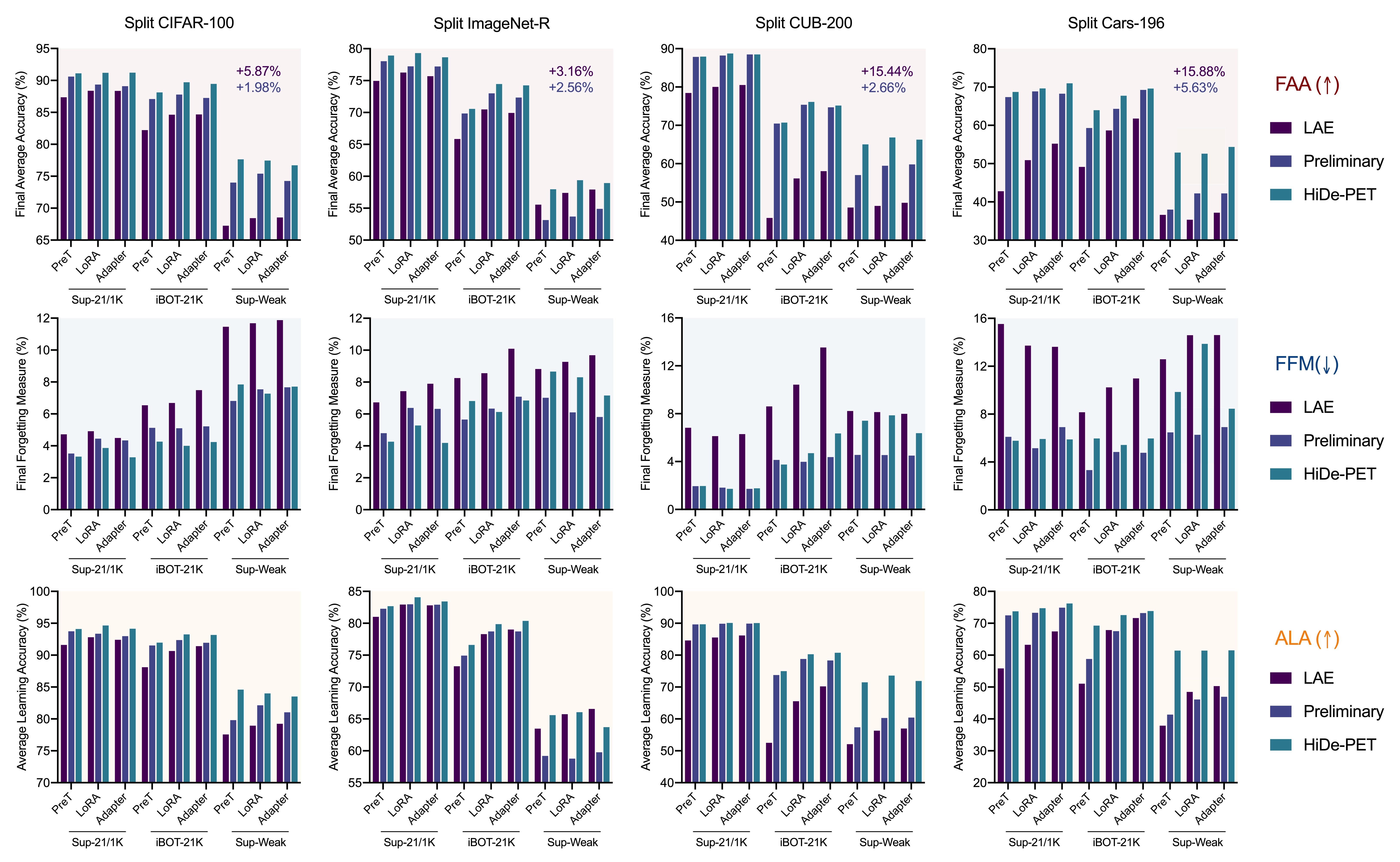}
    \vspace{-0.2cm}
\caption{Comparison of our HiDe-PET, our preliminary version \cite{wang2023hierarchical} and LAE \cite{gao2023unified} implemented with different PET techniques. FAA (\%): final average accuracy. FFM (\%): final forgetting measure of old tasks. ALA (\%): average learning accuracy of new tasks. Note the different range and scale of the y-axis. We present in the first row the average FAA lead of our HiDe-PET over LAE (dark purple) and our preliminary version (dark blue).
}
\vspace{-0.3cm}
\label{CL_Comparison}
\end{figure*}

\textbf{Benchmark:} 
We focus on four datasets that are widely used in previous work to evaluate CL \cite{wang2022learning_l2p,wang2022dualprompt,smith2023coda,gao2023unified,zhang2023slca}, and split each into 10 target tasks of disjoint classes for CIL experiments. The first two are \emph{general datasets}, such as CIFAR-100 \cite{krizhevsky2009learning_cifar} of 100-class small-scale images that are common objects in the real world and ImageNet-R \cite{hendrycks2021many} of 200-class large-scale images that are hard examples of ImageNet-21K \cite{ridnik2021imagenet21k} or newly collected examples of different styles. The latter two are \emph{fine-grained datasets}, such as CUB-200 \cite{wah2011caltech} of 200-class bird images and Cars-196 \cite{krause20133d} of 196-type car images. We mainly consider three representative pre-trained checkpoints that differ in paradigm and dataset, including Sup-21/1K, iBOT-21K and Sup-Weak. Specifically, Sup-21/1K \cite{smith2023coda} is essentially a supervised checkpoint that performs self-supervised learning on ImageNet-21K and then supervised fine-tuning on ImageNet-1K.
iBOT-21K \cite{zhouimage} is a self-supervised checkpoint that currently achieves the first-place classification performance for self-supervised learning on ImageNet-21K. 
Sup-Weak \cite{kim2022multi} is a supervised checkpoint on a subset of ImageNet-1K, in which 389 similar classes to subsequent CL are intentionally removed. 

\textbf{Baseline:} We compare our HiDe-PET with a range of recent strong baselines as described in Sec.~\ref{sec:pet_method}, including L2P \cite{wang2022learning_l2p}, DualPrompt \cite{wang2022dualprompt}, S-Prompt \cite{wang2022sprompts}, CODA-Prompt \cite{smith2023coda} and LAE \cite{gao2023unified}.
In brief, these baselines cover different PET architectures, but mainly target the Prompt-based PET.
LAE \cite{gao2023unified} is the most recent baseline among them, and is also the most relevant one to ours as it applies to a variety of mainstream PET techniques.
Similar to the previous work \cite{wang2023hierarchical}, we modify the implementation of S-Prompt \cite{wang2022sprompts} by inserting task-specific prompts into multiple MSA layers in a PreT manner and using a single-head output layer, in order to evaluate the impact of PET architecture and better adapt to the CIL experiments. The modified S-Prompt is referred to as S-Prompt++.
Following LAE \cite{gao2023unified}, we consider three kinds of mainstream PET techniques in our HiDe-PET and our preliminary version \cite{wang2023hierarchical}, including PreT \cite{li2021prefix}, LoRA \cite{hu2021lora} and Adapter \cite{rebuffi2017learning}.

\textbf{Evaluation:} We use $A_{i, i'}$ to denote the prediction accuracy on task $i$ after learning task $i'$ (with single-head evaluation for CIL), and define the average accuracy of all seen tasks as $AA_{i'} = \frac{1}{i'}\sum_{i=1}^{i'} A_{i, i'}$. After learning all tasks, we report both the final average accuracy ${\rm{FAA}} = AA_{t}$ that serves as the primary metric to evaluate CL performance, and the cumulative average accuracy ${\rm{CAA}} = \frac{1}{t} \sum_{i=1}^t AA_{i}$ that further reflects the historical performance. Moreover, we evaluate the behaviors of different PET techniques with the average learning accuracy ${\rm{ALA}} = \frac{1}{t-1}\sum_{i=2}^t A_{i, i-1}$ for learning plasticity and the final forgetting measure ${\rm{FFM}} = \frac{1}{t-1} \sum_{i=1}^{t-1} \max_{i' \in [t-1]}(A_{i, i'} - A_{i, t})$ for memory stability, as well as evaluate the TII performance in our HiDe-PET.

\textbf{Implementation:} We follow similar implementations as previous work.\footnote{The training regime and supervised checkpoints are identical to those in CODA-Prompt \cite{smith2023coda}, which are slightly different from those in our preliminary version \cite{wang2023hierarchical} and lead to some performance differences.} 
Specifically, we employ a pre-trained ViT-B/16 backbone with an Adam optimizer ($\beta_1=0.9$, $\beta_2=0.999$), a batch size of 128, and a cosine-decaying learning rate of 0.001. We train Split CIFAR-100 for 20 epochs and other benchmarks for 50 epochs to ensure convergence on each task.
The image inputs are resized to $224\times224$ and normalized to $[0,1]$.
The PET architecture of each baseline is also similar to its original paper, which has been shown to yield strong performance. Specifically, L2P \cite{wang2022learning_l2p} is implemented with the prompt pool size $M=30$, the prompt length $d_{\boldsymbol{p}}=5$ and the Top-5 keys.
DualPrompt \cite{wang2022dualprompt} is implemented with the prompt length $d_{\boldsymbol{p}}=5$ of task-shared prompts inserted into layers 1-2 and the prompt length $d_{\boldsymbol{p}}=20$ of task-specific prompts inserted into layers 3-5.
S-Prompt++ \cite{wang2022sprompts} is implemented similarly to DualPrompt but replaces all task-shared prompts with task-specific prompts, inserted into layers 1-5 with the prompt length $d_{\boldsymbol{p}}=20$. 
CODA-Prompt \cite{smith2023coda} is implemented with the prompt pool size $M=100$ and the prompt length $d_{\boldsymbol{p}}=8$, inserted into layers 1-5.
LAE \cite{gao2023unified} and our HiDe-PET are implemented with the prompt length $d_{\boldsymbol{p}}=20$ for PreT, and the low-dimension bottleneck $r=10$ for Adapter and LoRA, inserted into layers 1-5. We insert the Adapter modules in both sequential and parallel manners, while employ LoRA to update both $\boldsymbol{W}_K$ and $\boldsymbol{W}_V$. Therefore, the extra parameter costs of PreT, Adapter and LoRA are identical \cite{gao2023unified}. Note that, our HiDe-PET and our preliminary version \cite{wang2023hierarchical} adopt a similar PET architecture as S-Prompt++, but replace the task-specific keys with an auxiliary output layer $\hat{h}_{\omega}$ to predict the task identity and further preserve statistical information of pre-trained representations.\footnote{We consider a lightweight implementation in terms of the auxiliary output layer and representation recovery, which slightly compromise the performance but largely improve resource efficiency.} 

\subsection{Experimental Result}\label{sec:exp_result}
Now we present the results of our empirical investigation, including the overall performance of all methods, an ablation study of the three hierarchical components, the distinct impacts of implementation strategy, PET technique and PET architecture, as well as the adaptive knowledge accumulation over similar and dissimilar tasks.

\textbf{Overall Performance:} 
Table~\ref{table:overall} summarizes the results of all methods across various pre-trained checkpoints and CL benchmarks. It can be seen that our HiDe-PET implemented with three mainstream PET techniques achieves consistently the highest performance, and the performance lead tends to be more pronounced under the more challenging scenarios. Specifically, the performance of all methods is affected to varying degrees when considering fine-grained tasks (i.e., CUB-200 and Cars-196) and weakened pre-training in terms of self-supervised paradigm (i.e., iBOT-21K) and reduced pre-training samples (i.e., Sup-Weak). Among these competitors, the sub-optimality of Prompt-based PET in CL is clearly exposed, which underperforms LoRA/Adapter-based PET within and across methods. The LoRA/Adapter version of LAE \cite{gao2023unified} is the strongest competitor but still severely affected by the double challenges of pre-trained checkpoints and CL benchmarks. In contrast, our HiDe-PET adapts to them effectively with strong generality. 

\begin{table}[t]
	\centering
    \vspace{-0.1cm}
    \caption{Overall performance of continual learning under DINOv2~\cite{oquabdinov2}. FAA (\%): final average accuracy. CAA (\%): cumulative average accuracy.}
      \vspace{-0.2cm}
	\smallskip
      \renewcommand\arraystretch{1.15}
     \addtolength{\tabcolsep}{-2pt}
	\resizebox{0.48\textwidth}{!}{ 
	\begin{tabular}{c|l|cc|cc}
	 \hline
        \multirow{2}{*}{Setup} & \multirow{2}{*}{\,\,\,\,\,Method} & \multicolumn{2}{c|}{Split ImageNet-R} & \multicolumn{2}{c}{Split Cars-196}\\
        & & FAA ($\uparrow$) & CAA ($\uparrow$) & FAA ($\uparrow$) & CAA ($\uparrow$) \\
        \hline
        \multirow{6}*{\tabincell{c}{DINOv2 \\ LVD-142M}} 
        &LAE-PreT~\cite{gao2023unified} &78.98 &85.26 &45.53 &58.19 \\
        &LAE-LoRA~\cite{gao2023unified} &79.13 &86.04 &52.48 &61.92 \\
        &LAE-Adapter~\cite{gao2023unified} &77.43 &83.98 &57.30 &67.08 \\
        \cdashline{2-6}[2pt/2pt]
        &HiDe-PreT &85.68 &87.70 &\textbf{85.65} &82.20 \\
        &HiDe-LoRA &\textbf{86.26} &89.14 &85.53 &\textbf{84.95} \\
        &HiDe-Adapter &86.03 &\textbf{90.05}&83.48 &84.11 \\
        \hline
	\end{tabular}
	} 
	\label{table:dinov2}
	\vspace{-0.3cm}
\end{table}

It is noteworthy that self-supervised pre-training is often considered more practical than supervised pre-training, owing to the expense of annotating massive pre-training samples \cite{wang2023hierarchical,zhang2023slca}. Meanwhile, Sup-Weak avoids potential overlap between PTMs and target tasks, providing a more restrictive scenario for CL \cite{kim2022multi}. Sup-Weak is also more analogous to the widely used setting of CIL experiments in literature, i.e., the model first learns half of the classes and then learns the other classes in multiple incremental phases, where the baselines of Prompt-based PET have been shown to perform poorly on it \cite{tang2023prompt}. 
These considerations underscore more profound advantages of our HiDe-PET in CL.
We further evaluate CL under DINOv2 of ViT-B/14~\cite{oquabdinov2}, a state-of-the-art self-supervised checkpoint that largely improves iBOT-21K with additional training tricks and more pre-training data, and HiDe-PET consistently outperforms LAE~\cite{gao2023unified} by a wide margin (see Table~\ref{table:dinov2}).

\begin{table}[t]
	\centering
    \vspace{-0.1cm}
    \caption{Ablation study of the three hierarchical components in HiDe-PET. Naive: a naive baseline of only task-specific parameters.}  
      \vspace{-0.2cm}
	\smallskip
      \renewcommand\arraystretch{1.15}
     \addtolength{\tabcolsep}{-2pt}
	\resizebox{0.48\textwidth}{!}{ 
	\begin{tabular}{c|l|cc|cc}
	 \hline
        \multirow{2}{*}{Setup} & \multirow{2}{*}{\,\,\,\,\,\,\,\,\, Method} & \multicolumn{2}{c|}{Split ImageNet-R} & \multicolumn{2}{c}{Split Cars-196}\\
        & & FAA ($\uparrow$) & CAA ($\uparrow$) & FAA ($\uparrow$) & CAA ($\uparrow$) \\
        \hline
       \multirow{5}*{\tabincell{c}{Sup-21/1K \\ PreT}} 
       &Naive &69.77 &76.36 &46.08 &53.59 \\
       &WTP&73.01 &78.20 &48.23 &54.11 \\
       &WTP+TII&75.68 &80.95 &52.89 &54.18 \\
       &WTP+TAP&78.06 &82.69 &61.38 &64.46 \\
       &WTP+TII+TAP &\textbf{78.93} &\textbf{83.44} &\textbf{68.73} &\textbf{69.19} \\
        \hline
        \multirow{5}*{\tabincell{c}{Sup-21/1K \\ LoRA}} 
       &Naive &74.54 &80.66 &45.39 &54.28\\
       &WTP&75.59 &81.31 &48.01 &53.85 \\
       &WTP+TII&76.03 &81.69 &49.62 &57.29 \\
       &WTP+TAP&78.33 &83.68 &63.28 &65.87 \\
       &WTP+TII+TAP&\textbf{79.32} &\textbf{83.97} &\textbf{69.65} &\textbf{69.36} \\
        \hline
        \multirow{5}*{\tabincell{c}{Sup-21/1K \\ Adapter}} 
       &Naive &75.17 &81.46 &47.20 &55.69 \\
       &WTP&76.10 &82.23 &53.12 &59.35 \\
       &WTP+TII&76.80 &82.60 &55.93 &60.89 \\
       &WTP+TAP&76.98 &82.73 &64.65 &67.38 \\
       &WTP+TII+TAP&\textbf{78.65} &\textbf{83.55} &\textbf{70.98} &\textbf{71.31} \\
        \hline
       \multirow{5}*{\tabincell{c}{iBOT-21K \\ PreT}} 
       &Naive &63.78 &73.47 &41.54 &47.96 \\
       &WTP&64.98 &73.54 &52.99 &56.31 \\
       &WTP+TII&66.33 &74.98 &53.89 &57.01 \\
       &WTP+TAP&69.84 &77.02 &59.75 &61.28 \\
       &WTP+TII+TAP&\textbf{70.57} &\textbf{77.89} &\textbf{63.98} &\textbf{64.18} \\
       
        \hline
        \multirow{5}*{\tabincell{c}{iBOT-21K \\ LoRA}} 
       &Naive &67.07 &77.07 &53.13 &59.07 \\
       &WTP&68.06 &77.39 &56.03 &61.18 \\
       &WTP+TII&68.54 &77.60 &59.48 &63.36 \\
       &WTP+TAP&72.60 &79.95 &61.50 &64.88 \\
       &WTP+TII+TAP&\textbf{74.46} &\textbf{80.89} &\textbf{67.73} &\textbf{68.64} \\
        \hline
        \multirow{5}*{\tabincell{c}{iBOT-21K \\ Adapter}} 
       &Naive &68.17 &77.57 &53.58 &59.69 \\
       &WTP&69.11 &77.29 &57.71 &62.21 \\
       &WTP+TII&69.65 &77.90 &62.12 &65.66 \\
       &WTP+TAP&71.32 &79.17 &62.78 &65.59 \\
       &WTP+TII+TAP&\textbf{74.25} &\textbf{80.48} &\textbf{69.62} &\textbf{70.11} \\
        \hline
	\end{tabular}
	} 
	\label{table:ablation}
	\vspace{-0.2cm}
\end{table}

\textbf{Ablation Study:} Table~\ref{table:ablation} presents an ablation study to validate the effectiveness of optimizing the three hierarchical components in HiDe-PET.
Specifically, we progressively incorporate the designs of within-task prediction (WTP), task-identity inference (TII) and task-adaptive prediction (TAP) on the top of a naive architecture, which consists of only task-specific parameters $ \boldsymbol{e}_{1},...,\boldsymbol{e}_{t} $ implemented via mainstream PET techniques. In general, the optimization of each component all contributes to the strong performance of HiDe-PET. Although their contributions are relatively comparable under supervised pre-training and general tasks, the improvement of TAP becomes more significant under self-supervised pre-training and fine-grained tasks, demonstrating the necessity of TAP within the CL objective. 
Besides, the improvement of TII often becomes more apparent with WTP+TAP rather than with WTP alone, suggesting that WTP, TII and TAP operate in concert rather than in isolation.

\textbf{Implementation Strategy:} 
Now we evaluate the implementation strategy of task-shared parameters and representation recovery, which are both important for the optimization of our HiDe-PET and potentially shared by many recent methods. 
In contrast to task-specific parameters discussed above, task-shared parameters $\boldsymbol{g}$ aim to improve pre-trained representations in a task-agnostic manner, demanding effective strategies to mitigate catastrophic forgetting.
Various strategies have been employed in previous work, including 
(1) fix-and-tuning (F\&T) \cite{gao2023unified}, which updates the output layer with frozen $\boldsymbol{g}$ in earlier epochs and then updates $\boldsymbol{g}$ for representation learning in later epochs;
(2) first-session adaptation (FSA) \cite{panos2023first}, which updates $\boldsymbol{g}$ for representation learning exclusively from the first task and then fixes $\boldsymbol{g}$ in subsequent tasks;
(3) slow learner (SL) \cite{zhang2023slca}, which reduces the learning rate of $\boldsymbol{g}$ in all tasks; and
(4) exponential moving average (EMA) \cite{gao2023unified}, which employs an interim copy of $\boldsymbol{g}$ to learn each task and then updates $\boldsymbol{g}$ with a small momentum.

\begin{table}[t]
	\centering
    \caption{Comparison of different strategies for learning task-shared parameters. TII (\%): performance of task identity inference. FAA-U (\%): final average accuracy of learning all classes from uninstructed representations. F\&T: fix-and-tuning. FSA: first-session adaptation. SL: slow learner. EMA: exponential moving average.
    } 
      \vspace{-0.2cm}
	\smallskip
      \renewcommand\arraystretch{1.15}
	\resizebox{0.48\textwidth}{!}{ 
	\begin{tabular}{c|l|cc|cc}
	 \hline
        \multirow{2}{*}{Setup} & \multirow{2}{*}{\,\,Method} & \multicolumn{2}{c|}{Split ImageNet-R} & \multicolumn{2}{c}{Split Cars-196}\\
        & & TII ($\uparrow$) & FAA-U ($\uparrow$) & TII ($\uparrow$) & FAA-U ($\uparrow$)\\
        \hline
       \multirow{5}*{\tabincell{c}{Sup-21/1K \\ PreT}} 
       &F\&T \cite{gao2023unified}&76.45 &74.50 &59.46 &52.94 \\
       &FSA \cite{panos2023first}&75.85 & 73.76 &68.42 &58.85\\
       &SL \cite{zhang2023slca}&77.06 &74.68 &66.13 &56.67 \\
       &EMA \cite{gao2023unified}&76.17 &73.93 &68.35 &58.99 \\
       &FSA+SL &\textbf{77.15} &\textbf{75.02} &\textbf{69.43} &\textbf{59.32} \\
        \hline
       \multirow{5}*{\tabincell{c}{Sup-21/1K \\ LoRA}} 
       &F\&T \cite{gao2023unified}&71.90 &69.85 &64.65 &57.68 \\
       &FSA \cite{panos2023first}&77.74 &75.72 &70.90 &61.37 \\
       &SL \cite{zhang2023slca}&77.26 &75.41 &68.68 &59.33 \\
       &EMA \cite{gao2023unified}&78.33 &\textbf{76.51} &71.20 &61.87 \\
       &FSA+SL &\textbf{78.43} &76.35 &\textbf{71.92} &\textbf{62.89} \\
        \hline
       \multirow{5}*{\tabincell{c}{Sup-21/1K \\ Adapter}} 
       &F\&T \cite{gao2023unified}&75.45 &73.88 &57.16 &52.11 \\
       &FSA \cite{panos2023first}&78.15 &76.30 &72.75 &63.51 \\
       &SL \cite{zhang2023slca}&77.52 &75.98 &63.20 &56.16 \\
       &EMA \cite{gao2023unified}&78.29 &76.30 &73.59 &64.46 \\
       &FSA+SL &\textbf{80.09} &\textbf{78.52} &\textbf{73.71} &\textbf{64.93} \\
       \hline
	\end{tabular}
	} 
	\label{table:shared_pet}
	\vspace{-0.2cm}
\end{table}

However, most of these strategies have their own limitations. Both F\&T and SL impose restrictions on the extent of updates, sacrificing the effectiveness of representation learning and suffering from potential representation shifts in subsequent recovery. FSA adeptly integrates knowledge from the first task and completely avoids representation shifts, but is unable to perform subsequent representation learning. 
Considering their complementary properties, we propose a simple but effective strategy that employs FSA for learning the first task and SL for learning subsequent tasks, which clearly outperforms other strategies (see Table~\ref{table:shared_pet}).
Notably, EMA can be seen as a coarse implementation of FSA+SL and indeed achieves the second-highest performance. Therefore, the task-shared parameters in HiDe-PET may also be implemented with EMA by updating $\boldsymbol{g}$ from each $\boldsymbol{e}_i$ for $i \in [t]$, offering a slight compromise in performance but reducing a half of the training cost. 

On the other hand, we evaluate effective strategies for representation recovery. For classification tasks, the pre-trained representations of each class tend to be single-peaked and therefore can be modeled as a Gaussian with dedicated mean and covariance. Although the covariance achieves considerable performance as shown in Table~\ref{table:representation}, it requires a storage complexity $O(d^2)$ for embedding dimension $d$, which is comparable to the MSA layer of the backbone. There are three alternatives that reduces the storage complexity to $O(d)$, such as simplifying the covariance to variance, preserving randomly selected prototypes, and obtaining multiple representation centroids with $k$NNs. Among them, the multi-centroid demonstrates superior performance and is applicable to different task types, which therefore becomes our default implementation. Interestingly, the variance achieves comparable performance as the covariance and the multi-centroid under general tasks (i.e., Split ImageNet-R) and supervised pre-training (i.e., Sup-21/1K) while requires negligible parameter costs. This further strengthens the advantages of our HiDe-PET in such scenarios targeted by previous work.

Besides, we note that the proposed PET ensemble of task-specific parameters ensures efficiency and scalability due to the lightweight nature of mainstream PET techniques. As described in Sec.~\ref{sec:pet_tech}, ProT and PreT employ $\boldsymbol{p} \in \mathbb{R}^{d_{\boldsymbol{p}} \times d}$ with $d_{\boldsymbol{p}} \ll d$, while Adapter and LoRA employ $\boldsymbol{W}_{{\rm{down}}} \in \mathbb{R}^{d \times r}$ and $\boldsymbol{W}_{{\rm{up}}} \in \mathbb{R}^{r \times d}$ with $r \ll d$. In our default implementation, the additional parameter costs of PreT, Adapter, and LoRA for learning each task are kept the same, i.e., $d_{\boldsymbol{p}}=20$ for PreT, and $r=10$ for Adapter and LoRA, inserted into layers 1-5. Therefore, the additional parameter cost is 0.073M with embedding dimension $d=768$. Even though the model needs to learn a sufficiently long task sequence, e.g., 100 tasks, the total parameter cost is only 7.3M (around 8.5\% of the ViT-B/16 backbone).

\textbf{PET Technique:} 
While mainstream PET techniques universally amount to modulating specific hidden states of the PTMs \cite{he2021towards}, their potential differences in CL are noteworthy. As mentioned above, Prompt-based PET usually lags behind LoRA/Adapter-based PET for both LAE and HiDe-PET (see Table~\ref{table:overall} and Fig.~\ref{CL_Comparison}). The performance gap tends to be more pronounced under the more challenging scenarios of pre-trained checkpoints and CL benchmarks.
A major cause is the limited capacity of Prompt-based PET (i.e., updating attached tokens) in representation learning compared to LoRA/Adapter-based PET (i.e., updating network parameters), especially when handling self-supervised pre-training~\cite{yoo2023improving} and fine-grained tasks~\cite{ma2023visual}, as validated in our results on both task-specific parameters (see WTP in Table~\ref{table:ablation}) and task-shared parameters (see FSA+SL in Table~\ref{table:shared_pet}).

Beyond the overall performance, the choice of PET technique also exerts distinct influences on the three hierarchical components. Compared to Prompt-based PET, LoRA/Adapter-based PET excels in WTP performance through more effectively incorporating task-specific knowledge, but reveals a heightened sensitivity to TII performance, manifested in the errors of predicting an inappropriate set of task-specific parameters (i.e., mismatched representations for each task lead to more errors in the final prediction). This effect is further compensated by learning a robust TAP function.
As shown in Table~\ref{table:ablation}, the effectiveness of TII is remarkably pronounced when coupled with WTP+TAP for LoRA/Adapter-based PET, whereas diminishes for either Prompt-based PET or WTP alone. Moreover, our HiDe-PET outperforms its preliminary version \cite{wang2023hierarchical} especially in LoRA/Adapter-based PET and the more challenging scenarios (see Fig.~\ref{CL_Comparison}), thanks to the improved TII performance with task-shared parameters.

\begin{table}[t]
	\centering
    \caption{Comparison of different strategies for representation recovery. We set $k=10$ for $k$NNs to obtain Multi-Centroid. \#Param: average parameter costs per class, where $d=768$ in this case.} 
      \vspace{-0.2cm}
	\smallskip
      \renewcommand\arraystretch{1.15}
     \addtolength{\tabcolsep}{-2pt}
	\resizebox{0.5\textwidth}{!}{ 
	\begin{tabular}{c|l|cc|cc}
	 \hline
        \multirow{2}{*}{Setup} & \multirow{2}{*}{\,\,\,\,\,\,\,\,\,Method} & \multicolumn{2}{c|}{Split ImageNet-R} & \multicolumn{2}{c}{Split Cars-196}\\
        & & FAA ($\uparrow$) & \#Param ($\downarrow$) & FAA ($\uparrow$) & \#Param ($\downarrow$) \\
        \hline
       \multirow{5}*{\tabincell{c}{Sup-21/1K \\ PreT}} 
       &No Recovery&75.68 &0 &52.89 &0 \\ 
       &Prototype&76.88 &10$d$ &62.65 & 10$d$ \\
       &Variance&77.54 &1$d$ &57.04 & 1$d$ \\
       &Covariance&77.58 &$d^2$ &\textbf{73.14} & $d^2$\\
       &Multi-Centroid&\textbf{78.93} &9.5$d$ &68.73 &8.0$d$ \\
        \hline
        \multirow{5}*{\tabincell{c}{iBOT-21K \\ PreT}} 
       &No Recovery&58.88 &0 &41.89 &0 \\
       &Prototype&67.08 &10$d$ &46.34 & 10$d$ \\
       &Variance&70.55 &1$d$ &48.27 & 1$d$ \\
       &Covariance&68.85 &$d^2$ &\textbf{66.42} & $d^2$ \\
       &Multi-Centroid&\textbf{70.57} &9.5$d$ &63.98 & 8.7$d$ \\
        \hline
        \multirow{5}*{\tabincell{c}{Sup-Weak \\ PreT}} 
       &No Recovery&54.63 &0 &44.35 & 0 \\
       &Prototype&55.06 &10$d$ &46.91 & 10$d$ \\
       &Variance&55.49 &1$d$ &47.77 & 1$d$ \\
       &Covariance&57.46 &$d^2$ &\textbf{56.06} & $d^2$ \\
       &Multi-Centroid&\textbf{57.98} &9.1$d$ &52.89 &8.7$d$  \\
        \hline
	\end{tabular}
	} 
	\label{table:representation}
	\vspace{-0.4cm}
\end{table}

\begin{table*}[ht]
	\centering
    \vspace{-0.2cm}
    \caption{Evaluation of adaptive knowledge accumulation (AKA) with LoRA-based PET. Full (\%): average accuracy of learning subsequent tasks with all training samples. Few (\%): average accuracy of learning subsequent tasks with a few training samples (5 per class).}
      \vspace{-0.2cm}
	\smallskip
      \renewcommand\arraystretch{1.15}
         \resizebox{0.90\textwidth}{!}{ 
	\begin{tabular}{c|c|cc|cc|cc|cc|cc}
	 \hline
        \multirow{2}{*}{PTM} & \multirow{2}{*}{Method} &\multicolumn{2}{c|}{Split Dogs-120} & \multicolumn{2}{c|}{Split CUB-200} & \multicolumn{2}{c|}{Split Cars-196} & \multicolumn{2}{c|}{Split Aircraft-102}  & \multicolumn{2}{c}{CL of Mixture}\\
        & & Full ($\uparrow$) & Few ($\uparrow$) & Full ($\uparrow$) & Few ($\uparrow$) & Full ($\uparrow$) & Few ($\uparrow$) & Full ($\uparrow$) & Few ($\uparrow$)  & FAA ($\uparrow$) & CAA ($\uparrow$)\\
        \hline
        \multirow{2}*{\tabincell{c}{Sup-21/1K}} 
       & w/o AKA &92.32 &88.36 &89.51 &81.26 &83.77 &57.04 &77.98 &55.56 &77.32 &81.14 \\ 
       & w/ \, AKA &92.50 &88.92 &91.39 &84.91 &89.46 &64.30 &83.38 &62.37 &83.27 &86.78 \\ 
        \hline
       \multirow{2}*{\tabincell{c}{iBOT-21K}} 
       & w/o AKA &81.70 &54.08 &74.66 &45.79 &79.27 &38.15 &79.24 &53.97 &68.38 &74.57 \\ 
       & w/ \, AKA &84.41 &65.10 &82.95 &62.30 &88.32 &56.12 &85.06 &62.76 &74.99 &81.55 \\ 
       \hline
       \multirow{2}*{\tabincell{c}{Sup-Weak}} 
       & w/o AKA &88.14 &80.88 &71.10 &47.46 &59.03 &36.51 &62.68 &35.78 &53.23 &58.54 \\ 
       & w/ \, AKA &88.17 &81.38 &77.50 &56.92 &77.42 &50.37 &72.64 &46.32 &65.48 &70.41 \\ 
       \hline
	\end{tabular}
	} 
	\label{table:adaptive_knowledge}
	\vspace{-0.2cm}
\end{table*}

\textbf{PET Architecture:} The generality of our HiDe-PET is also reflected in its PET architecture, which strategically exploits both task-specific and task-shared parameters for representation learning. These two kinds of parameters are used to acquire knowledge with different levels of differentiation and need to overcome their respective challenges as described in Sec.~\ref{sec:pet_method}.
Within HiDe-PET, they both contribute to the outstanding performance in Table~\ref{table:overall} and complement each other (see WTP in Table~\ref{table:ablation} and FSA+SL in Table~\ref{table:shared_pet}). 
In contrast, our preliminary version \cite{wang2023hierarchical} and LAE \cite{gao2023unified} exclusively engage either task-specific or task-shared parameters, missing out on fully harnessing the benefits of PTMs and PET.
We further present an extensive comparison of our HiDe-PET, our preliminary version and LAE in terms of the overall performance, memory stability and learning plasticity, so as to better demonstrate the respective contributions of different PET architectures (see Fig.~\ref{CL_Comparison}). 
It can be seen that the use of task-shared parameters in HiDe-PET can largely improve ALA of new tasks and FAA of all tasks, but probably results in a slight increase in FFM of old tasks due to the ongoing updates in our default implementation (i.e., FSA+SL). This trend is comparably more pronounced under Sup-Weak that is more demanding to update the pre-trained representations, with FAA, ALA and FFM increasing on average by 6.83\%, 9.64\% and 2.21\%, respectively. We further evaluate alternative implementations of task-shared parameters in Table~\ref{table:shared_pet}, where FSA has no extra forgetting but performs less well than FSA+SL. The users may select appropriate implementations according to their customised requirements of evaluation metrics.

It is noteworthy that previous work such as L2P \cite{wang2022learning_l2p} and DualPrompt \cite{wang2022dualprompt} also explicitly or implicitly exploit both task-specific and task-shared prompts, but in a \emph{sequential} manner to instruct representation learning of each task (corresponding to WTP in our framework). 
In contrast, our HiDe-PET optimizes these two kinds of parameters in a \emph{parallel} manner to improve the three hierarchical components, allowing for a more adequate differentiation of the acquired knowledge.
Interesting, using only task-shared parameters coupled with representation recovery within HiDe-PET (i.e., FSA+SL in Table~\ref{table:shared_pet}) has already achieved better performance than these methods (see Sec.~\ref{sec:discussion} for a conceptual explanation), serving as a strong baseline to evaluate current progress. 
The inherent connections of task-specific and task-shared parameters will be further explored below with a PET hierarchy inspired by OOD detection.

\textbf{Adaptive Knowledge Accumulation:}
As analyzed in Sec.~\ref{sec:ood_method}, the use of $\boldsymbol{e}_{1},...,\boldsymbol{e}_{t}$ and $\boldsymbol{g}$ can be seen as a special case tailored for target tasks randomly split from the same dataset, i.e., there is no actual task structure. 
When considering more realistic CL scenarios with apparent similarity and dissimilarity between task distributions, we devise a hierarchy of expandable parameter sets $\boldsymbol{g}_1,...,\boldsymbol{g}_k$ upon OOD detection to achieve adaptive knowledge accumulation (AKA), and focus on the LoRA-based PET to examine if the pre-trained knowledge can evolve flexibly with target tasks in CL.
Here we construct such scenario with the two fine-grained datasets (i.e., CUB-200 \cite{wah2011caltech} and Cars-196 \cite{krause20133d}) and another two (i.e., Dogs-120 \cite{khosla2011novel} and Aircraft-102 \cite{maji2013fine}), which cover both natural and artificial objects.
Each dataset is randomly split into 10 tasks. We collect 5 tasks per dataset and mix then as a task sequence (20 tasks in total) for CL, while leaving the rest 5 tasks per dataset for validation. 

\begin{figure}[t]
    \centering
    \includegraphics[width=0.50\textwidth]{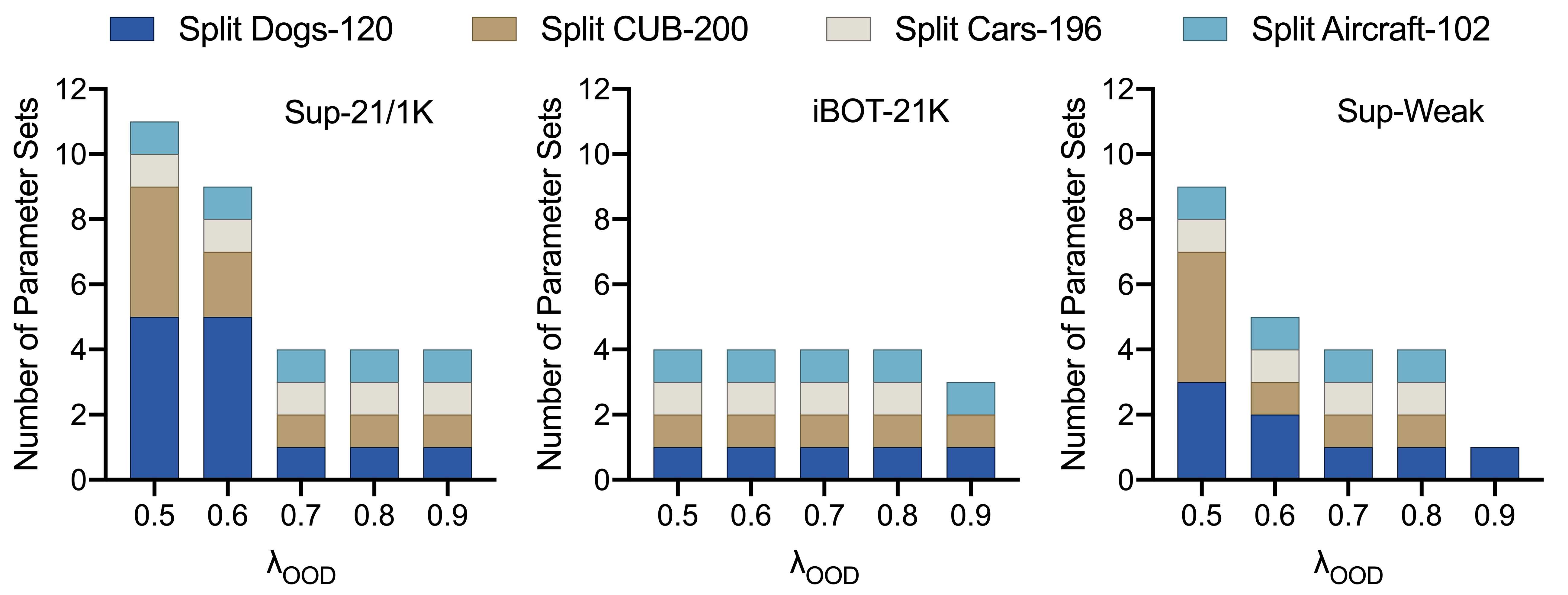}
    \vspace{-0.7cm}
\caption{OOD detection threshold $\lambda_{{\rm{OOD}}}$ and the number of expanded parameter sets $\boldsymbol{g}_{1},...,\boldsymbol{g}_{k}$ under different pre-trained checkpoints.}
\vspace{-0.4cm}
\label{OOD_Threshold}
\end{figure}

In CL, the OOD detection threshold $\lambda_{{\rm{OOD}}}$ determines the expansion of parameter sets. As shown in Fig.~\ref{OOD_Threshold}, using a larger $\lambda_{{\rm{OOD}}}$ tends to expand fewer parameter sets, and vice versa. In particular, the choice of $\lambda_{{\rm{OOD}}}$ is relatively insensitive and can consistently construct one parameter set for each dataset (with $\lambda_{{\rm{OOD}}}=0.7$~or~$0.8$) under different pre-trained checkpoints. Then we validate the effectiveness of AKA in Table~\ref{table:adaptive_knowledge}.
Inspired by a recent work \cite{liao2023does}, we evaluate the improvements of pre-trained knowledge through the average accuracy of learning the validation tasks under large-shot or few-shot setting. Through selecting the most relevant $\boldsymbol{g}_j$ and adding it to $\theta$, the pre-trained backbone $f_{\theta}$ is able to learn each task more effectively. With the improved $f_{\theta}$, the overall performance of CL (i.e., FAA and CAA) for the mixed task sequence is also significantly enhanced.

Interestingly, the idea of updating the pre-trained backbone with a mixture of LoRA experts \cite{wu2023mole} has been shown effective to accumulate knowledge from multi-task learning, which is consistent with our results. In contrast, the design of the OOD detection, FSA+SL, and representation recovery enables our HiDe-PET to achieve this purpose in a lifelong manner. Besides, our HiDe-PET can also adapt to task-agnostic CL \cite{wang2023comprehensive} through expanding $\boldsymbol{e}_{1},...,\boldsymbol{e}_{t}$ upon the OOD detection.
We leave it as a further work.

\section{Discussion and Conclusion}\label{sec:discussion}

In this work, we present a unified framework for CL with PTMs and PET, in order to advance this direction with improved effectiveness and generality. 
Our framework features a profound integration of theoretical and empirical insights, a broad coverage of relevant techniques, as well as a robust adaptation to different scenarios. Considering the particular impact of pre-trained knowledge on CL, we decompose the CL objective into three hierarchical components, i.e., WTP, TII and TAP, and devise an innovative approach to explicitly optimize them with mainstream PET techniques.
During the optimization process, pre-trained representations are effectively instructed via task-specific and task-shared PET architectures, and are efficiently recovered through preserving their statistical information.

Our framework allows for a comprehensive evaluation of various technical elements inherent in CL with PTMs and PET.
Through an extensive empirical investigation, we demonstrate the better performance of LoRA/Adapter-based PET over Prompt-based PET within both task-specific and task-shared PET architectures, which tends to be more evident under the more challenging scenarios in terms of pre-trained checkpoints and CL benchmarks.
We also unravel the distinct behaviors of different PET techniques in response to the three hierarchical components, as well as the respective challenges and complementary effects of different PET architectures. 
These technical elements are potentially shared by many recent methods, making it possible to dissect their specific implementations and incorporate the most appropriate ones. 
Owning to the above extensive explorations, our approach achieves remarkably superior performance across various CL scenarios over a wide range of recent strong baselines.

Intriguingly, the correspondence of our approach to the three hierarchical components suggests a more profound connection between existing methods. As discussed in Sec.~\ref{sec:pet_method}, the use of task-specific parameters \cite{wang2022learning_l2p,wang2022dualprompt,smith2023coda,wang2022sprompts,wang2023hierarchical} necessitates learning to predict their identities, equivalent to optimizing the decomposed WTP performance $P(\boldsymbol{x} \in \mathcal{X}_{\bar{i},\bar{j}}|\boldsymbol{x} \in \mathcal{X}_{\bar{i}},\mathcal{D},\theta)$ and TII performance $P(\boldsymbol{x} \in \mathcal{X}_{\bar{i}}|\mathcal{D},\theta)$. In contrast, the use of task-shared parameters \cite{wang2022learning_l2p,wang2022dualprompt,mcdonnell2023ranpac,gao2023unified,zhang2023slca} needs to overcome catastrophic forgetting, equivalent to optimizing the pre-decomposed performance $P(\boldsymbol{x} \in \mathcal{X}_{\bar{i},\bar{j}}|\mathcal{D},\theta)$. On the top of representation learning, the use of representation recovery \cite{mcdonnell2023ranpac,zhang2023slca,tran2023koppa} to rectify the output layer further improves the TAP performance $P(\boldsymbol{x} \in \mathcal{X}^{y}|\mathcal{D},\theta)$. This connection is summarized by the multi-objective optimization problem in Eq.~\eqref{FinalObjective}, and also demonstrates why our approach clearly outperforms other baselines and why the use of only task-shared parameters and representation recovery (i.e., FSA+SL in Table~\ref{table:shared_pet}) is powerful enough. Subsequent efforts in CL with PTMs and PET could employ this as a theoretical reference to develop more advanced methods.

Moreover, the hierarchical decomposition along with the design of our approach showcase a close relationship with the mechanisms of robust biological CL. 
In the mammalian brain, the memory of an experience is consolidated with the interplay of hippocampus and neocortex, known as the complementary learning system theory \cite{kumaran2016learning,mcclelland1995there} that has been widely used to inspire CL of AI. The hippocampus-depended and neocortex-depended memories tend to be more specific and more generalized, respectively \cite{frankland2005organization,wang2021triple}, and the retrieval of these two memory paths is adaptively switched from the concrete scenarios \cite{goshen2011dynamics}. 
Within hippocampus, the activation of distinct populations of memory cells also undergoes adaptive switching \cite{lei2022social}, and the neural representations of previous experiences are frequently recovered \cite{kudithipudi2022biological}.
The entire process is consistent with the parallel organization of task-specific and task-shared parameters, the exclusive selection of the former and the representation recovery of task distributions.

In the era of large-scale PTMs, we would emphasize the pressing need for these adaptive algorithms that are designed with machine learning fundamentals and real-world considerations.
By leveraging the power of PTMs and the adaptability of CL, we can customize solutions to address the unique challenges posed by specific domains, and envision extending our approach to numerous areas such as healthcare, robotics and industrial manufacturing.
Such an elevated goal requires extending the target of CL from homogeneous to heterogeneous tasks, which also provides novel opportunities to explore generalizable knowledge behind them.
Taken together, we expect this work to not only facilitate direct applications but also set the stage for the robustness, adaptability and reliability of future AI systems, as a general purpose of CL research.

\ifCLASSOPTIONcompsoc
  \section*{Acknowledgments}
\else
  \section*{Acknowledgment}
\fi
This work was supported by the NSFC Projects (Nos.~62406160, 62350080, 62106123, 62106120, 92370124, 92248303), Beijing Natural Science Foundation L247011, Tsinghua Institute for Guo Qiang, and the High Performance Computing Center, Tsinghua University. L.W. is also supported by the Postdoctoral Fellowship Program of CPSF under Grant Number GZB20230350 and the Shuimu Tsinghua Scholar. J.Z. is also supported by the XPlorer Prize.

\ifCLASSOPTIONcaptionsoff
  \newpage
\fi



%
\bibliographystyle{IEEEtran} 
\bibliography{egbib}

\begin{thebibliography}{10}
\providecommand{\url}[1]{#1}
\csname url@samestyle\endcsname
\providecommand{\newblock}{\relax}
\providecommand{\bibinfo}[2]{#2}
\providecommand{\BIBentrySTDinterwordspacing}{\spaceskip=0pt\relax}
\providecommand{\BIBentryALTinterwordstretchfactor}{4}
\providecommand{\BIBentryALTinterwordspacing}{\spaceskip=\fontdimen2\font plus
\BIBentryALTinterwordstretchfactor\fontdimen3\font minus \fontdimen4\font\relax}
\providecommand{\BIBforeignlanguage}[2]{{%
\expandafter\ifx\csname l@#1\endcsname\relax
\typeout{** WARNING: IEEEtran.bst: No hyphenation pattern has been}%
\typeout{** loaded for the language `#1'. Using the pattern for}%
\typeout{** the default language instead.}%
\else
\language=\csname l@#1\endcsname
\fi
#2}}
\providecommand{\BIBdecl}{\relax}
\BIBdecl

\bibitem{ramasesh2021effect}
V.~V. Ramasesh, A.~Lewkowycz, and E.~Dyer, ``Effect of scale on catastrophic forgetting in neural networks,'' in \emph{ICLR}, 2021.

\bibitem{mehta2021empirical}
S.~V. Mehta \emph{~et~al.}, ``An empirical investigation of the role of pre-training in lifelong learning,'' \emph{arXiv preprint arXiv:2112.09153}, 2021.

\bibitem{wang2023comprehensive}
L.~Wang \emph{~et~al.}, ``A comprehensive survey of continual learning: Theory, method and application,'' \emph{IEEE TPAMI}, 2024.

\bibitem{zhang2023slca}
G.~Zhang \emph{~et~al.}, ``Slca: Slow learner with classifier alignment for continual learning on a pre-trained model,'' in \emph{ICCV}, 2023.

\bibitem{wang2022learning_l2p}
Z.~Wang \emph{~et~al.}, ``Learning to prompt for continual learning,'' in \emph{CVPR}, 2022.

\bibitem{wang2022dualprompt}
Z.~Wang \emph{~et~al.}, ``Dualprompt: Complementary prompting for rehearsal-free continual learning,'' in \emph{ECCV}, 2022.

\bibitem{wang2022sprompts}
Y.~Wang, Z.~Huang, and X.~Hong, ``S-prompts learning with pre-trained transformers: An occam's razor for domain incremental learning,'' \emph{NeurIPS}, 2022.

\bibitem{smith2023coda}
J.~S. Smith \emph{~et~al.}, ``Coda-prompt: Continual decomposed attention-based prompting for rehearsal-free continual learning,'' in \emph{CVPR}, 2023.

\bibitem{lester2021power}
B.~Lester, R.~Al-Rfou, and N.~Constant, ``The power of scale for parameter-efficient prompt tuning,'' in \emph{EMNLP}, 2021.

\bibitem{li2021prefix}
X.~L. Li and P.~Liang, ``Prefix-tuning: Optimizing continuous prompts for generation,'' in \emph{ACL-IJCNLP}, 2021.

\bibitem{rebuffi2017learning}
S.-A. Rebuffi, H.~Bilen, and A.~Vedaldi, ``Learning multiple visual domains with residual adapters,'' \emph{NeurIPS}, 2017.

\bibitem{hu2021lora}
E.~J. Hu \emph{~et~al.}, ``Lora: Low-rank adaptation of large language models,'' \emph{arXiv preprint arXiv:2106.09685}, 2021.

\bibitem{wang2023hierarchical}
L.~Wang \emph{~et~al.}, ``Hierarchical decomposition of prompt-based continual learning: Rethinking obscured sub-optimality,'' \emph{NeurIPS}, 2023.

\bibitem{gao2023unified}
Q.~Gao \emph{~et~al.}, ``A unified continual learning framework with general parameter-efficient tuning,'' in \emph{ICCV}, 2023.

\bibitem{parisi2019continual}
G.~I. Parisi \emph{~et~al.}, ``Continual lifelong learning with neural networks: A review,'' \emph{Neur. Netw.}, 2019.

\bibitem{kirkpatrick2017overcoming}
J.~Kirkpatrick \emph{~et~al.}, ``Overcoming catastrophic forgetting in neural networks,'' \emph{PNAS}, 2017.

\bibitem{wang2021afec}
L.~Wang \emph{~et~al.}, ``Afec: Active forgetting of negative transfer in continual learning,'' \emph{NeurIPS}, 2021.

\bibitem{aljundi2018memory_mas}
R.~Aljundi \emph{~et~al.}, ``Memory aware synapses: Learning what (not) to forget,'' in \emph{ECCV}, 2018.

\bibitem{rebuffi2017icarl}
S.-A. Rebuffi \emph{~et~al.}, ``icarl: Incremental classifier and representation learning,'' in \emph{CVPR}, 2017.

\bibitem{shin2017continual_dgr}
H.~Shin \emph{~et~al.}, ``Continual learning with deep generative replay,'' \emph{NeurIPS}, 2017.

\bibitem{wang2021memory}
L.~Wang \emph{~et~al.}, ``Memory replay with data compression for continual learning,'' in \emph{ICLR}, 2021.

\bibitem{pham2021dualnet}
Q.~Pham, C.~Liu, and S.~Hoi, ``Dualnet: Continual learning, fast and slow,'' \emph{NeurIPS}, 2021.

\bibitem{cha2021co2l}
H.~Cha, J.~Lee, and J.~Shin, ``Co2l: Contrastive continual learning,'' in \emph{ICCV}, 2021.

\bibitem{ostapenko2022foundational}
O.~Ostapenko \emph{~et~al.}, ``Foundational models for continual learning: An empirical study of latent replay,'' in \emph{CoLLAs}, 2022.

\bibitem{lopez2017gradient_gem}
D.~Lopez-Paz and M.~Ranzato, ``Gradient episodic memory for continual learning,'' \emph{NeurIPS}, 2017.

\bibitem{wang2021training}
S.~Wang \emph{~et~al.}, ``Training networks in null space of feature covariance for continual learning,'' in \emph{CVPR}, 2021.

\bibitem{saha2020gradient}
G.~Saha, I.~Garg, and K.~Roy, ``Gradient projection memory for continual learning,'' in \emph{ICLR}, 2020.

\bibitem{serra2018overcoming}
J.~Serra \emph{~et~al.}, ``Overcoming catastrophic forgetting with hard attention to the task,'' in \emph{ICML}, 2018.

\bibitem{wang2022coscl}
L.~Wang \emph{~et~al.}, ``Coscl: Cooperation of small continual learners is stronger than a big one,'' in \emph{ECCV}, 2022.

\bibitem{wang2023incorporating}
L.~Wang \emph{~et~al.}, ``Incorporating neuro-inspired adaptability for continual learning in artificial intelligence,'' \emph{Nat. Mach. Intell.}, 2023.

\bibitem{wu2019large_bic}
Y.~Wu \emph{~et~al.}, ``Large scale incremental learning,'' in \emph{CVPR}, 2019.

\bibitem{knoblauch2020optimal}
J.~Knoblauch, H.~Husain, and T.~Diethe, ``Optimal continual learning has perfect memory and is np-hard,'' in \emph{ICML}, 2020.

\bibitem{he2021towards}
J.~He \emph{~et~al.}, ``Towards a unified view of parameter-efficient transfer learning,'' in \emph{ICLR}, 2021.

\bibitem{mcdonnell2023ranpac}
M.~D. McDonnell \emph{~et~al.}, ``Ranpac: Random projections and pre-trained models for continual learning,'' \emph{NeurIPS}, 2023.

\bibitem{jia2022visual}
M.~Jia \emph{~et~al.}, ``Visual prompt tuning,'' in \emph{ECCV}, 2022.

\bibitem{yoo2023improving}
S.~Yoo \emph{~et~al.}, ``Improving visual prompt tuning for self-supervised vision transformers,'' in \emph{ICML}, 2023.

\bibitem{van2019three}
G.~M. Van~de Ven and A.~S. Tolias, ``Three scenarios for continual learning,'' \emph{arXiv preprint arXiv:1904.07734}, 2019.

\bibitem{vaswani2017attention}
A.~Vaswani \emph{~et~al.}, ``Attention is all you need,'' \emph{NeurIPS}, 2017.

\bibitem{dosovitskiy2020image_vit}
A.~Dosovitskiy \emph{~et~al.}, ``An image is worth 16x16 words: Transformers for image recognition at scale,'' in \emph{ICLR}, 2020.

\bibitem{ma2023visual}
X.~Ma \emph{~et~al.}, ``When visual prompt tuning meets source-free domain adaptive semantic segmentation,'' \emph{NeurIPS}, 2023.

\bibitem{kim2022theoretical}
G.~Kim \emph{~et~al.}, ``A theoretical study on solving continual learning,'' \emph{NeurIPS}, 2022.

\bibitem{yang2021generalized}
J.~Yang \emph{~et~al.}, ``Generalized out-of-distribution detection: A survey,'' \emph{arXiv preprint arXiv:2110.11334}, 2021.

\bibitem{panos2023first}
A.~Panos \emph{~et~al.}, ``First session adaptation: A strong replay-free baseline for class-incremental learning,'' \emph{arXiv preprint arXiv:2303.13199}, 2023.

\bibitem{tran2023koppa}
Q.~Tran \emph{~et~al.}, ``Koppa: Improving prompt-based continual learning with key-query orthogonal projection and prototype-based one-versus-all,'' \emph{arXiv preprint arXiv:2311.15414}, 2023.

\bibitem{sun2022out}
Y.~Sun \emph{~et~al.}, ``Out-of-distribution detection with deep nearest neighbors,'' in \emph{ICML}, 2022.

\bibitem{krizhevsky2009learning_cifar}
A.~Krizhevsky, G.~Hinton \emph{~et~al.}, ``Learning multiple layers of features from tiny images,'' Tech. Rep., 2009.

\bibitem{hendrycks2021many}
D.~Hendrycks, S.~Basart \emph{~et~al.}, ``The many faces of robustness: A critical analysis of out-of-distribution generalization,'' in \emph{ICCV}, 2021.

\bibitem{ridnik2021imagenet21k}
T.~Ridnik \emph{~et~al.}, ``Imagenet-21k pretraining for the masses,'' \emph{arXiv preprint arXiv:2104.10972}, 2021.

\bibitem{wah2011caltech}
C.~Wah \emph{~et~al.}, ``The caltech-ucsd birds-200-2011 dataset,'' 2011.

\bibitem{krause20133d}
J.~Krause \emph{~et~al.}, ``3d object representations for fine-grained categorization,'' in \emph{ICCVW}, 2013.

\bibitem{zhouimage}
J.~Zhou \emph{~et~al.}, ``Image bert pre-training with online tokenizer,'' in \emph{ICLR}, 2021.

\bibitem{kim2022multi}
G.~Kim, B.~Liu, and Z.~Ke, ``A multi-head model for continual learning via out-of-distribution replay,'' in \emph{CoLLAs}, 2022.

\bibitem{oquabdinov2}
M.~Oquab \emph{~et~al.}, ``Dinov2: Learning robust visual features without supervision,'' \emph{TMLR}.

\bibitem{tang2023prompt}
Y.-M. Tang, Y.-X. Peng, and W.-S. Zheng, ``When prompt-based incremental learning does not meet strong pretraining,'' in \emph{ICCV}, 2023.

\bibitem{khosla2011novel}
A.~Khosla \emph{~et~al.}, ``Novel dataset for fine-grained image categorization: Stanford dogs,'' in \emph{CVPRW}, 2011.

\bibitem{maji2013fine}
S.~Maji \emph{~et~al.}, ``Fine-grained visual classification of aircraft,'' \emph{arXiv preprint arXiv:1306.5151}, 2013.

\bibitem{liao2023does}
W.~Liao \emph{~et~al.}, ``Does continual learning meet compositionality? new benchmarks and an evaluation framework,'' \emph{NeurIPS}, 2023.

\bibitem{wu2023mole}
X.~Wu, S.~Huang, and F.~Wei, ``Mole: Mixture of lora experts,'' in \emph{ICLR}, 2023.

\bibitem{kumaran2016learning}
D.~Kumaran, D.~Hassabis, and J.~L. McClelland, ``What learning systems do intelligent agents need? complementary learning systems theory updated,'' \emph{Trends Cogn. Sci.}, 2016.

\bibitem{mcclelland1995there}
J.~L. McClelland, B.~L. McNaughton, and R.~C. O'Reilly, ``Why there are complementary learning systems in the hippocampus and neocortex: insights from the successes and failures of connectionist models of learning and memory.'' \emph{Psychol. Rev.}, 1995.

\bibitem{frankland2005organization}
P.~W. Frankland and B.~Bontempi, ``The organization of recent and remote memories,'' \emph{Nature Rev. Neurosci.}, 2005.

\bibitem{wang2021triple}
L.~Wang \emph{~et~al.}, ``Triple-memory networks: A brain-inspired method for continual learning,'' \emph{IEEE TNNLS}, 2021.

\bibitem{goshen2011dynamics}
I.~Goshen \emph{~et~al.}, ``Dynamics of retrieval strategies for remote memories,'' \emph{Cell}, 2011.

\bibitem{lei2022social}
B.~Lei \emph{~et~al.}, ``Social experiences switch states of memory engrams through regulating hippocampal rac1 activity,'' \emph{PNAS}, 2022.

\bibitem{kudithipudi2022biological}
D.~Kudithipudi \emph{~et~al.}, ``Biological underpinnings for lifelong learning machines,'' \emph{Nat. Mach. Intell.}, 2022.

\bibitem{zuo2024hierarchical}
Y.~Zuo \emph{~et~al.}, ``Hierarchical prompts for rehearsal-free continual learning,'' \emph{arXiv preprint arXiv:2401.11544}, 2024.

\end{thebibliography}


%


\vspace{-35pt}
\begin{IEEEbiography}
[{\includegraphics[width=1in,height=1.25in,clip,keepaspectratio]{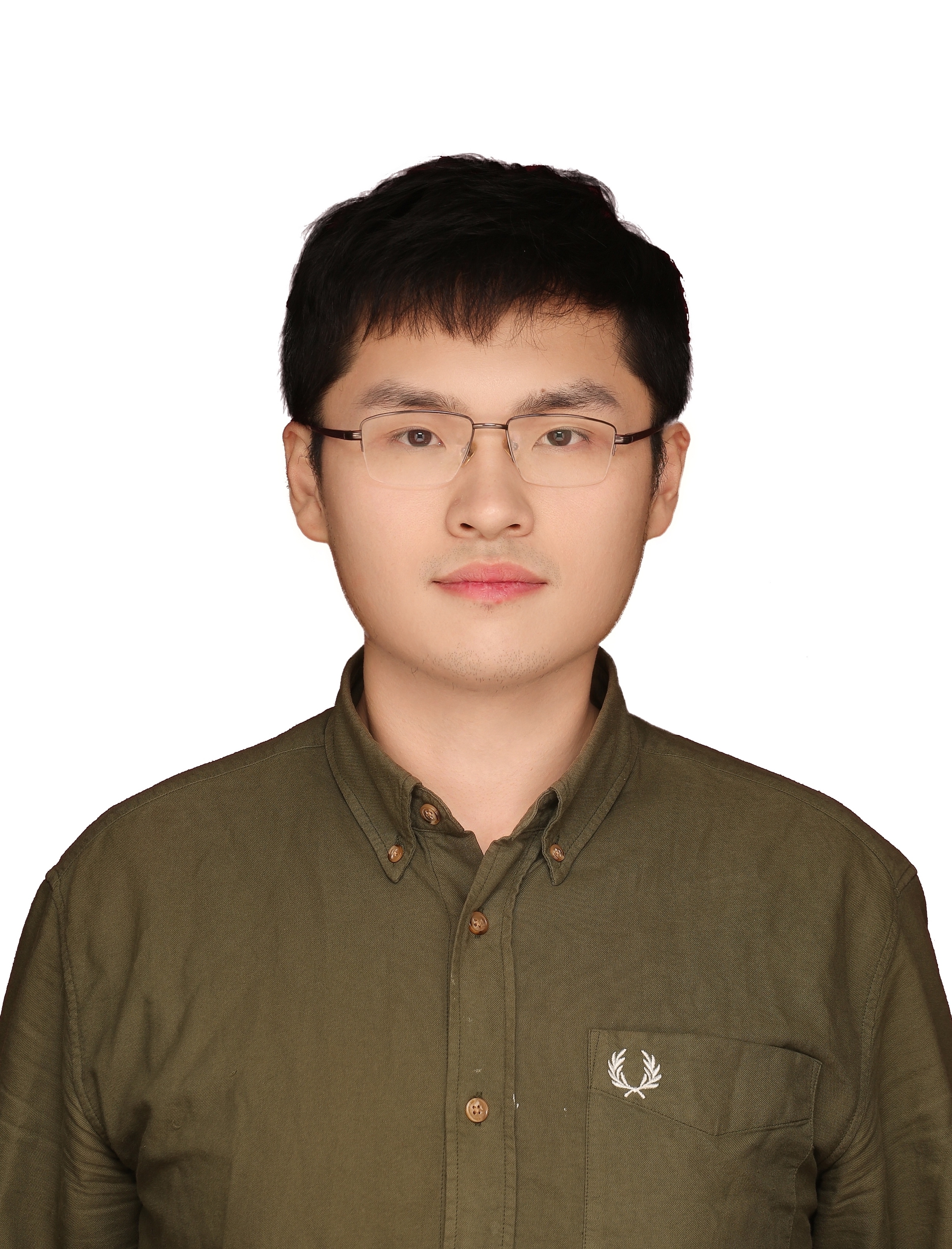}}]{Liyuan Wang}
is currently a postdoctoral researcher in Tsinghua University, working with Prof.~Jun Zhu at the Department of Computer Science and Technology. Before that, he received the B.S. and Ph.D. degrees from Tsinghua University. His research interests include continual learning, incremental learning, lifelong learning and brain-inspired AI. His work in continual learning has been published in major conferences and journals in related fields, such as Nature Machine Intelligence, IEEE TPAMI, IEEE TNNLS, NeurIPS, ICLR, CVPR, ICCV, ECCV, etc.
\end{IEEEbiography}

\vspace{-35pt}
\begin{IEEEbiography}
[{\includegraphics[width=1in,height=1.25in,clip,keepaspectratio]{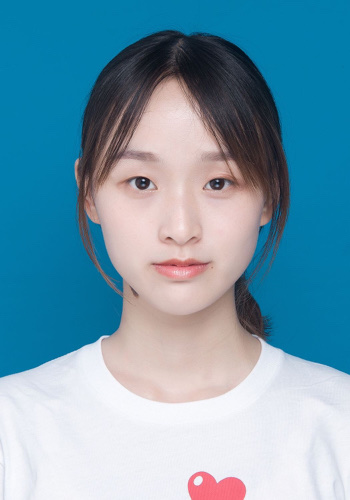}}]{Jingyi Xie} is currently a researcher engineer in Prof.~Jun Zhu's group. She received the B.Sc. and M.Sc. degree with the Department of Mathematics and Statistics, Wuhan University, Wuhan, China. 
Her current research interests include representation learning, continual learning, and deep learning.
\end{IEEEbiography}

\vspace{-35pt}
\begin{IEEEbiography}
[{\includegraphics[width=1in,height=1.25in,clip,keepaspectratio]{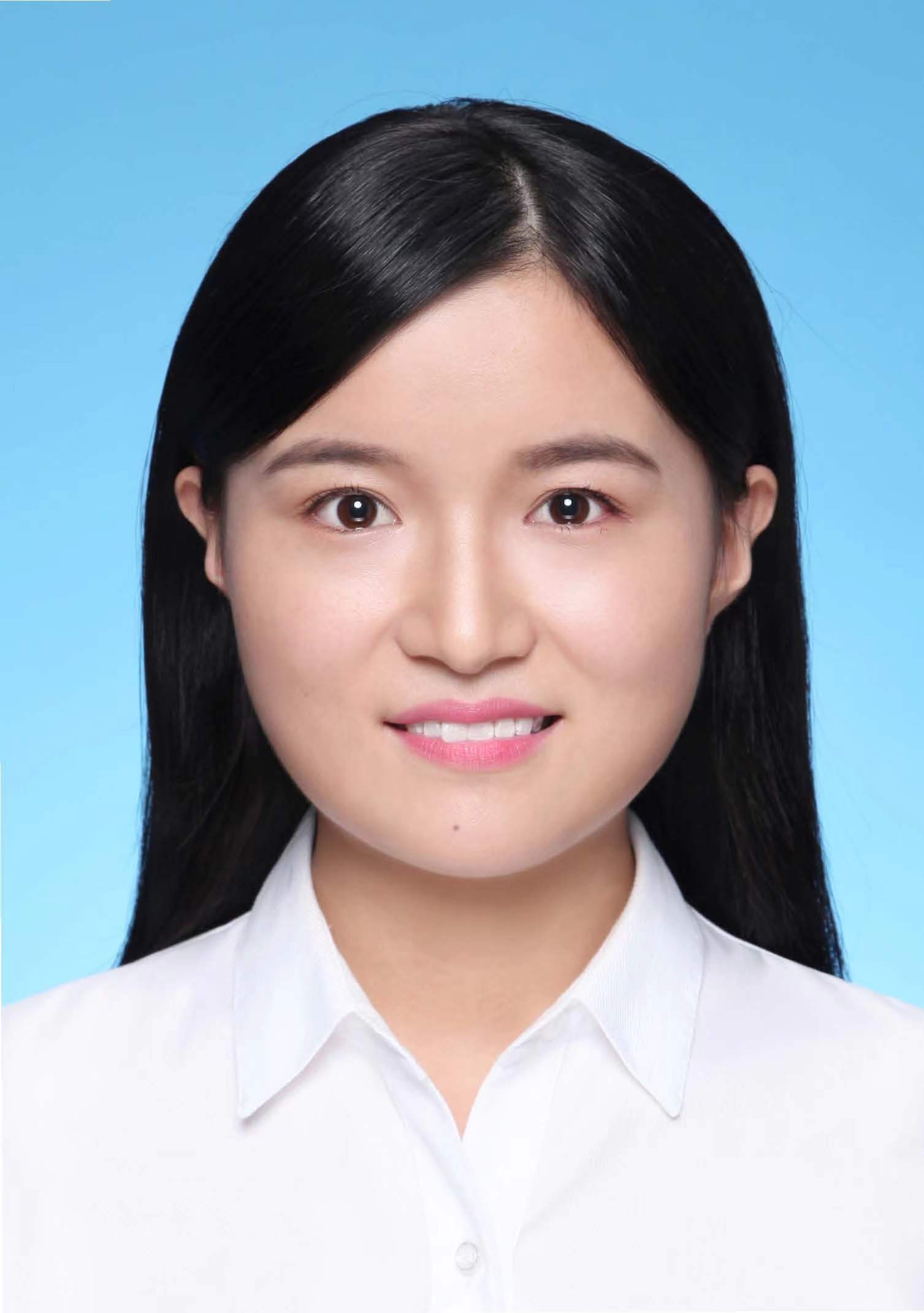}}]{Xingxing Zhang}
received the Ph.D. degree from the Institute of Information Science, Beijing Jiaotong University in 2020 and B.E. degree in 2015. She was also a visiting student with the Department of Computer Science, University of Rochester, USA, from 2018 to 2019. She was a postdoc in the Department of Computer Science, Tsinghua University, from 2020 to 2022. Her research interests include continual learning and zero/few-shot learning. She has received the excellent Ph.D. thesis award from the Chinese Institute of Electronics in 2020.
\end{IEEEbiography}

\vspace{-35pt}
\begin{IEEEbiography}
[{\includegraphics[width=1in,height=1.25in,clip,keepaspectratio]{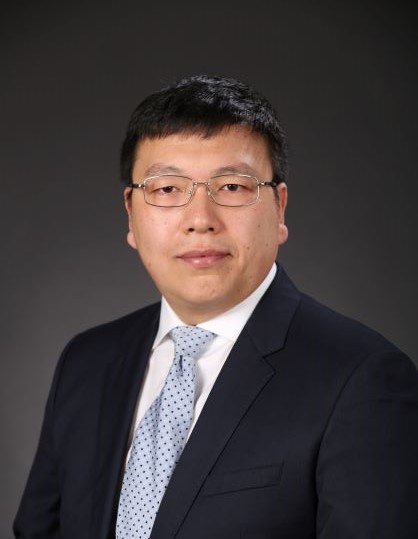}}]{Hang Su}, IEEE member, is an associated professor in the department of computer science and technology at Tsinghua University. His research interests lie in the adversarial machine learning and robust computer vision, based on which he has published more than 50 papers including CVPR, ECCV, TMI, etc. He has served as area chair in NeurIPS and the workshop co-chair in AAAI22. he received ``Young Investigator Award” from MICCAI2012, the ``Best Paper Award'' in AVSS2012, and ``Platinum Best Paper Award'' in ICME2018.
\end{IEEEbiography}

\vspace{-35pt}
\begin{IEEEbiography}
[{\includegraphics[width=1in,height=1.25in,clip,keepaspectratio]{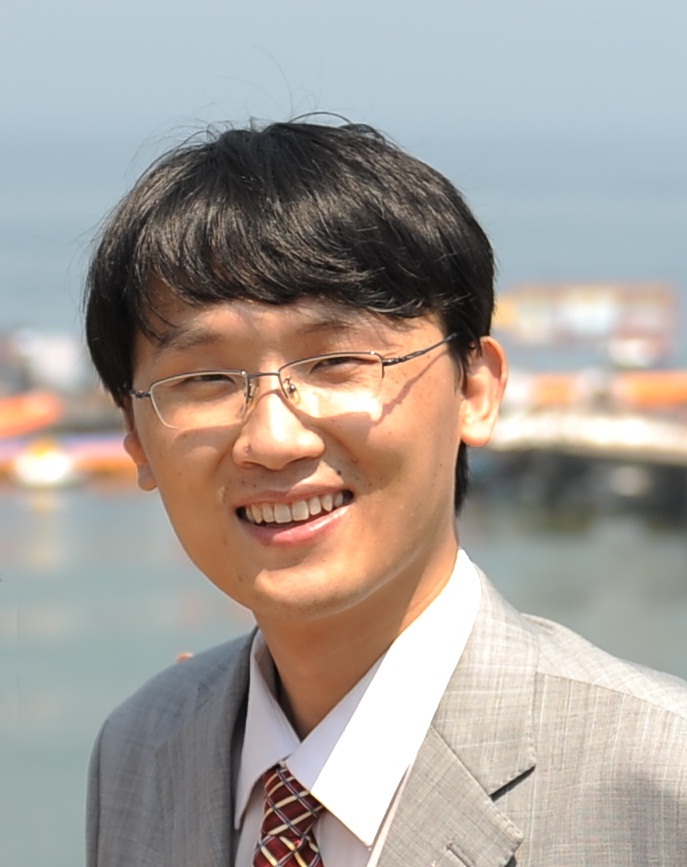}}]{Jun Zhu}  
received his B.S. and Ph.D. degrees from the Department of Computer Science and Technology in Tsinghua University, where he is currently a Bosch AI professor. He was an adjunct faculty and postdoctoral fellow in the Machine Learning Department, Carnegie Mellon University. His research interest is primarily on developing machine learning methods to understand scientific and engineering data arising from various fields. He regularly serves as senior Area Chairs and Area Chairs at prestigious conferences, including ICML, NeurIPS, ICLR, IJCAI and AAAI. He was selected as ``AI's 10 to Watch'' by IEEE Intelligent Systems. He is a Fellow of the IEEE and an associate editor-in-chief of IEEE TPAMI. 
\end{IEEEbiography}





\clearpage
\appendices

\section{Theoretical Foundation I}\label{Appendix_A_theory}
In this section, we present the complete proof of our hierarchical decomposition under different CL scenarios.

\subsection{Class-Incremental Learning (CIL)}

\noindent
\textbf{Proof of Theorem~\ref{LossError1}}\\
For CIL with pre-training, assume $ \mathbb{E}_{\boldsymbol{x}} [{H}_{\rm{WTP}}(\boldsymbol{x})] \leq \delta$, $\mathbb{E}_{\boldsymbol{x}} [{H}_{\rm{TII}}(\boldsymbol{x})] \leq \epsilon$, and $\mathbb{E}_{\boldsymbol{x}} [{H}_{\rm{TAP}}(\boldsymbol{x})] \leq \eta$.
Let $y \in \mathcal{Y}_{\bar{i},\bar{j}}$ be the ground truth of an $\boldsymbol{x}$, where $\bar{i}\in [t]$ and $\bar{j} \in [|\mathcal{Y}_{\bar{i}}|]$ denote the task identity and within-task index, respectively.

As we defined,
\begin{small}
\begin{align}\label{wtp}
\begin{split}
   {H}_{\rm{WTP}}(\boldsymbol{x}) &= \mathcal{H}(\boldsymbol{1}_{\bar{j}},\{P(\boldsymbol{x} \in \mathcal{X}_{\bar{i},j}|\boldsymbol{x} \in \mathcal{X}_{\bar{i}},\mathcal{D},\theta)\}_j)\\ 
   &= -\log P(\boldsymbol{x} \in \mathcal{X}_{\bar{i},\bar{j}}|\boldsymbol{x} \in \mathcal{X}_{\bar{i}},\mathcal{D},\theta),
   \end{split}
\end{align}
\end{small}

\begin{small}
\begin{align}\label{tii}
\begin{split}
    {H}_{\rm{TII}}(\boldsymbol{x}) &= \mathcal{H}(\boldsymbol{1}_{\bar{i}},\{P(\boldsymbol{x} \in \mathcal{X}_{i}|\mathcal{D},\theta)\}_{i})\\
    &=-\log P(\boldsymbol{x} \in \mathcal{X}_{\bar{i}}|\mathcal{D},\theta),
     \end{split}
\end{align}
\end{small}

\begin{small}
\begin{align}\label{tap}
\begin{split}
   {H}_{\rm{TAP}}(\boldsymbol{x}) &= \mathcal{H}(\boldsymbol{1}_{y}, \{P(\boldsymbol{x} \in \mathcal{X}^{c}|\mathcal{D},\theta)\}_{c} )\\
   & = -\log P(\boldsymbol{x} \in \mathcal{X}^{y}|\mathcal{D},\theta).
   \end{split}
\end{align}
\end{small}

Then, we have 
\begin{small}
\begin{align}\label{L1}
\begin{split}
 &\mathcal{H}(\boldsymbol{1}_{\bar{i},\bar{j}},\{P(\boldsymbol{x} \in \mathcal{X}_{{i},{j}}|\mathcal{D},\theta)\}_{{i},{j}})\\
&=-\log P(\boldsymbol{x} \in\mathcal{X}_{\bar{i},\bar{j}}|\mathcal{D},\theta)\\
&=-\log (P(\boldsymbol{x} \in \mathcal{X}_{\bar{i},\bar{j}}|\boldsymbol{x} \in \mathcal{X}_{\bar{i}},\mathcal{D},\theta) P(\boldsymbol{x} \in \mathcal{X}_{\bar{i}}|\mathcal{D},\theta))\\
&=-\log  P(\boldsymbol{x} \in \mathcal{X}_{\bar{i},\bar{j}}|\boldsymbol{x} \in \mathcal{X}_{\bar{i}},\mathcal{D},\theta) -\log P(\boldsymbol{x} \in \mathcal{X}_{\bar{i}}|\mathcal{D},\theta)\\
&= {H}_{\rm{WTP}}(\boldsymbol{x}) + {H}_{\rm{TII}}(\boldsymbol{x}).
 \end{split}
\end{align}
\end{small}

Taking expectations on Eq.~(\ref{tap}), we have
\begin{small}
\begin{align}\label{TAP-E}
\begin{split}
\mathcal {L}_1=\mathbb{E}_{\boldsymbol{x}} [{H}_{\rm{TAP}}(\boldsymbol{x})] \leq \eta.
\end{split}
\end{align}
\end{small}

Taking expectations on both sides of Eq.~(\ref{L1}), we have
\begin{small}
\begin{align}\label{L2}
\begin{split}
\mathcal {L}_2 &= \mathbb{E}_{\boldsymbol{x}} [\mathcal{H}(\boldsymbol{1}_{\bar{i},\bar{j}},\{P(\boldsymbol{x} \in \mathcal{X}_{{i},{j}}|\mathcal{D},\theta)\}_{{i},{j}})]\\
&=\mathbb{E}_{\boldsymbol{x}} [{H}_{\rm{WTP}}(\boldsymbol{x})] +\mathbb{E}_{\boldsymbol{x}} [{H}_{\rm{TII}}(\boldsymbol{x})]\\
&\leq \delta+\epsilon.
\end{split}
\end{align}
\end{small}

Considering the multi-objective optimization problem $\max [P(\boldsymbol{x} \in \mathcal{X}_{\bar{i},\bar{j}}|\mathcal{D},\theta),P(\boldsymbol{x} \in \mathcal{X}^{y}|\mathcal{D},\theta)]$ in Eq.~\eqref{FinalObjective}, we have the loss error 
\begin{small}
\begin{align}\label{Losserror_all}
\begin{split}
\mathcal{L} &= \max\{\mathbb{E}_{\boldsymbol{x}} [\mathcal{H}(\boldsymbol{1}_{\bar{i},\bar{j}},\{P(\boldsymbol{x} \in \mathcal{X}_{{i},{j}}|\mathcal{D},\theta)\}_{{i},{j}})], \mathbb{E}_{\boldsymbol{x}} [{H}_{\rm{TAP}}(\boldsymbol{x})] \}\\
& = \max\{ \mathcal{L}_2,\mathcal{L}_1\}\\
& =  \max\{\delta+\epsilon,\eta \}.
\end{split}
\end{align}
\end{small}
 This finishes the proof.

\noindent
\textbf{Proof of Theorem~\ref{LossError2}}\\
For CIL with pre-training, its loss error $\mathcal{L}\leq \xi $. Assume $\boldsymbol{x} \in \mathcal{X}_{\bar{i},\bar{j}} \subseteq \mathcal{X}_{\bar{i}} $.
 According to the proof of Theorem~\ref{LossError1}, we have
 \begin{small}
\begin{align}\label{wtp-2}
\begin{split}
   {H}_{\rm{WTP}}(\boldsymbol{x}) &= -\log P(\boldsymbol{x} \in \mathcal{X}_{\bar{i},\bar{j}}|\boldsymbol{x} \in \mathcal{X}_{\bar{i}},\mathcal{D},\theta)\\
   & = -\log \frac{P(\boldsymbol{x} \in \mathcal{X}_{\bar{i},\bar{j}}|\mathcal{D},\theta)}{P(\boldsymbol{x} \in \mathcal{X}_{\bar{i}}|\mathcal{D},\theta)}\\
   &\leq -\log P(\boldsymbol{x} \in \mathcal{X}_{\bar{i},\bar{j}}|\mathcal{D},\theta)\\
   & = \mathcal{H}(\boldsymbol{1}_{\bar{i},\bar{j}},\{P(\boldsymbol{x} \in \mathcal{X}_{{i},{j}}|\mathcal{D},\theta)\}_{{i},{j}})\\
   &=\mathcal{L}_2\leq \xi.
   \end{split}
\end{align}
\end{small}

Likewise, we have
\begin{small}
\begin{align}\label{tii-2}
\begin{split}
    {H}_{\rm{TII}}(\boldsymbol{x}) 
    &=-\log P(\boldsymbol{x} \in \mathcal{X}_{\bar{i}}|\mathcal{D},\theta)\\
     & = -\log \frac{P(\boldsymbol{x} \in \mathcal{X}_{\bar{i},\bar{j}}|\mathcal{D},\theta)}{P(\boldsymbol{x} \in \mathcal{X}_{\bar{i},\bar{j}}|\boldsymbol{x} \in \mathcal{X}_{\bar{i}},\mathcal{D},\theta)}\\
    &\leq -\log P(\boldsymbol{x} \in \mathcal{X}_{\bar{i},\bar{j}}|\mathcal{D},\theta)\\
   & = \mathcal{H}(\boldsymbol{1}_{\bar{i},\bar{j}},\{P(\boldsymbol{x} \in \mathcal{X}_{{i},{j}}|\mathcal{D},\theta)\}_{{i},{j}})\\
   & =\mathcal{L}_2 \leq \xi.
     \end{split}
\end{align}
\end{small}

Considering the multi-objective optimization problem $\max [P(\boldsymbol{x} \in \mathcal{X}_{\bar{i},\bar{j}}|\mathcal{D},\theta),P(\boldsymbol{x} \in \mathcal{X}^{y}|\mathcal{D},\theta)]$ in Eq.~\eqref{FinalObjective}, each component must guarantee the loss error less than $\xi$, i.e., 
\begin{small}
\begin{align}\label{tap-2}
\begin{split}
   {H}_{\rm{TAP}}(\boldsymbol{x}) 
   & = -\log P(\boldsymbol{x} \in \mathcal{X}^{y}|\mathcal{D},\theta)\\
   &= \mathcal{L}_1  \leq \xi.
   \end{split}
\end{align}
\end{small}
 This finishes the proof.
 
\subsection{Domain-Incremental Learning (DIL)}

For DIL, let each ``task''\footnote{For naming consistency, here we still use ``task'' to denote the sequentially arriving ``domain'' in DIL.} $\mathcal{D}_i$ consist of domain $\mathcal{X}_i=\bigcup_{j}\mathcal{X}_{i,j}$ 
and label $ \mathcal{Y}_i=\bigcup_{j}\mathcal{Y}_{i,j}$, where $j \in [|\mathcal{Y}_i|]$ denotes the $j$-th class in task $i \in [t]$, and $\mathcal{Y}_i=\mathcal{Y}_{i'}$ for $\forall i \neq i'$. 
Similar to the analysis of CIL, the goal is to learn a projection from $\mathcal{X} = \bigcup_{i=1}^{t} \mathcal{X}_i$ to $\mathcal{Y} = \bigcup_{i=1}^{t} \mathcal{Y}_i$ so as to achieve TAP. When training from scratch, the TAP performance $P(\boldsymbol{x} \in \mathcal{X}^{y}|\mathcal{D})$ is to predict across all classes without distinguishing tasks, where $\mathcal{D} = \{\mathcal{D}_1, ..., \mathcal{D}_{t}\}$, $y \in [|\bigcup_{i=1}^{t} \mathcal{Y}_i|]$ denotes the ground truth label of $\boldsymbol{x}$, and $\mathcal{X}^{y}$ denotes the domain of class $y$. Given the assumptions of disjoint domains, the DIL probability can be expressed as a hierarchical process of TII and WTP:
\begin{small}
\begin{align}\label{BayesTheorem-DIL_Scratch}
   \underbrace{P(\boldsymbol{x} \in \mathcal{X}_{*,j}|\mathcal{D})}_{\text{DIL}} = 
   \sum_{i} 
   \underbrace{P(\boldsymbol{x} \in \mathcal{X}_{i}|\mathcal{D})}_{\text{TII}}
   \underbrace{P(\boldsymbol{x} \in \mathcal{X}_{i,j}|\boldsymbol{x} \in \mathcal{X}_{i},\mathcal{D})}_{\text{WTP}}.
\end{align}
\end{small}
In this case, the TAP performance $P(\boldsymbol{x} \in \mathcal{X}^{y}|\mathcal{D})$ is essentially equivalent to the DIL performance $P(\boldsymbol{x} \in \mathcal{X}_{*,\bar{j}}|\mathcal{D},\theta)$, where $\bar{j} \in [ |\mathcal{Y}_t| ]$ denotes the ground truth of an $\boldsymbol{x}$ w.r.t. the within-task index.

When considering the pre-trained knowledge carried by parameters $\theta$, the TAP is redefined as $P(\boldsymbol{x} \in \mathcal{X}^{y}|\mathcal{D},\theta)$, while the DIL probability of TII and WTP is re-written as
\begin{small}
\begin{align}\label{BayesTheorem-DIL}
   \underbrace{P(\boldsymbol{x} \in \mathcal{X}_{*,j}|\mathcal{D},\theta)}_{\text{DIL}} = 
   \sum_{i} 
   \underbrace{P(\boldsymbol{x} \in \mathcal{X}_{i}|\mathcal{D},\theta)}_{\text{TII}}
   \underbrace{P(\boldsymbol{x} \in \mathcal{X}_{i,j}|\boldsymbol{x} \in \mathcal{X}_{i},\mathcal{D},\theta)}_{\text{WTP}}.
\end{align}
\end{small}
It can be seen that both the TAP performance $P(\boldsymbol{x} \in \mathcal{X}^{y}|\mathcal{D},\theta)$ and the DIL performance $P(\boldsymbol{x} \in \mathcal{X}_{*,\bar{j}}|\mathcal{D},\theta)$ are now affected by $\theta$, but in different ways.
Therefore, we propose to further optimize TAP along with the improved TII and WTP, formulating the ultimate goal of DIL as a multi-objective optimization problem, i.e., 
\begin{small}
\begin{align}\label{FinalObjective}
\max [\, \underbrace{P(\boldsymbol{x} \in \mathcal{X}_{*,\bar{j}}|\mathcal{D},\theta)}_{\text{DIL}}, 
\underbrace{P(\boldsymbol{x} \in \mathcal{X}^{y}|\mathcal{D},\theta)}_{\text{TAP}} \,].
\end{align}
\end{small}
We further derive the following theorems in terms of the sufficient and necessary conditions for improving CL.

\begin{theorem}
    \label{theo3}
    For DIL with pre-training, 
    if $ \mathbb{E}_{\boldsymbol{x}} [{H}_{\rm{WTP}}(\boldsymbol{x})] \leq \delta$, $\mathbb{E}_{\boldsymbol{x}} [{H}_{\rm{TII}}(\boldsymbol{x})] \leq \epsilon$, and $\mathbb{E}_{\boldsymbol{x}} [{H}_{\rm{TAP}}(\boldsymbol{x})] \leq \eta$, we have the loss error $\mathcal{L} \in [0, \max\{{\delta +\epsilon+\log t},\eta\}]$, regardless whether the WTP predictor, TII predictor and TAP predictor are trained together or separately.
\end{theorem}

\noindent
\textbf{Proof of Theorem~\ref{theo3}}\\
As similarly defined in CIL, here
\begin{small}
\begin{align}\label{wtp-DIL}
\begin{split}
   {H}_{\rm{WTP}}(\boldsymbol{x}) &= \mathcal{H}(\boldsymbol{1}_{\bar{j}},\{P(\boldsymbol{x} \in \mathcal{X}_{i,j}|\boldsymbol{x} \in \mathcal{X}_{i},\mathcal{D},\theta)\}_j)\\
   &= -\log P(\boldsymbol{x} \in \mathcal{X}_{i,\bar{j}}|\boldsymbol{x} \in \mathcal{X}_{i},\mathcal{D},\theta),
   \end{split}
\end{align}
\end{small}

\begin{small}
\begin{align}\label{tii-DIL}
\begin{split}
    {H}_{\rm{TII}}(\boldsymbol{x}) &= \mathcal{H}(\gamma,\{P(\boldsymbol{x} \in \mathcal{X}_{i}|\mathcal{D},\theta)\}_{i})\\
    &=-\gamma_i\log P(\boldsymbol{x} \in \mathcal{X}_{i}|\mathcal{D},\theta),
     \end{split}
\end{align}
\end{small}

\begin{small}
\begin{align}\label{tap-DIL}
\begin{split}
   {H}_{\rm{TAP}}(\boldsymbol{x}) &= \mathcal{H}(\boldsymbol{1}_{y}, \{P(\boldsymbol{x} \in \mathcal{X}^{c}|\mathcal{D},\theta)\}_{c} )\\
   & = -\log P(\boldsymbol{x} \in \mathcal{X}^{y}|\mathcal{D},\theta),
   \end{split}
\end{align}
\end{small}
where $\gamma = \{\gamma_i\}_{i=1}^{t}$ represents the possibility of $\boldsymbol{x}$ belonging to each observed domain, $\gamma_i \in [0,1]$ and $\sum_i \gamma_i =1$.

Then, for any simplex $ \gamma$, we have 
\begin{small}
\begin{align}\label{L1-DIL}
\begin{split}
 &\mathcal{H}(\boldsymbol{1}_{\bar{j}},\{P(\boldsymbol{x} \in \mathcal{X}_{*,{j}}|\mathcal{D},\theta)\}_{{j}})\\
&=-\log P(\boldsymbol{x} \in\mathcal{X}_{*,\bar{j}}|\mathcal{D},\theta)\\
&=-\log (\sum_{i} P(\boldsymbol{x} \in \mathcal{X}_{i,\bar {j}}|\boldsymbol{x} \in \mathcal{X}_{i},\mathcal{D},\theta) P(\boldsymbol{x} \in \mathcal{X}_{i}|\mathcal{D},\theta))\\
& \leq -\sum_i \gamma_i \log (\frac{P(\boldsymbol{x} \in \mathcal{X}_{i,\bar {j}}|\boldsymbol{x} \in \mathcal{X}_{i},\mathcal{D},\theta) P(\boldsymbol{x} \in \mathcal{X}_{i}|\mathcal{D},\theta)}{\gamma_i})\\
& = -\sum_i \gamma_i \log P(\boldsymbol{x} \in \mathcal{X}_{i,\bar {j}}|\boldsymbol{x} \in \mathcal{X}_{i},\mathcal{D},\theta)  \\
&-\sum_i \gamma_i \log P(\boldsymbol{x} \in \mathcal{X}_{i}|\mathcal{D},\theta)
+ \sum_i \gamma_i \log(\gamma_i)\\
& =\sum_i \gamma_i  {H}_{\rm{WTP}}  
+ {H}_{\rm{TII}} + \mathcal{H} (\gamma).
\end{split}
\end{align}
\end{small}

Taking expectations on Eq.~(\ref{tap-DIL}), we have
\begin{small}
\begin{align}\label{TAP-E-DIL}
\begin{split}
\mathcal {L}_1=\mathbb{E}_{\boldsymbol{x}} [{H}_{\rm{TAP}}(\boldsymbol{x})] \leq \eta.
\end{split}
\end{align}
\end{small}

Taking expectations on both sides of Eq.~(\ref{L1-DIL}), we have
\begin{small}
\begin{align}\label{L2-DIL}
\begin{split}
\mathcal {L}_2 &= \mathbb{E}_{\boldsymbol{x}} [\mathcal{H}(\boldsymbol{1}_{\bar{j}},\{P(\boldsymbol{x} \in \mathcal{X}_{*,{j}}|\mathcal{D},\theta)\}_{j}]\\
&\leq \sum_i \gamma_i \mathbb{E}_{\boldsymbol{x}} [{H}_{\rm{WTP}}(\boldsymbol{x})] +\mathbb{E}_{\boldsymbol{x}} [{H}_{\rm{TII}}(\boldsymbol{x})] + \mathcal{H} (\gamma)\\
&\leq \delta+\epsilon+\log t.
\end{split}
\end{align}
\end{small}

Considering the multi-objective optimization problem $\max [P(\boldsymbol{x} \in \mathcal{X}_{*,\bar{j}}|\mathcal{D},\theta),P(\boldsymbol{x} \in \mathcal{X}^{y}|\mathcal{D},\theta)]$, we have the loss error 
\begin{small}
\begin{align}\label{Losserror_all-DIL}
\begin{split}
\mathcal{L} &= \max\{\mathbb{E}_{\boldsymbol{x}} [\mathcal{H}(\boldsymbol{1}_{\bar{j}},\{P(\boldsymbol{x} \in \mathcal{X}_{*,{j}}|\mathcal{D},\theta)\}_{{j}})], \mathbb{E}_{\boldsymbol{x}} [{H}_{\rm{TAP}}(\boldsymbol{x})] \}\\
& = \max\{ \mathcal{L}_2,\mathcal{L}_1\}\\
& =  \max\{\delta+\epsilon+\log t,\eta \}.
\end{split}
\end{align}
\end{small}
 This finishes the proof.

\begin{theorem}
    \label{theo4}
     For DIL with pre-training, if the loss error $\mathcal{L}  \leq \xi $, then there always exist (1) a WTP predictor, s.t. ${H}_{\rm{WTP}} \leq \xi$; (2) a TII predictor, s.t. ${H}_{\rm{TII}} \leq \xi$; and (3) a TAP predictor, s.t. ${H}_{\rm{TAP}} \leq \xi$.
\end{theorem}

\noindent
\textbf{Proof of Theorem~\ref{theo4}}\\
For DIL with pre-training, its loss error $\mathcal{L}= \max [\mathcal{L}_1, \mathcal{L}_2]\leq \xi $.
 Assume $\boldsymbol{x} \in \mathcal{X}_{*,\bar{j}} \subseteq \mathcal{X}^{y} $.
 According to the proof of Theorem~\ref{theo3}, if we define 
 $P(\boldsymbol{x} \in \mathcal{X}_{i,\bar{j}}|\mathcal{D},\theta) = P(\boldsymbol{x} \in \mathcal{X}_{*,\bar{j}}|\mathcal{D},\theta)$, we have
 \begin{small}
\begin{align}\label{wtp-2-DIL}
\begin{split}
   {H}_{\rm{WTP}}(\boldsymbol{x}) &= -\log P(\boldsymbol{x} \in \mathcal{X}_{i,\bar{j}}|\boldsymbol{x} \in \mathcal{X}_{i},\mathcal{D},\theta)\\
   & = -\log \frac{P(\boldsymbol{x} \in \mathcal{X}_{i,\bar{j}}|\mathcal{D},\theta)}{P(\boldsymbol{x} \in \mathcal{X}_{i}|\mathcal{D},\theta)}\\
   &\leq -\log P(\boldsymbol{x} \in \mathcal{X}_{i,\bar{j}}|\mathcal{D},\theta)\\
   &= -\log P(\boldsymbol{x} \in \mathcal{X}_{*,\bar{j}}|\mathcal{D},\theta)\\
   & = \mathcal{H}(\boldsymbol{1}_{\bar{j}},\{P(\boldsymbol{x} \in \mathcal{X}_{*,{j}}|\mathcal{D},\theta)\}_{{j}})\\
   &= \mathcal{L}_2 \leq \xi.
   \end{split}
\end{align}
\end{small}

Likewise, if we define $\gamma_i=1$ and $\gamma_{i'}=0$ for $\forall i \neq i', i \in [t]$, we have
\begin{small}
\begin{align}\label{tii-2-DIL}
\begin{split}
    {H}_{\rm{TII}}(\boldsymbol{x}) 
    &=- \sum_i \gamma_i \log P(\boldsymbol{x} \in \mathcal{X}_{{i}}|\mathcal{D},\theta)\\
    & = -\log P(\boldsymbol{x} \in \mathcal{X}_{{i}}|\mathcal{D},\theta)\\
    & = - \log \frac{P(\boldsymbol{x} \in \mathcal{X}_{i,\bar{j}}|\mathcal{D},\theta)}{P(\boldsymbol{x} \in \mathcal{X}_{i,\bar{j}}|\boldsymbol{x} \in \mathcal{X}_{i},\mathcal{D},\theta)}\\
    & \leq - \log (\boldsymbol{x} \in \mathcal{X}_{i,\bar{j}}|\mathcal{D},\theta)\\
    & =  - \log (\boldsymbol{x} \in \mathcal{X}_{*,\bar{j}}|\mathcal{D},\theta)\\
    & = \mathcal{H}(\boldsymbol{1}_{\bar{j}},\{P(\boldsymbol{x} \in \mathcal{X}_{*,{j}}|\mathcal{D},\theta)\}_{{j}})\\
    &= \mathcal{L}_2 \leq \xi.
      \end{split}
\end{align}
\end{small}

Considering the multi-objective optimization problem $\max [P(\boldsymbol{x} \in \mathcal{X}_{*,\bar{j}}|\mathcal{D},\theta),P(\boldsymbol{x} \in \mathcal{X}^{y}|\mathcal{D},\theta)]$, each component must guarantee the loss error less than $\xi$, i.e., 

\begin{small}
\begin{align}
\begin{split}
   {H}_{\rm{TAP}}(\boldsymbol{x}) 
   & = -\log P(\boldsymbol{x} \in \mathcal{X}^{y}|\mathcal{D},\theta)\\
      & = \mathcal{L}_1 \leq \xi.
   \end{split}
\end{align}
\end{small}
 This finishes the proof.

\subsection{Task-Incremental Learning (TIL)}
For task-incremental learning (TIL), let each task $\mathcal{D}_i$ consist of domain $\mathcal{X}_i=\bigcup_{j}\mathcal{X}_{i,j}$ 
and label $\mathcal{Y}_i=\bigcup_{j}\mathcal{Y}_{i,j}$, where $j \in [|\mathcal{Y}_i|]$ denotes the $j$-th class in task $i \in [t]$. Unlike CIL and DIL, TIL has the task identity provided during the testing phase. Whether or not the impact of pre-trained parameters $\theta$ is taken into account, the TAP objective is to learn $P(\boldsymbol{x} \in \mathcal{X}_{\bar{i},j}|\boldsymbol{x} \in \mathcal{X}_{\bar{i}}, \mathcal{D})$ or $P(\boldsymbol{x} \in \mathcal{X}_{\bar{i},j}|\boldsymbol{x} \in \mathcal{X}_{\bar{i}}, \mathcal{D},\theta)$, where $\mathcal{D} = \{\mathcal{D}_1, ..., \mathcal{D}_{t}\}$, $\bar{i}\in [t]$, and $j \in [|\mathcal{Y}_{\bar{i}}|]$. In fact, this is equivalent to WTP alone. For completeness, we further derive the following theorems in terms of the sufficient and necessary conditions for improving CL.

\begin{theorem}
    \label{theo5}
    For TIL with pre-training, $\mathbb{E}_{\boldsymbol{x}} [{H}_{\rm{TII}}(\boldsymbol{x})] =0$, and TAP is degraded into WTP.
    If $ \mathbb{E}_{\boldsymbol{x}} [{H}_{\rm{WTP}}(\boldsymbol{x})] \leq \delta$, we have the loss error $\mathcal{L} \in [0, \delta ]$.
\end{theorem}

\noindent
\textbf{Proof of Theorem~\ref{theo5}}\\
For TIL with pre-training, assume $ \mathbb{E}_{\boldsymbol{x}} [{H}_{\rm{WTP}}(\boldsymbol{x})] \leq \delta$.
Given an $\boldsymbol{x}$ with the task identity $\bar{i} \in [t]$, let $\bar{j} \in [|\mathcal{Y}_{\bar{i}}|]$ be the ground truth of $\boldsymbol{x}$ w.r.t. the within-task index, and $y \in [|\bigcup_{i=1}^{t} \mathcal{Y}_i|]$ be the ground truth label of $\boldsymbol{x}$.

As similarly defined in CIL, here
\begin{small}
\begin{align}\label{wtp-TIL}
\begin{split}
   {H}_{\rm{WTP}}(\boldsymbol{x}) &= \mathcal{H}(\boldsymbol{1}_{\bar{j}},\{P(\boldsymbol{x} \in \mathcal{X}_{\bar{i},j}|\boldsymbol{x} \in \mathcal{X}_{\bar{i}},\mathcal{D},\theta)\}_j)\\ 
   &= -\log P(\boldsymbol{x} \in \mathcal{X}_{\bar{i},\bar{j}}|\boldsymbol{x} \in \mathcal{X}_{\bar{i}},\mathcal{D},\theta),
   \end{split}
\end{align}
\end{small}

\begin{small}
\begin{align}\label{tii-TIL}
\begin{split}
    {H}_{\rm{TII}}(\boldsymbol{x}) &= \mathcal{H}(\boldsymbol{1}_{\bar{i}},\{P(\boldsymbol{x} \in \mathcal{X}_{i}|\mathcal{D},\theta)\}_{i})\\
    &=-\log P(\boldsymbol{x} \in \mathcal{X}_{\bar{i}}|\mathcal{D},\theta)\\
    &=-\log 1 \\
    &=0,
     \end{split}
\end{align}
\end{small}

\begin{small}
\begin{align}\label{tap-TIL}
\begin{split}
   {H}_{\rm{TAP}}(\boldsymbol{x}) &= \mathcal{H}(\boldsymbol{1}_{y}, \{P(\boldsymbol{x} \in \mathcal{X}^{c}|\boldsymbol{x} \in \mathcal{X}_{\bar{i}},\mathcal{D},\theta)\}_{c})\\
   & = -\log P(\boldsymbol{x} \in \mathcal{X}^{y}|\boldsymbol{x} \in \mathcal{X}_{\bar{i}},\mathcal{D},\theta)\\
   &={H}_{\rm{WTP}}(\boldsymbol{x}).
   \end{split}
\end{align}
\end{small}

Then, we have 
\begin{small}
\begin{align}\label{L1-TIL}
\begin{split}
 &\mathcal{H}(\boldsymbol{1}_{\bar{i},\bar{j}},\{P(\boldsymbol{x} \in \mathcal{X}_{{i},{j}}|\mathcal{D},\theta)\}_{{i},{j}})\\
&=-\log P(\boldsymbol{x} \in\mathcal{X}_{\bar{i},\bar{j}}|\mathcal{D},\theta)\\
&=-\log (P(\boldsymbol{x} \in \mathcal{X}_{\bar{i},\bar{j}}|\boldsymbol{x} \in \mathcal{X}_{\bar{i}},\mathcal{D},\theta) P(\boldsymbol{x} \in \mathcal{X}_{\bar{i}}|\mathcal{D},\theta))\\
&=-\log  P(\boldsymbol{x} \in \mathcal{X}_{\bar{i},\bar{j}}|\boldsymbol{x} \in \mathcal{X}_{\bar{i}},\mathcal{D},\theta) -\log P(\boldsymbol{x} \in \mathcal{X}_{\bar{i}}|\mathcal{D},\theta)\\
&= {H}_{\rm{WTP}}(\boldsymbol{x}) + {H}_{\rm{TII}}(\boldsymbol{x})\\
& = {H}_{\rm{WTP}}(\boldsymbol{x}).
 \end{split}
\end{align}
\end{small}

Taking expectations on both sides of Eq.~(\ref{L1-TIL}), we have
\begin{small}
\begin{align}\label{L-TIL}
\begin{split}
\mathcal {L} &= \mathbb{E}_{\boldsymbol{x}} [\mathcal{H}(\boldsymbol{1}_{\bar{i},\bar{j}},\{P(\boldsymbol{x} \in \mathcal{X}_{{i},{j}}|\mathcal{D},\theta)\}_{{i},{j}})]\\
&=\mathbb{E}_{\boldsymbol{x}} [{H}_{\rm{WTP}}(\boldsymbol{x})] \\
&\leq \delta.
\end{split}
\end{align}
\end{small}

Considering the TIL objective $P(\boldsymbol{x} \in \mathcal{X}_{\bar{i},\bar{j}}|\boldsymbol{x} \in \mathcal{X}_{\bar{i}},\mathcal{D},\theta)$, we have the loss error $\mathcal {L} \leq \delta$.
This finishes the proof.
 
\begin{theorem}
    \label{theo6}
     For TIL with pre-training, if the loss error $\mathcal{L}\leq \xi $, then there always exists a WTP predictor, s.t. ${H}_{\rm{WTP}} \leq \xi$.
\end{theorem}

\noindent
\textbf{Proof of Theorem~\ref{theo6}}\\
For TIL with pre-training, its loss error $\mathcal{L}\leq \xi $. Assume $\boldsymbol{x} \in \mathcal{X}_{\bar{i},\bar{j}} \subseteq \mathcal{X}_{\bar{i}} $.
According to the proof of Theorem~\ref{theo5}, we have
 \begin{small}
\begin{align}
\begin{split}
   {H}_{\rm{WTP}}(\boldsymbol{x}) &= -\log P(\boldsymbol{x} \in \mathcal{X}_{\bar{i},\bar{j}}|\boldsymbol{x} \in \mathcal{X}_{\bar{i}},\mathcal{D},\theta)\\
   & = -\log \frac{P(\boldsymbol{x} \in \mathcal{X}_{\bar{i},\bar{j}}|\mathcal{D},\theta)}{P(\boldsymbol{x} \in \mathcal{X}_{\bar{i}}|\mathcal{D},\theta)}\\
   &\leq -\log P(\boldsymbol{x} \in \mathcal{X}_{\bar{i},\bar{j}}|\mathcal{D},\theta)\\
   & = \mathcal{H}(\boldsymbol{1}_{\bar{i},\bar{j}},\{P(\boldsymbol{x} \in \mathcal{X}_{{i},{j}}|\mathcal{D},\theta)\}_{{i},{j}})\\
   &\leq \xi.
   \end{split}
\end{align}
\end{small}
This finishes the proof.

\vspace{-0.2cm}
\section{Theoretical Foundation II}\label{LossOODProof}
In this section, we first present the complete proof of connecting TII to OOD detection, and then derive the sufficient and necessary conditions of improving CL with WTP, OOD detection and TAP.
\vspace{-0.2cm}

\subsection{TII to OOD Detection}
\noindent
\textbf{Proof of Theorem~\ref{LossOOD1}}\\
For CL in a pre-training context, define the TII probability as $P(\boldsymbol{x} \in \mathcal{X}_{i}|\mathcal{D},\theta)=\frac{P_i(\boldsymbol{x} \in \mathcal{X}_{i}|\mathcal{D},\theta)}{\sum_{j}P_j(\boldsymbol{x} \in \mathcal{X}_{j}|\mathcal{D},\theta)}$.
If $H_{{\rm{OOD}},i}(\boldsymbol{x}) \leq \epsilon_i$ for $i \in [t]$, then we have 
\begin{small}
\begin{equation}
\begin{split}
&H_{{\rm{OOD}},i}(\boldsymbol{x})= \\
&\begin{cases}
\mathcal{H}(1, P_i(\boldsymbol{x} \in \mathcal{X}_{i}|\mathcal{D},\theta))=-\log P_i(\boldsymbol{x} \in \mathcal{X}_{i}|\mathcal{D},\theta)\leq \epsilon_i, & \boldsymbol{x} \in \mathcal{X}_{i}\\ 
\mathcal{H}(0, P_i(\boldsymbol{x} \in \mathcal{X}_{i}|\mathcal{D},\theta))=-\log P_i(\boldsymbol{x} \notin \mathcal{X}_{i}|\mathcal{D},\theta)\leq \epsilon_i, & \boldsymbol{x} \notin \mathcal{X}_{i} \\
\end{cases}.
\end{split}
\end{equation}
\end{small}
This means $P_i(\boldsymbol{x} \in \mathcal{X}_{i}|\mathcal{D},\theta) \geq e^{-\epsilon_i}$ for $\boldsymbol{x} \in \mathcal{X}_{i}$, and $P_i(\boldsymbol{x} \notin \mathcal{X}_{i}|\mathcal{D},\theta) \geq  e^{-\epsilon_i}$ (i.e., $P_i(\boldsymbol{x} \in \mathcal{X}_{i}|\mathcal{D},\theta) \leq  1-e^{-\epsilon_i}$) for $\boldsymbol{x} \notin \mathcal{X}_{i} $. 

Let $\bar{i}\in [t]$ denote the task identity of $\boldsymbol{x}$, then we have
\begin{small}
\begin{align}\label{tii-ood}
\begin{split}
{H}_{\rm{TII}}(\boldsymbol{x}) &= \mathcal{H}(\boldsymbol{1}_{\bar{i}},\{P(\boldsymbol{x} \in \mathcal{X}_{i}|\mathcal{D},\theta)\}_{i})\\
    &=-\log P(\boldsymbol{x} \in \mathcal{X}_{\bar{i}}|\mathcal{D},\theta)\\
    &=-\log \frac{P_{\bar{i}}(\boldsymbol{x} \in \mathcal{X}_{\bar{i}}|\mathcal{D},\theta)}{\sum_{j} P_j(\boldsymbol{x} \in \mathcal{X}_{j}|\mathcal{D},\theta)}\\
    &=\log  [1+\frac{\sum_{j\neq {\bar i}} P_{j}(\boldsymbol{x} \in \mathcal{X}_{j}|\mathcal{D},\theta)}{P_{\bar{i}}(\boldsymbol{x} \in \mathcal{X}_{\bar{i}}|\mathcal{D},\theta)}]\\
    & \leq \log [1+ \frac{\sum_{j\neq {\bar i}} 1-e^{-\epsilon_j}}{e^{-\epsilon_{\bar{i}}}}]\\
    & = \log [1+{e^{\epsilon_{\bar{i}}}}\sum_{j\neq {\bar i}} 1-e^{-\epsilon_j}]\\
    & \leq {e^{\epsilon_{\bar{i}}}}\sum_{j\neq {\bar i}} 1-e^{-\epsilon_j}\\
    &=(\sum_i \boldsymbol{1}_{\boldsymbol{x}\in \mathcal{X}_{i}}e^{\epsilon_i})(\sum_i \boldsymbol{1}_{\boldsymbol{x}\notin \mathcal{X}_{i}}(1-e^{-\epsilon_i})).
\end{split}
\end{align}
\end{small}
The last inequation holds due to $\log(1+z)\leq z$ for $z\geq 0$.

Now, let us move on the proof of adequacy for TII and OOD detection. If $H_{\rm{TII}}(\boldsymbol{x})\leq \epsilon$, then we have
\begin{small}
\begin{equation}
\begin{split}
{H}_{\rm{TII}}(\boldsymbol{x}) &= \mathcal{H}(\boldsymbol{1}_{\bar{i}},\{P(\boldsymbol{x} \in \mathcal{X}_{i}|\mathcal{D},\theta)\}_{i})\\
    &=-\log P(\boldsymbol{x} \in \mathcal{X}_{\bar{i}}|\mathcal{D},\theta)\\
    &\leq \epsilon.
 \end{split}
\end{equation}
\end{small}

Further, for $P(\boldsymbol{x} \in \mathcal{X}_{i}|\mathcal{D},\theta)=\frac{P_i(\boldsymbol{x} \in \mathcal{X}_{i}|\mathcal{D},\theta)}{\sum_{j}P_j(\boldsymbol{x} \in \mathcal{X}_{j}|\mathcal{D},\theta)}$, we have 
\begin{small}
\begin{equation}
\begin{split}
&H_{{\rm{OOD}},i}(\boldsymbol{x})= \mathcal{H}(1, P_i(\boldsymbol{x} \in \mathcal{X}_{i}|\mathcal{D},\theta))\\
&=-\log P_i(\boldsymbol{x} \in \mathcal{X}_{i}|\mathcal{D},\theta)\\
&=-\log (P(\boldsymbol{x} \in \mathcal{X}_{i}|\mathcal{D},\theta) {\sum_{j}P_j(\boldsymbol{x} \in \mathcal{X}_{j}|\mathcal{D},\theta)})\\
& = -\log P(\boldsymbol{x} \in \mathcal{X}_{i}|\mathcal{D},\theta) -\log{\sum_{j}P_j(\boldsymbol{x} \in \mathcal{X}_{j}|\mathcal{D},\theta)}\\
& ={H}_{\rm{TII}}(\boldsymbol{x})-\log{\sum_{j}P_j(\boldsymbol{x} \in \mathcal{X}_{j}|\mathcal{D},\theta)}\\
&\leq \epsilon
\end{split}
\end{equation}
\end{small}
The last inequation holds due to $\sum_{j}P_j(\boldsymbol{x} \in \mathcal{X}_{j}|\mathcal{D},\theta)\geq 1$.

Likewise, for $\boldsymbol{x} \notin \mathcal{X}_{i} $, we have
\begin{small}
\begin{equation}
\begin{split}
&H_{{\rm{OOD}},i}(\boldsymbol{x})= \mathcal{H}(0, P_i(\boldsymbol{x} \in \mathcal{X}_{i}|\mathcal{D},\theta))\\
&=-\log P_i(\boldsymbol{x} \notin \mathcal{X}_{i}|\mathcal{D},\theta)\\
&=-\log (1-P_i(\boldsymbol{x} \in \mathcal{X}_{i}|\mathcal{D},\theta))\\
& =-\log (1-P(\boldsymbol{x} \in \mathcal{X}_{i}|\mathcal{D},\theta) {\sum_{j}P_j(\boldsymbol{x} \in \mathcal{X}_{j}|\mathcal{D},\theta)})\\
& \leq -\log P(\boldsymbol{x} \in \mathcal{X}_{\bar{i}}|\mathcal{D},\theta)\\
&\leq \epsilon 
 \end{split}
\end{equation}
\end{small}
This finishes the proof.

\subsection{Sufficient and Necessary Conditions}
Now we discuss the upper bound of CIL in relation to WTP, OOD detection and TAP. 
\begin{theorem}
    \label{LossOOD2}
    For CIL with pre-training (i.e., $\theta$), define the TII probability as $P(\boldsymbol{x} \in \mathcal{X}_{i}|\mathcal{D},\theta)=\frac{P_i(\boldsymbol{x} \in \mathcal{X}_{i}|\mathcal{D},\theta)}{\sum_{j}P_j(\boldsymbol{x} \in \mathcal{X}_{j}|\mathcal{D},\theta)}$.
     If $ {H}_{\rm{WTP}}(\boldsymbol{x}) \leq \delta$, ${H}_{\rm{TAP}}(\boldsymbol{x}) \leq \eta$, and
     $H_{{\rm{OOD}},i}(\boldsymbol{x})  \leq \epsilon_i$ for $i \in [t]$, then we have the loss error 
$$\mathcal{L} \in [0, \max\{{\delta +(\sum_i \boldsymbol{1}_{\boldsymbol{x}\in \mathcal{X}_{i}}e^{\epsilon_i})(\sum_i \boldsymbol{1}_{\boldsymbol{x}\notin \mathcal{X}_{i}}(1-e^{-\epsilon_i}))},\eta\}].$$
\end{theorem}

As shown in Theorem~\ref{LossOOD2}, the good performance of WTP, TAP, and OOD detection are sufficient to guarantee a good CIL performance. Now we further study the necessary conditions of a well-performed CIL model.
\begin{theorem}
    \label{LossOOD3}
 For CIL with pre-training (i.e., $\theta$), define the TII probability as $P(\boldsymbol{x} \in \mathcal{X}_{i}|\mathcal{D},\theta)=\frac{P_i(\boldsymbol{x} \in \mathcal{X}_{i}|\mathcal{D},\theta)}{\sum_{j}P_j(\boldsymbol{x} \in \mathcal{X}_{j}|\mathcal{D},\theta)}$. If the loss error $\mathcal{L}\leq \xi $, then there always exist (1) a WTP predictor, s.t. ${H}_{\rm{WTP}} \leq \xi$; (2) a TII predictor, s.t. ${H}_{\rm{TII}} \leq \xi$; (3) a TAP predictor, s.t. ${H}_{\rm{TAP}} \leq \xi$; and (4) an OOD detector for each task, s.t. ${H}_{{\rm{OOD}},i} \leq \xi$ for $i \in [t]$.
\end{theorem}

This theorem shows that if a good CIL model is trained, then a good WTP, a good TII, a good TAP, and a good OOD detector for each task are always implied.

\noindent
\textbf{Proof of Theorem~\ref{LossOOD2}}\\
For CIL with pre-training (i.e., $\theta$), define TII as $P(\boldsymbol{x} \in \mathcal{X}_{i}|\mathcal{D},\theta)=\frac{P_i(\boldsymbol{x} \in \mathcal{X}_{i}|\mathcal{D},\theta)}{\sum_{j}P_j(\boldsymbol{x} \in \mathcal{X}_{j}|\mathcal{D},\theta)}$.
     If $ {H}_{\rm{WTP}}(\boldsymbol{x}) \leq \delta$, ${H}_{\rm{TAP}}(\boldsymbol{x}) \leq \eta$, and
     $H_{{\rm{OOD}},i}(\boldsymbol{x})  \leq \epsilon_i$ for $i \in [t]$, then using Theorem~\ref{LossOOD2} we have $H_{\rm{TII}}(\boldsymbol{x})\leq (\sum_i \boldsymbol{1}_{\boldsymbol{x}\in \mathcal{X}_{i}}e^{\epsilon_i})(\sum_i \boldsymbol{1}_{\boldsymbol{x}\notin \mathcal{X}_{i}}(1-e^{-\epsilon_i}))$. 
     Further, using Theorem~\ref{LossError1} we have the loss error 
\begin{small}
\begin{equation}\label{cil-ood1}
\begin{split}
\mathcal{L} &= \max\{\mathbb{E}_{\boldsymbol{x}} [\mathcal{H}(\boldsymbol{1}_{\bar{i},\bar{j}},\{P(\boldsymbol{x} \in \mathcal{X}_{{i},{j}}|\mathcal{D},\theta)\}_{{i},{j}})], \mathbb{E}_{\boldsymbol{x}} [{H}_{\rm{TAP}}(\boldsymbol{x})] \}\\
& = \max\{ \mathbb{E}_{\boldsymbol{x}} [{H}_{\rm{WTP}}(\boldsymbol{x})] +\mathbb{E}_{\boldsymbol{x}} [{H}_{\rm{TII}}(\boldsymbol{x})],\mathcal{L}_1\}\\
& =  \max\{\delta+(\sum_i \boldsymbol{1}_{\boldsymbol{x}\in \mathcal{X}_{i}}e^{\epsilon_i})(\sum_i \boldsymbol{1}_{\boldsymbol{x}\notin \mathcal{X}_{i}}(1-e^{-\epsilon_i})),\eta \}.
\end{split}
\end{equation}
\end{small}
This finishes the proof.

\noindent
\textbf{Proof of Theorem~\ref{LossOOD3}}\\
For CIL with pre-training (i.e., $\theta$), define the TII probability as $P(\boldsymbol{x} \in \mathcal{X}_{i}|\mathcal{D},\theta)=\frac{P_i(\boldsymbol{x} \in \mathcal{X}_{i}|\mathcal{D},\theta)}{\sum_{j}P_j(\boldsymbol{x} \in \mathcal{X}_{j}|\mathcal{D},\theta)}$. If the loss error $\mathcal{L}\leq \xi $, then using Theorem~\ref{LossError2} there always exist (1) a WTP predictor, s.t. ${H}_{\rm{WTP}} \leq \xi$; (2) a TII predictor, s.t. ${H}_{\rm{TII}} \leq \xi$; and (3) a TAP predictor, s.t. ${H}_{\rm{TAP}} \leq \xi$.
Furthermore, if ${H}_{\rm{TII}} \leq \xi$, using Theorem~\ref{LossOOD1}, we always have an OOD detector for each task, s.t. ${H}_{{\rm{OOD}},i} \leq \xi$ for $i \in [t]$. This finishes the proof.

\begin{table*}[th]
    \renewcommand\arraystretch{1.8}
     \vspace{-0.1cm}
     \caption{Comparison of recent CL methods relevant to PTMs and PET. $t$ is the total number of tasks. $d$ is the embedding dimension. $s$ is the expansion rate of embedding dimension. Full: fine-tuning of full backbone parameters. General: applicable to mainstream PET techniques, such as ProT, PreT, LoRA, Adapter, etc. Note that $t << d$ in general, e.g., $t=10$ and $d=768$ for all cases in this work. $s=100$ in \cite{mcdonnell2023ranpac}.} 
     \vspace{-0.2cm}
       \centering
      \smallskip
       \resizebox{16.5 cm}{!}{ 
      \begin{tabular}{l|c|c|c|c|c|c}
	 \hline
      Method & Year & Avenue & PET Technique & Task-Specific Parameters & Task-Shared Parameters & Representation Recovery \\
      \hline
       L2P \cite{wang2022learning_l2p} & 2022 & CVPR & ProT & \ding{51} & \ding{51} & N/A \\
       DualPrompt \cite{wang2022dualprompt} & 2022 & ECCV & PreT & \ding{51} & \ding{51} & N/A \\
       S-Prompt \cite{wang2022sprompts} & 2022 & NeurIPS & ProT & \ding{51} & N/A & N/A \\
       CODA-Prompt \cite{smith2023coda} & 2023 & CVPR & PreT & \ding{51} & N/A & N/A \\
       SLCA \cite{zhang2023slca} & 2023 & ICCV & Full & N/A & \ding{51} & $O(td^2)$ \\
       FSA \cite{panos2023first} & 2023 & ICCV & Full & N/A & \ding{51} & $O(d^2)$  \\
       LAE \cite{gao2023unified} & 2023 & ICCV & General & N/A & \ding{51} & N/A \\
       RanPAC \cite{mcdonnell2023ranpac} & 2023 & NeurIPS & PreT & N/A & \ding{51} & $O(s^2d^2)$ \\
       KOPPA \cite{tran2023koppa} & 2023 & arXiv & PreT & \ding{51} & N/A & $O(td)$ \\
       H-Prompt \cite{zuo2024hierarchical} & 2024 & arXiv & PreT & \ding{51} & \ding{51} & $O(td^2)$ \\
       \cdashline{1-7}[2pt/2pt]
       HiDe-Prompt \cite{wang2023hierarchical} & 2023 & NeurIPS & PreT & \ding{51} & N/A & $O(td^2)$ \\
       HiDe-PET & 2024 & Current & General & \ding{51} & \ding{51} & $O(td)$ \\
      \hline
	\end{tabular}
	}
 
	\label{table:survey}
    \vspace{-.1cm}
\end{table*}

\section{Implementation Details}\label{implementation}

Here we describe the supplementary implementation details of the empirical investigation. 

\textbf{Comparison with Preliminary Version:} 
The major technical difference between our HiDe-PET and our preliminary version \cite{wang2023hierarchical} lies in the use of task-shared parameters $\boldsymbol{g}$ to improve TII, which is critical for LoRA/Adapter-based PET that is sensitive to the TII errors. To mitigate catastrophic forgetting in $\boldsymbol{g}$, we set a cosine-decaying learning rate of 0.01 for FSA, a cosine-decaying learning rate of 0.001 for SL, and  a momentum of 0.1 for EMA. The PET ensemble strategy sets $\alpha=0.1$ in all cases. To ensure generality and resource efficiency, the specific implementations are slightly modified in three aspects. 
First, our preliminary version \cite{wang2023hierarchical} followed the implementation of L2P \cite{wang2022learning_l2p} and DualPrompt \cite{wang2022dualprompt}, which employed a constant learning rate of 0.005 and a supervised checkpoint of ImageNet-21K (i.e., Sup-21K). We notice that many recent methods followed the implementation of CODA-Prompt \cite{smith2023coda}, which employed a cosine-decaying learning rate of 0.001, a self-supervised/supervised checkpoint on ImageNet-21/1K (i.e., Sup-21/1K) and a different split of ImageNet-R. Considering that a smaller learning rate with cosine decay has been more commonly used for fine-tuning large-scale PTMs, we reproduce all baselines with the implementation of CODA-Prompt \cite{smith2023coda} in the current manuscript. This consideration further ensures the generality of our HiDe-PET in adapting to different experimental settings.
Second, our preliminary version \cite{wang2023hierarchical} devised a contrastive regularization (CR) term to balance the instructed representations for WTP and TAP, which brings some benefits to the performance of Prompt-based PET. In subsequent explorations, we observe that the CR term cannot improve the performance of LoRA/Adapter-based PET, and therefore remove it in the current manuscript to ensure generality in adapting to different PET techniques.
Third, our preliminary version \cite{wang2023hierarchical} employed dedicated covariance matrices (additional $d^2$ parameters for each class) in representation recovery and a two-layer MLP (additional $d^2$ parameters for the first layer) in $\hat{h}_{\omega}$, in order to acquire better performance. In contrast, the current manuscript employs multi-centroid (additional $<10d$ parameters for each class) in representation recovery and a one-layer MLP in $\hat{h}_{\omega}$, which slightly compromise the performance but largely improve resource efficiency. 

\textbf{Adaptive Knowledge Accumulation:}
In Sec.~\ref{sec:ood_method}, we devise a PET hierarchy $\boldsymbol{g}_{1},...,\boldsymbol{g}_{k}$ inspired by OOD detection to demonstrate the connections between task-specific and task-shared PET architecture. 
We further consider a specialized implementation of $\boldsymbol{g}_{1},...,\boldsymbol{g}_{k}$ for HiDe-PET in Algorithm~\ref{alg:algorithm}, serving as a plug-in module to achieve adaptive knowledge accumulation from pronounced distribution changes. 
As described in Sec.~\ref{sec:ood_method}, $\boldsymbol{g}_{1},...,\boldsymbol{g}_{k}$ are adaptively expanded or retrieved upon the distance ${\rm{Dis}}(\boldsymbol{x}, \hat{\mathcal{G}}_{i})$ to each previous task $i \in [t]$. The learning of each $\boldsymbol{g}_{j}$ for $j \in [k]$ is identical to the learning of $\boldsymbol{g}$, i.e., a combination of FSA and SL.
As for the exploitation of $\boldsymbol{g}_{1},...,\boldsymbol{g}_{k}$, before learning each task $i$, the most relevant $\boldsymbol{g}_{j}$ is first retrieved upon the current training samples $\mathcal{D}_{i}$ and then temporarily added to the backbone parameters $\theta$ to better incorporate task-specific knowledge (the improved $\theta$ is denoted as $\theta'$). 
The improved backbone $f_{\theta'}$ is used to obtain $\mathcal{G}_{i,c}$, i.e., Step 10 in Algorithm~\ref{alg:algorithm}. Whereas, the original backbone $f_{\theta}$ is still used to obtain $\hat{\mathcal{G}}_{i,c}$, i.e., Step 9 in Algorithm~\ref{alg:algorithm}. At the testing phase, the most relevant $\boldsymbol{g}_{j}$ is retrieved upon the current testing samples and temporarily added to $\theta$.

\end{document}